\documentclass[twoside,11pt,a4paper]{article}

\newcommand{\bhamstudentname}{Joseph Bartlett}
\newcommand{\bhamthesistitle}{Accelerated First Order Methods for Variational Imaging}
\newcommand{\bhamfronttitle}{Accelerated First Order Methods\\for Variational Imaging}
\newcommand{\bhamschool}{School of Computer Science}
\newcommand{\bhamcollege}{Engineering and Physical Sciences}
\newcommand{\bhamdegree}{MSc. Artificial Intelligence and Machine Learning}
\newcommand{\bhamid}{1811247}
\newcommand{\bhamsupervisor}{Dr. Jinming Duan}
\newcommand{\bhamyear}{2021}

\newcommand\zwe[1]{\rlap{${}^{#1}$}}

\usepackage[breaklinks = true]{hyperref}
\usepackage{fancyhdr}
\usepackage[sort]{natbib} 
\usepackage{comment} 
\usepackage{dirtree} 
\usepackage{longtable} 

\usepackage{amsmath}  
\usepackage{amsfonts} 
\usepackage{cleveref}
\usepackage[parfill]{parskip}

\usepackage{subfigure}
\usepackage{amsmath}
\usepackage[algo2e,ruled]{algorithm2e}
\usepackage{amsthm} 
\usepackage[affil-it]{authblk} 
\usepackage{amsopn} 
\usepackage{amssymb} 
\usepackage{mathrsfs} 
\usepackage{booktabs} 
\usepackage{natbib} 
\usepackage{tikz} 
\usepackage{algorithm} 
\usepackage{float} 
\usepackage{caption} 
\usepackage{abstract} 
\usepackage{listings} 
\usepackage{color} 

\newfloat{algorithm}{t}{lop}

\numberwithin{equation}{section}

\newtheorem{lemma}{Lemma}[section]
\newtheorem{theorem}[lemma]{Theorem}
\newtheorem{proposition}[lemma]{Proposition}

\theoremstyle{definition}
\newtheorem{definition}{Definition}[section]

\theoremstyle{remark}

\fancyhfoffset[L]{0cm} 
\newcommand{\HRule}{\rule{\linewidth}{0.5mm}}
\renewcommand{\headrulewidth}{0pt}

\newcommand{\minitab}{\hspace*{0.25em}}

\DeclareMathOperator*{\argminB}{argmin}
\DeclareMathOperator*{\argmaxB}{argmax}

\usepackage{perpage}
\MakePerPage{footnote} 

\pdfoutput=1 
\usepackage[margin=3cm]{geometry}
\usepackage{titling}	
\usepackage{titlesec}
\titleformat*{\section}{\normalsize	\bfseries}
\titleformat*{\subsection}{\small \bfseries}
\titleformat*{\subsubsection}{\footnotesize \bfseries}
\titlespacing*{\section} {0pt}{3ex plus 1ex minus .2ex}{2ex plus .2ex}
\titlespacing*{\subsection} {0pt}{2.25ex plus 1ex minus .2ex}{0.75ex plus .2ex}
\titlespacing*{\subsubsection}{0pt}{2.ex plus 1ex minus .2ex}{0.5ex plus .2ex}

\usepackage{bm}
\usepackage{pdfpages}
\usepackage{lastpage}
\usepackage{amsmath}
\usepackage{amsfonts}
\usepackage{amssymb}

\usepackage{epstopdf} 

\usepackage{listings}
\title{MSc. Project\\\bhamthesistitle}
\author{\textsf{\bhamstudentname  {\textsf{BSc (Hons)}}}}

\date{}
\begin{document}

\pagenumbering{gobble} 
\begin{titlepage}
\begin{center}
\begin{minipage}{6in}
  \centering
  \raisebox{-0.5\height}{\includegraphics[width=1.25in]{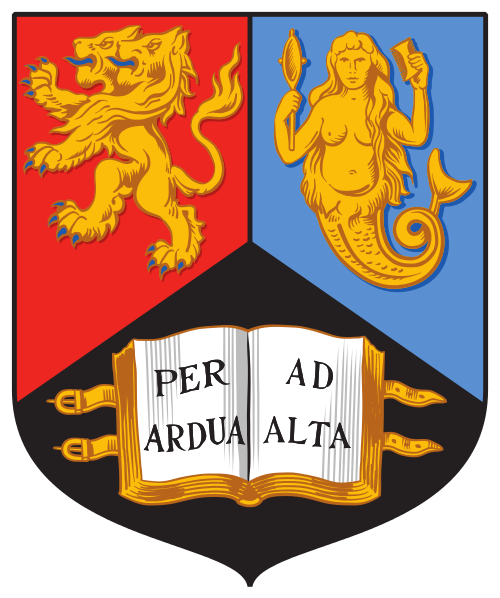}}
  \hspace*{.2in}
  \raisebox{-0.5\height}{\includegraphics[height=0.9375in]{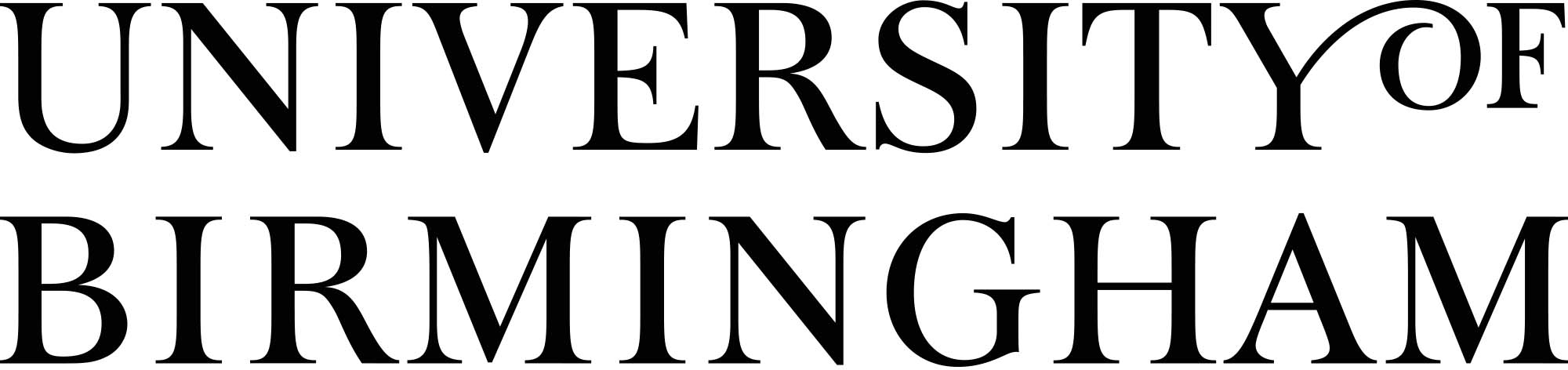}}
  \end{minipage}
  \\ [1.0cm]
\textsc{{\LARGE \bhamschool\\}College of \bhamcollege}\\[3.5cm]

\textsc{\Large MSc. Project}\\[0.5cm]

\HRule \\[0.4cm]
\begin{center}\Huge
\bhamfronttitle
\end{center}
\HRule \\[1.5cm]

\begin{center}
Submitted in conformity with the requirements\\ for the degree of \bhamdegree\\
\bhamschool\\ University of Birmingham\\
\vspace{2cm}
\bhamstudentname, BSc. (Hons)\\
Student ID: \bhamid\\
Supervisor: \bhamsupervisor      
\end{center}
\vfill

{\large September \bhamyear}

\end{center}
\end{titlepage}
\clearpage
\section*{Declaration}
The material contained within this report has not previously been
submitted for a degree at the University of Birmingham or any other university.
The research reported within this report has been conducted by the author
unless indicated otherwise.\\

\vfill
\clearpage
\begin{center}
\vspace*{\fill}
\begin{minipage}{6in}

\centering \Large{``If I have seen further it is by standing on the shoulders of Giants."}\\{\normalsize{\textsc{Isaac Newton}}}

  \end{minipage}
  \vspace*{\fill}
\end{center}
\clearpage
\subsection*{Acknowledgements}
In loving memory of my grandmother Mavis Kirk, who I would have loved to have shared this work with.

I would like to thank my project supervisor Dr. Jinming Duan, he has inspired me constantly with his enthusiasm and dedication. He has recognised my interests and introduced me to this fascinating area of study, and offered guidance and support throughout. 

I would like to thank my family and friends for their support throughout this project, their continuity and love has been invaluable.    

Finally I would like to thank Abigail for her patience and help throughout. 
\clearpage
\maketitle
\vspace{-5.5em} 
\begingroup
    \fontsize{9pt}{11pt}\selectfont
\tableofcontents
\endgroup
\clearpage
\phantomsection

\clearpage
\pagenumbering{arabic}
\setcounter{page}{1}
\lhead{}\chead{MSc. Project Report : \nouppercase{Section \thesection\minitab : \leftmark}}\rhead{}
\rfoot{Page \thepage \hspace*{0.2pt} of \pageref{LastPage}}
\renewcommand{\headrulewidth}{0.4pt}

\section{Overview}

\subsection{Abstract}
In this thesis, we offer a thorough investigation of different regularisation terms used in variational imaging problems, together with detailed optimisation processes of these problems. We begin by studying smooth problems and partially non-smooth problems in the form of Tikhonov denoising and Total Variation (TV) denoising, respectively. 

For Tikhonov denoising, we study an accelerated gradient method with adaptive restart, which shows a very rapid convergence rate. However, it is not straightforward to apply this fast algorithm to TV denoising, due to the non-smoothness of its built-in regularisation. To tackle this issue, we propose to utilise duality to convert such a non-smooth problem into a smooth one so that the accelerated gradient method with restart applies naturally. 

However, we notice that both Tikhonov and TV regularisations have drawbacks, in the form of blurred image edges and staircase artefacts, respectively. To overcome these drawbacks, I propose a novel adaption to Total Generalised Variation (TGV) regularisation called Total Smooth Variation (TSV), which retains edges and meanwhile does not produce results which contain staircase artefacts. To optimise TSV effectively, I then propose the Accelerated Proximal Gradient Algorithm (APGA) which also utilises adaptive restart techniques. Compared to existing state-of-the-art regularisations (e.g. TV), TSV is shown to obtain more effective results on denoising problems as well as advanced imaging applications such as magnetic resonance imaging (MRI) reconstruction and optical flow. TSV removes the staircase artefacts observed when using TV regularisation, but has the added advantage over TGV that it can be efficiently optimised using gradient based methods with Nesterov acceleration and adaptive restart.  Code is available at \href{https://github.com/Jbartlett6/Accelerated-First-Order-Method-for-Variational-Imaging}{https://github.com/Jbartlett6/Accelerated-First-Order-Method-for-Variational-Imaging}.
\newpage

\clearpage
\section{Introduction}
\subsection{Introduction to Variational Models}
Variational imaging models cover a group of models which can be applied to a wide range of imaging problems. They consist of finding an optimum value to some objective function, which has been designed to produce the required solution. Such objective functions contain a data fidelity term and a regularisation term. The data fidelity term ensures that the solution remains similar to the input data in some way; and the regularisation term allows our prior knowledge to influence the solution. In the case of variational methods, the regularisation term contains the image gradient, as we try to minimise this value in an attempt to remove high frequency components of the image, and therefore remove noise. 

A variational method is an example of a model based method that does not require large amounts of data, unlike popular data driven methods such as deep learning, that by definition do. They therefore are effective at tasks where ground truth data is expensive or likely impossible to obtain. Another case when data is not available is image denoising, when only one image with the given noise distribution is available. These are just some scenarios where model based methods, such as variational networks, have an advantage over popular data driven methods.  

In this project I will investigate two major components of variational imaging models: the optimisation algorithms applied to the objective function and the regularisation term within the objective function. These two components go hand in hand. Regularisation affects the properties of the solution first and foremost, but has a strong relationship with the optimisation algorithms which can be applied to the problem, which dictate how long it takes to reach the solution. By considering these two factors together, I aim to improve the quality of solutions obtained by applying variational methods to imaging problems as well as reducing the time taken to obtain such images. 

To extend upon this, many state-of-the-art deep learning architectures are now using model based methods to improve architectures and produce the best results on a range of imaging tasks. This highlights the need for research in the area of model based methods in order to find new avenues to explore in the domains of both data driven methods and model based methods. Variational imaging is particularly important in this regard as architectures, such as the variational network by \cite{hammernik2018learning} and VS-Net by \cite{duan2019vs}, achieve state-of-the-art results in their respective fields by using ideas from variational imaging.

The area I specifically address is applying adaptive restart techniques to a range of problems with different regularisation terms. This allows gradient based optimisation algorithms to be accelerated effectively when applied to such problems. I show the efficacy of this method applied to two established regularisation terms: Tikhonov regularisation and total variation denoising. In both scenarios, this method achieves positive results, however both methods have their shortfalls. Tikhonov denoising produces blurry images, with poor edge retention, and total variation introduces staircase artefacts. 

Total generalised variation is a regularisation term designed to overcome the problems faced by Tikhonov and total variation denoising. However, it cannot be optimised using adaptive restart methods. I therefore propose some changes to total generalised variation in order to obtain a regularisation term which maintains the desirable properties of its solution, and can be solved using adaptive restart methods. By doing so I obtain a novel regularisation term - total smooth variation, which brings together the positive properties of total generalised variation regularisation and the quick convergence rate of adaptive restart methods. 

\subsection{Literature Review}
\subsubsection{Optimisation}
Since the advent of the backpropagation algorithm, which has enabled the training of neural networks via gradient descent, optimisation has come to the forefront of machine learning - however, in this thesis we will focus on convex optimisation. Gradient descent has long been an established method in optimisation and \cite{zbMATH03850830} proposed additional acceleration schemes to improve the rate of convergence from $\mathcal{O}(1/k)$ to $\mathcal{O}(1/k^2)$. When the strongly convex parameter is known, it was also shown that such acceleration methods can achieve linear convergence. Nesterov extends this to strongly convex problems where the strongly convex parameter is unknown \cite{nesterov2013gradient}. The method proposed involves starting with some initial estimate, and adapting this value depending on the convergence of the function. 

The method previously mentioned lacked elegance, despite providing insight into how to deal with strongly convex problems. The idea was then extended to heuristic, adaptive restart methods in \textit{Adaptive Restart for Accelerated Gradient Schemes} by \cite*{o2015adaptive}. This paper illustrates how effective such adaptive restart methods can be on a range of complex examples. The two adaptive restart schemes use cheap observations at each iteration to decide whether to restart the algorithm from the current iterate. Additionally, \cite{chambolle2016introduction} consider a range of continuous optimisation methods applied to imaging problems in \textit{An introduction to continuous optimization for imaging}. They acknowledge the findings from the paper written by \cite{o2015adaptive}, as well as considering other established algorithms, such as the Alternating Direction Method of Multipliers (ADMM) and the Primal Dual algorithm (PD). They also offer effective algorithms for the optimisation of denoising, MRI reconstruction and optical flow problems.

ADMM, originally proposed by \cite{gabay1976dual} has since become a very popular method and a large amount of work has been carried out on its convergence analysis. In \textit{Distributed Optimization and Statistical Learning Via the Alternating Direction Method of Multipliers} by \cite{boyd2011distributed}, the method of ADMM was applied to statistical problems, and a thorough outline of the algorithm is given. It has since been applied to a huge range of problems from support vector machines to sparse signal recovery. Common patterns which occur when applying ADMM to problems are also outlined, as well as a guide on how to apply ADMM to functions containing the $L_{1}$ norm. 

The fast ADMM algorithm (an acceleration scheme applied to standard ADMM) was  proposed by \cite{goldstein2014fast}. This method uses the same acceleration scheme proposed by Nesterov, only applied to ADMM instead. Additionally, a restart method for ADMM which relies on the primal and dual errors was proposed; however, this method has no convergence guarantee. Fast ADMM was then applied to a range of denoising and reconstruction problems, and was shown to converge quicker than regular ADMM. 
 
\cite{chambolle2011first} also introduced a primal dual algorithm, specifically to solve problems involving total variation (TV) regularisation. This algorithm is applied directly to the primal dual objective function of the TV denoising problem, which is a mini-max problem. Such problems have recently gained a lot of interest in optimisation, specifically as they are relevant to training generative adversarial networks. 

Recently a state-of-the-art algorithm for optimising mini-max problems was proposed by \cite{lin2020near}. The algorithm achieves a near optimal convergence rate on strongly convex, strongly concave problems, and matches the performance of other algorithms on non-strongly convex/concave problems. They also outline how the method can be accelerated using Nesterov acceleration, on strongly convex, strongly concave problems, however no outline of restart methods are given when applying the method to non-strongly convex/concave problems. Such a method has potential to be applied to the primal dual formulation of variational methods discussed in this paper. 

\subsubsection{Regularisation}
\cite{tikhonov1963solution} regularisation was the original method of regularisation used to solve ill-posed inverse problems, utilising the favourable properties of the $L_{2}$ norm squared. The shortfalls of Tikhonov denoising however, are obvious, as the edges become unclear and the whole image becomes too blurred. These issues are discussed by \cite{rudin1987images} in his PhD thesis \textit{Images, Numerical Analysis of Singularities and Shock Filters} suggesting that the $L_{1}$ norm is a better fit for imaging problems. TV denoising is then formulated fully by \cite{rudin1992nonlinear} in their paper \textit{Nonlinear Total Variation Based Noise Removal Algorithms}. They propose the TV regularisation problem and then solved it using the gradient projection method.

Since its introduction, the staircase artefacts which appear in images upon which TV regularisation is applied, have now been widely discussed in the literature, and several methods of overcoming these shortfalls have been proposed - the main solution being higher order regularisation. In \textit{A Fast Total Variation Minimization Method for Image Restoration} by \cite{huang2008fast}, a fitting term was introduced to decouple the regularisation and data terms. This finding allowed for a fast optimisation algorithm to solve the problem of image restoration. The Total Generalised Variation (TGV) was introduced by \cite{bredies2010total}, and this successfully overcame the staircase artefacts of TV, and has been applied to more advanced problems such as MRI Reconstruction by \cite{https://doi.org/10.1002/mrm.22595}. On such a problem, it was shown to yield impressive empirical results. The current trend in regularisation is moving away from model based regularisation towards learnt regularisation. \cite{kobler2020total} introduces such regularisation, and shows its strong performance on a range of problems. Such methods build on the ideas already used in current regularisation.

\subsubsection{MRI Reconstruction}
MRI reconstruction has developed a lot since first being introduced by \cite{lustig2007sparse}. Initial methods used TV regularisation to exploit sparsity of the image in some domain. This domain varies between the wavelet domain and the finite difference domain – which is explored later in this thesis to allow the Nyquist Shannon limit to be violated and accurate images to still be reproduced. In \textit{Second Order Total Generalized Variation (TGV) for MRI} by \cite{knoll2011second}, it is illustrated that TGV produced results which are better to view than that of TV - the staircase artefacts are removed and sharp edges are retained. This was shown to be the case on several medical images.

Since these developments, MRI reconstruction has been approved by the FDA for medical diagnosis, opening the potential for increased research into the area. Research has shifted towards deep learning methods of MRI reconstruction, due to real time inference and high accuracy [\cite{ye2019compressed}]. However, they have the disadvantage that they require data rather than being a model based method. One such example is given by \cite{schlemper2017deep}, in which a deep convolutional network is applied to dynamic MRI reconstruction problems. 
Many of the state-of-the-art deep learning methods implement a combination of deep learning and model based methods, such as in VS Net by \cite{duan2019vs}, ADMM Net by \cite{sun2016deep} and variational networks by \cite{hammernik2018learning}. Hence it is important that model based methods are still developed to optimise the output of such deep learning algorithms. 

\subsubsection{Optical Flow}
Optical flow varies from other methods in the sense that there is no ground truth data for a neural network to be trained on. Therefore, optimisation methods generally achieve state-of-the-art accuracy in this field. The original work on optical flow was proposed by \cite{horn1981determining} - they proposed a model and an algorithm to solve the optical flow problem. Since then, different types of regularisation and data terms such as $L_{1}$ TV have been used to solve the optical flow problem. Different optimisation techniques can be used, with dual proximal algorithm being preferred by \cite{zach2007duality}. TGV has also built on the success of TV to incorporate smooth gradients in the image [\cite{fortun2015optical}]. Voxelmorph is an unsupervised deep learning method  for optical flow, proposed by \cite{balakrishnan2019voxelmorph}. The deep learning architecture learns the function which maps the pixels between the two images. This method offers real time inference, and doesn't require ground truth data; however, it doesn’t perform better than iterative methods in terms of accuracy. 

\subsubsection{My Contributions}
In this thesis, I will expand on the works by \cite{bredies2010total} and \cite{o2015adaptive}, on TGV and adaptive restart, respectively, and then apply this to the advanced imaging applications highlighted in the literature review. The research conducted on adaptive restart methods is an area which has been under-utilised in the field of variational imaging models, as there is no material that highlights the empirical improvement of results which can be obtained by applying this to such problems. 

Then, using this as motivation we analyse different regularisation terms for which this method can be applied to. For total variation this is straight forward, and works as a baseline, however, we reiterate the shortfalls of this regularisation already present in the literature. We introduce TGV and then adapt this regularisation so the adaptive restart techniques established by \cite{o2015adaptive} can be applied. By utilising state-of-the-art methods in both optimisation and regularisation, we can obtain an algorithm which can be applied to advanced imaging techniques. MRI reconstruction is primarily used as an illustrative example; due to the accessibility of data, the field is now dominated by deep learning. However, due to the fact that there is no ground truth data for optical flow, model-based optimisation techniques are still state-of-the-art in this field, especially in terms of accuracy. 

Not only does this research apply state-of-the-art methods to important current problems, it highlights some of the pitfalls, as well as adaptations which have to be made in order to fit such methods to current problems. Optimisation and regularisation go hand in hand, but the following piece of work joins two pieces of research which have been carried out in the field together in order to obtain strong results on a range of problems. It also compares the developed method with other state-of-the-art model-based methods, highlighted in the literature review, to offer a sufficient overview of the field. Throughout, I compare methods in terms of perceptual quality, quantitative results and convergence rate. 

My main contributions throughout this piece of work can be summarised as follows:
\begin{itemize}
    \item I extend the application of adaptive restart methods to partially non-smooth problems using the Accelerated Dual Proximal Algorithm (ADPA). Such restart methods illustrate a significant improvement on applying the un-accelerated dual proximal gradient algorithm as well as accelerated proximal gradient descent without restart.
    \item A novel regularisation term, total smoothed variation, is introduced, which performs comparably with the state-of-the-art TGV regularisation. However, this can be optimised effectively using the ADPA algorithm.
    \item A comparison of a range of state-of-the-art algorithms and regularisation terms applied to different imaging problems is carried out, illustrating the advantages and disadvantages of different methods. 
\end{itemize}

\subsection{Report Structure}
The structure of my thesis from this point onward is as follows: in section \ref{sec:MathBack}, I introduce some mathematical basics which are required to understand the problems analysed in this thesis, as well as my reasoning for the changes made in some of the methods. Then in section \ref{sec:Tikh}, I introduce the idea of adaptive restart methods, and illustrate their empirical results on Tikhonov denoising. This example is used to not only illustrate the shortfalls of Tikhonov denoising but also illustrate how effective the adaptive restart methods can be. Then in section \ref{sec:TV}, total variation denoising is introduced to overcome the shortfalls of Tikhonov denoising, as well as introduce some concepts on non-smooth optimisation. Here I apply restart to the dual proximal algorithm, illustrating how this method can be generalised to partially non-smooth problems. In section \ref{sec:TSV}, I introduce TGV and how it overcomes the shortfalls of TV, and then explain how TGV can be adapted to obtain the novel regularisation term Total Smooth Variation (TSV). Then I apply adaptive restart methods to the TSV denoising problem. Finally in section \ref{sec:Advanced}, I show how TSV denoising can be applied to advanced imaging problems, both generally and then applied specifically to MRI reconstruction and optical flow.

\section{Mathematical Background}\label{sec:MathBack}
\subsection{Multivariable Calculus}
This thesis is based around optimising multivariate, scalar functions, specifically optimising multivariate, objective functions, $f: \mathbb{R}^{mn} \mapsto \mathbb{R}$. We will focus on functions for which the input is an $m$ by $n$ image, $\mathbf{u}\in \mathbb{R}^{mn}$, which has been converted into a vector using column vector ordering. All derivations throughout this paper will be done using this column vector notation of images, however, it should be noted that for many of the algorithms, they can be applied using equivalent operations applied to matrices, in an element-wise manner. 

Bold letters will be used to represent vectors in the image space e.g. $\mathbf{u}\in \mathbb{R}^{mn}$, and non-bold face letters will be used to represent vectors at individual pixels in the image e.g. $u_{i} \in \mathbb{R}^{k}$, where $i\in\{1,...,mn\}$ represents the pixel position and $k \in \mathbb{N}$ is how many channels the image has. In some cases vectors in the image space will have to be split into their respective channels, in which case boldface letters with subscripts will be used to distinguish the channels e.g. $\mathbf{p}_{1}$ and $\mathbf{p}_{2}$ represent the first and second channels of some 2-channeled vector, $\mathbf{p}$.  

\subsection{Finite Differences}\label{subsec:MathBack:FinDiff}
The objective functions applied throughout typically consist of a data term and a regularisation term (this will be covered in more detail later in the thesis). The regularisation term in the case of variational methods puts some constraint on the change of intensity in both the $x$ and $y$ dimensions of the image. For continuous functions, the gradient is used to measure the instantaneous rate of change at any given input of the function. Since the intensity of an image can be viewed as a discrete function (taking pixel position as input and outputting an intensity value) we can use the discretised gradient known as finite differences. Finite differences follow the same idea as continuous differentiation only it measures the rate of change between neighbouring pixels instead of over infinitesimally small distances. The formulas used to calculate both the forward and backwards finite difference of a 2-dimensional image can be seen in Definition \ref{def:findiff}. The following concepts are defined in the image space and then convert to the vector space. 
\begin{definition}\label{def:findiff}
For some function ${\rm{I}}: \{(x,y) \in \mathbb{Z}^{2}$ $|$ $x \in [1,m],y \in [1,n]\} \mapsto \mathbb{R}$ the forward and backwards finite differences in the $x$ and $y$ directions are as follows:
\begin{itemize}
    \item Forward Finite Differences:
        \begin{center}
            \[\partial_{x}{\rm{I}}(x,y) = \frac{{\rm{I}}(x+1,y) - {\rm{I}}(x,y)}{1} = {\rm{I}}(x+1,y) - {\rm{I}}(x,y)\]
            \[\partial_{y}{\rm{I}}(x,y) = \frac{{\rm{I}}(x,y+1) - {\rm{I}}(x,y)}{1} = {\rm{I}}(x,y+1) - {\rm{I}}(x,y)\]
        \end{center}
    \item Backward Finite Differences:
        \begin{center}
            \[-\partial^{*}_{x}{\rm{I}}(x,y) = \frac{{\rm{I}}(x,y) - {\rm{I}}(x-1,y)}{1} = {\rm{I}}(x,y) - {\rm{I}}(x-1,y)\]
            \[-\partial^{*}_{y}{\rm{I}}(x,y) = \frac{{\rm{I}}(x,y) - {\rm{I}}(x,y-1)}{1} = {\rm{I}}(x,y) - {\rm{I}}(x,y-1)\]
        \end{center}
\end{itemize}
\end{definition}
Calculating the finite differences of an image can be done using neighbouring pixels to calculate the value at any given pixel. This can in-fact be done via convolution, using a simple finite difference kernel. However it is slightly more challenging when the image has been converted to a vector, yet since the size of the original image is known to be $m$ by $n$ it is clear that: 
\begin{center}
    ${\rm{I}}(x,y) = u(i) \implies {\rm{I}}(x+1,y) = u(i+m)$
\end{center}Therefore when we convert the image into a vector we retain the ability to calculate the finite difference in both dimensions. 

The values of these finite differences will be undefined at some positions, specifically at $\partial_{x}{\rm{I}}(n,y)$ for all $y$, $\partial_{y}{\rm{I}}(x,m)$ for all $x$, $-\partial^{*}_{x}{\rm{I}}(1,y)$ for all $y$ and $-\partial^{*}_{y}{\rm{I}}(x,1)$ for all $x$. At these positions, we are required to define boundary conditions, i.e. what occurs at pixels which are not defined by the image. There are many common boundary conditions the choice of such conditions can significantly affect the appearance of the output image, as well as some aspects of how the methods are applied. The two types of boundary conditions which I will consider throughout my project are: 
\begin{itemize}
    \item Symmetric Boundary Condition
        \begin{equation}\label{eq:symBound}
            {\rm{I}}(n+1,y) = {\rm{I}}(n,y)
        \end{equation}
    \item Periodic Boundary Conditions
        \begin{equation}\label{perBound}
            {\rm{I}}(n+1,y) = {\rm{I}}(1,y)
        \end{equation}
\end{itemize}

Finite differences will usually be calculated for the image as a whole, rather than for individual pixels. This can be written as a matrix multiplication applied to the vector form of the image. For the majority of the paper I will be considering symmetric boundary conditions, therefore in \ref{eq:Tn} I have only included the 1D forward finite difference matrix over a vector of length $n$ with symmetric boundary conditions, denoted $\mathcal{T}_{n} \in \mathbb{R}_{n \times n}$:
\begin{equation} \label{eq:Tn}
\mathcal{T}_{n} = \left( {\begin{array}{*{20}{r}}
{-1}&1&&&&\\
&{-1}&1&&&\\
&&-1&1&&\\
&&&\ddots&\ddots&\\
&&&&-1&1\\
&&&&&-1
\end{array}} \right)
\end{equation}
The backwards finite difference matrix is equal to $-\mathcal{T}_{n}^{T}$. In case of 2D images, some modifications have to be made, which utilise the structure of the image when in vector form. It is however easy to generalise the above matrix to 2D using the kronecker product. The  2D finite difference matrices, when applied to a signal $\mathbf{u} \in \mathbb{R}^{mn}$, with original dimensions were $m \times n$, for the $x$ and $y$ dimensions, respectively, are as follows:
\begin{align}\label{eq:LxLy}
\begin{split}
L_{x} &= \mathcal{T}_{n} \otimes {\cal{I}}_{m}\\
L_{y} &= {\cal{I}}_{m} \otimes \mathcal{T}_{n} 
\end{split}
\end{align}
where ${\cal{I}}_{n} \in \mathbb{R}^{n}$ is the $n$ by $n$ identity matrix, and $\otimes$ is the kronecker product. The backwards finite differences in the $x$ and $y$ directions are $-L_{x}^{T}$ and $-L_{y}^{T}$ respectively. 

Now that we have defined the forward and backward finite difference matrices, we can discuss how they will fit into the more general notation used when deriving the algorithms in the rest of the thesis. 
\begin{definition}\label{def:ImgGrad}
Suppose $\mathbf{u} \in \mathbb{R}^{mn}$ then the discrete gradient operator $\nabla$: $\mathbb{R}^{mn} \mapsto \mathbb{R}_{mn \times 2}$, when applied to $\mathbf{u}$ is defined as follows:
\[\nabla \mathbf{u} = (L_{x}\mathbf{u},L_{y}\mathbf{u})\]
\end{definition}

$\nabla \mathbf{u}$ is therefore a 2-channelled image. The divergence operator $\nabla^{T}$ which can only be applied to a 2-channelled image and is defined as follows:
\begin{definition} \label{def:ImgDiv}
Suppose $\mathbf{p} \in \mathbb{R}_{mn \times 2}$ then the divergence operator $\nabla^{T}$: $\mathbb{R}_{mn \times 2} \mapsto \mathbb{R}^{mn}$, when applied to $\mathbf{p}$, is defined as follows:
\[\nabla^{T}\mathbf{p} = -L_{x}^{T}\mathbf{p}_{1} - L_{y}^{T}\mathbf{p}_{2}\]
\end{definition}
The gradient and divergence operators are closely related, this is illustrated by the following identity, which relates the two through the dot product:
\begin{equation}\label{eq:AdjointIdentity}
    \langle \nabla \mathbf{u,p} \rangle = - \langle \mathbf{u},\nabla^{T}\mathbf{p} \rangle 
\end{equation}
The derivation of this identity, can be seen below:
\begin{align*}
    \langle \nabla \mathbf{u,p} \rangle =& \langle L_{x} \mathbf{u, p}_{1} \rangle + \langle L_{y} \mathbf{u, p}_{2} \rangle \\
    =& (L_{x} \mathbf{u})^{T}\mathbf{p}_{1} + (L_{y} \mathbf{u})^{T}\mathbf{p}_{2} \\
    =& \mathbf{u}^{T}L_{x} ^{T} \mathbf{p}_{1} + \mathbf{u}^{T}L_{y} ^{T} \mathbf{p}_{2}\\
    =& \langle \mathbf{u},L_{x}^{T}\mathbf{p}_{1} \rangle + \langle \mathbf{u},L_{y}^{T}\mathbf{p}_{2} \rangle \\
    =& - \langle \mathbf{u},-L_{x}^{T} \mathbf{p}_{1} \rangle - \langle \mathbf{u},-L_{y}^{T} \mathbf{p}_{2} \rangle \\
    =& -\langle \mathbf{u}, -L_{x}^{T} \mathbf{p}_{1}-L_{y}^{T} \mathbf{p}_{2}\rangle \\
    =& - \langle \mathbf{u},\nabla^{T} \mathbf{p}\rangle \\
\end{align*}

\subsection{Vector Norms}
A norm can be used as a measure of size and proximity in $\mathbb{R}^{2}$ and $\mathbb{R}^{3}$ [\cite{horn1990norms}]. Norms have been studied in great deal for their abstract properties, however in this thesis we will only be considering the practical applications of norms. Specifically, we use them as a measure of proximity. Throughout this thesis we only concentrate on the $L_{2}$ norm. Special care is required when dealing with norms in the case of images, as global image norms, and individual pixel norms must be defined. 
\begin{definition} \label{def:EuclideanVectorNorm}
    Suppose we have some pixel $u$ in a $k$-channelled image, i.e.$u \in \mathbb{R}^{k}$, then the Euclidean vector norm, $|\cdot|:\mathbb{R}^{k} \mapsto \mathbb{R}$ is defined as:
    \[|u| = \sqrt{\sum_{i=1}^{k} u_{i}^{2}}\]
\end{definition}
In relation to my work, this is the notation which will be used when applying a norm to a vector at a pixel position within an image. Often, the $L_{2}$ norm squared must be considered, denoted $|\cdot|^{2}$ - in this case, the norm is simply the sum of squared components of the vector. 

Often, when dealing with images, a measure of proximity is required, so we will now define the squared euclidean norm for vectors in the image space: 
\begin{definition}\label{def:EuclideanImageNorm}
    Suppose we have some $k$-channeled image $\mathbf{u} \in \mathbb{R}_{mn \times k}$ then the Euclidean vector norm squared in the image space, $\|\cdot\|_{2}^{2}:\mathbb{R}_{mn \times k} \mapsto \mathbb{R}$ is defined as follows:
    \[\|\mathbf{u}\|^{2}_{2} = \sum_{i=1}^{mn}|u_{i}|^2\]
\end{definition}
In this case, $mn$ is the number of pixels in the image, and $k$ is the number of channels. Throughout this thesis, images will commonly either be one or two channeled, Only in the case of optical flow will images with more than two channels be considered.

\subsection{Vector Calculus}
Ontop of the requirement to calculate the finite difference of images, another key mathematical aspect of this thesis is to be able calculate the derivative of the objective function. This, however, differs from the case of finding the finite differences of the images, as it is in a continuous setting. The first key definition is to be able to find the gradient of a multivariate scalar function:
\begin{definition}\label{def:ObjGrad}
        Suppose $f:\mathbb{R}^{mn} \rightarrow \mathbb{R}$ then the gradient of the function $f$ denoted $\nabla f$ evaluated at $\mathbf{u} \in \mathbb{R}^{mn}$
        is a column vector i.e. $\nabla f(\mathbf{u}) \in \mathbb{R}^{mn}$ where each component is as follows, for $i=1,...,mn$:
            \[\nabla f(\mathbf{u})_{i} = \frac{\partial f}{\partial u_{i}}(\mathbf{u})\] 
\end{definition}
The gradient has many uses, such as calculating the first order Taylor expansion of the function $f$ as well as allowing us to identify turning points and perform gradient descent. Many of these applications will be discussed later in the thesis. 
\begin{definition}\label{def:ObjHess}
    Suppose $f:\mathbb{R}^{mn} \rightarrow \mathbb{R}$, then the Hessian matrix of the function $f$, denoted $\nabla^{2}f(\mathbf{u})$, is an $mn$ by $mn$ matrix, i.e. $\nabla^{2}f(\mathbf{u}) \in \mathbb{R}_{mn \times mn}$ whose entries are defined as follows for $i,j = 1,...,mn$:
    \begin{center}
    \[\nabla^{2}f(\mathbf{u})_{ij} = \frac{\partial^{2}f}{\partial u_{i} \partial u_{j}}(\mathbf{u})\] 
    \end{center}
\end{definition}
The Hessian matrix generalises the idea of second derivatives to multivariate functions. The Hessian matrix is used when calculating the second order Taylor expansion and plays a significant part in convex analysis. 
\subsection{Convex Functions}
All of the objective functions in this thesis are convex functions. This means they have favourable properties with respect to optimisation. Below are some key definitions which are used throughout, as well as some discussion about these key concepts. 
\begin{definition}\label{def:Convex}
        A function $f : \mathbb{R}^{mn} \rightarrow \mathbb{R}$ is convex if the domain of $f$, ${\rm{dom}}(f)$, is a convex set and if for all $\mathbf{u,v} \in$ ${\rm{dom}}(f)$ and $\theta$ with $0 \leq \theta \leq 1$ we have:
        \[f(\theta \mathbf{u}+(1-\theta)\mathbf{v})\leq \theta f(\mathbf{u}) + (1-\theta)f(\mathbf{v})\]
        
\end{definition}
It is difficult to determine whether a function is convex or not solely from this definition, however it offers useful insight into what it means for a function to be convex. For two points $\mathbf{u,v}$ $\in$ ${\rm{dom}}(f)$, the function $f$ lies below the segment connecting the point $(\mathbf{u},f(\mathbf{u}))$ and $(\mathbf{v},f(\mathbf{v}))$. Some of the nice properties that follow on from this definition are that the Taylor expansion of the function is a global underestimator of any convex function, and also any point $\mathbf{u}$ such that $\nabla f(\mathbf{u}) = 0$ is a global minimiser of a convex function.

There are both first order and second order conditions for convexity which can be used. The following proposition relates the eigenvalues of the Hessian matrix of $f$ to whether or not a function is convex.
\begin{proposition}\label{prop:ConvexHess}
    Assume a function, $f:\mathbb{R}^{mn} \rightarrow \mathbb{R}$, is twice differentiable, then $f$ is convex if and only if its Hessian matrix is positive semi-definite for all $\mathbf{u}\in {\rm{dom}}(f)$, i.e. 
            \[\nabla^{2}f(\mathbf{u}) \succeq 0\] 
\end{proposition}
This is a useful way of identifying whether a function is convex or not as the Hessian is often easily obtained. This condition can easily be simplified down to whether the eigenvalues of the Hessian matrix of $f$ are bounded below by 0. 

Another property which is useful in the analysis of convex functions is to have a bound on the rate of change of the function i.e. a bound on the derivative of the function.

\begin{definition}\label{def:lSmooth}
        Suppose the function $f:\mathbb{R}^{mn} \rightarrow \mathbb{R}$ is differentiable, then $f$ is said to be a $\ell$-smooth function if for all $\mathbf{u},\mathbf{v} \in {\rm{dom}}(f)$ the following holds:
            \[\|\nabla f(\mathbf{u}) - \nabla f(\mathbf{v})\|_{2} \leq \ell \|\mathbf{u} - \mathbf{v}\|_{2}\]
\end{definition}

The $\ell$-smooth parameter allows us to make statements about how quick the gradient of the function can change as the input of the function changes. Knowing the $\ell$-smooth parameter is particularly important when deciding the step size of gradient based methods. Once again it is difficult to determine the $\ell$-smooth parameter directly from Definition \ref{def:lSmooth}, so we instead use Proposition \ref{prop:ellHess}.

\vspace{1pt}
\begin{proposition}\label{prop:ellHess}
    Suppose the function $f:\mathbb{R}^{mn} \rightarrow \mathbb{R}$ is twice differentiable, then $f$ is $\ell$-smooth if and only if:  
    \[0 \preceq \nabla^{2}f(\mathbf{u}) \preceq \ell I\]  for all $\mathbf{u} \in$ {\rm{dom($f$)}}.
\end{proposition}
\vspace{1pt}

So if the eigenvalues of the Hessian of $f$ are all bounded above by some value $\ell$ then we have that $f$ is $\ell$-smooth. Therefore, to find this value we can simply find the largest eigenvalue of the Hessian matrix. Another property which can be useful when accelerating gradient based optimisation algorithms is strong convexity.

\begin{definition}\label{def:strongConv}
    Suppose $f:\mathbb{R}^{mn} \rightarrow \mathbb{R}$ is differentiable, then $f$ being strongly convex with strong convexity parameter $0<\mu \in \mathbb{R}$ is equivalent to dom($f$) being convex and:
        \[f(\mathbf{v}) \geq f(\mathbf{u}) + \nabla f(\mathbf{u})^{T}(\mathbf{v}-\mathbf{u}) + \frac{\mu}{2}\|\mathbf{v}-\mathbf{u}\|^{2}_{2}\] for all $\mathbf{u,v} \in$ dom($f$)
\end{definition}

This definition can in fact be extended to non-differentiable functions as the gradient in the second component can be replaced by the sub-gradient. The following proposition provides a method of calculating the strong convexity parameter: 

\begin{proposition}\label{prop:muHess}
    Suppose $f:\mathbb{R}^{mn} \rightarrow \mathbb{R}$ is twice differentiable and ${\rm{dom(}}f{\rm{)}}$ is a convex set, then $f$ is strongly convex with strong convexity parameter $0<\mu \in \mathbb{R}$ if and only if 
    \[\nabla^{2} f(\mathbf{u}) \succeq \mu I\] for all $\mathbf{u} \in {\rm{dom}}(f)$.
\end{proposition}

Which means that the eigenvalues of the Hessian matrix are bounded below by the strong convexity parameter $0<\mu \in \mathbb{R} $. If a function is both strongly convex and $\ell$-smooth, then all of the eigenvalues of the Hessian matrix will lie in the interval $[\mu,\ell]$. When this is the case, the condition number of the function $f$ can be defined to be as follows.

\begin{definition}\label{def:condNum}
    Suppose $f:\mathbb{R}^{mn} \rightarrow \mathbb{R}$ is $\ell$-smooth and strongly convex with strong convexity parameter $\mu$, then the condition number $\kappa$ of the function $f$, is defined as follows:
        \[\kappa = \frac{\ell}{\mu}\]
\end{definition}
Generally, a function with a large condition number is said to be ill-conditioned and has poor convergence properties. This will become apparent in later sections when discussing the convergence properties of various algorithms.

One final topic I would like to introduce is the first order optimality conditions for functions, this is highlighted in the below theorem:
\begin{theorem}\label{thm:FirstOrderOpt}
Let $f: \mathbb{R}^{mn} \mapsto \mathbb{R}$ be a function defined on a set ${\rm{dom}}(f) \subseteq \mathbb{R}^{n}$. Suppose that $\mathbf{u}^{\star}$ is a local optimum and that all of the partial derivatives of f exist at $\mathbf{u}^{\star}$. Then $\nabla f(\mathbf{u}^{\star}) = \mathbf{0}$
\end{theorem}
This condition is particularly useful with regards to strongly convex functions as this first order optimality condition is sufficient for $\mathbf{u}^{\star}$ to be the global optimum of the function.

\clearpage

\section{Smooth Optimisation - Tikhonov Denoising}\label{sec:Tikh}
\subsection{Introduction to Tikhonov Denoising}
To illustrate the properties of gradient descent and some of the methods of accelerating this algorithm,  I will use the example of Tikhonov denoising. To perform Tikhonov denoising, the following objective function must be optimised
\begin{equation} \label{eq:TikhObj}
\argminB_{\mathbf{u}} \frac{1}{2}  \|\mathbf{u}-\mathbf{f}\|_{2}^{2}  + \frac{\lambda}{2} \|\nabla \mathbf{u}\|_{2}^{2}    
\end{equation}

such that $\lambda \in \mathbb{R}$ and $\mathbf{u,f} \in \mathbb{R}^{mn}$, where $\mathbf{f}$ is the image which is being denoised. The following, $\|\mathbf{u}-\mathbf{f}\|_{2}^{2}$, is known as the data fidelity term, which penalises the solution if it is too far away from the input data, $\mathbf{f}$. The regularisation term, $\|\nabla \mathbf{u}\|_{2}^{2}$, is a means of influencing the solution using our prior knowledge. In the case of variational methods, we use the prior knowledge that a denoised image will be smooth, i.e. have fewer large changes in intensity. This is desirable since many large changes in intensity generally corresponds to noise. 

The parameter $\lambda$ is the regularisation or smoothing parameter which balances the impact of the data term and the regularisation term. The larger $\lambda$ is, the smoother the resulting output image will be; the smaller $\lambda$ is, the more similar to the original input image the resulting solution will be. All problems considered in this thesis are composed of both a data and a regularisation term. The value of $\lambda$ can be selected on an image by image basis, or if being applied to a dataset, cross validation can be used. This holds for all variational methods to follow.   

 Tikhonov denoising has some very favourable properties in terms of optimisation, namely, that the objective function is smooth and strongly convex. The smoothness property allows us to apply widely studied smooth optimisation techniques such as gradient descent; strong convexity allows us to optimally accelerate gradient descent applied to this problem.  

Since both the objective function and its derivative are smooth with respect to $\mathbf{u}$, it can be differentiated twice to obtain the following Hessian matrix
\[I - \lambda\nabla^{T}\nabla \]
The eigenvalues of this matrix can be calculated using a DCT inversion, more details of which can be found in Appendix \ref{App:FourInv}. The eigenvalues range from $1$ to $8\lambda + 1$ by referring to Definitions \ref{def:lSmooth}, \ref{def:strongConv} and \ref{def:condNum} we can see the function has the following properties:
\begin{itemize}
    \item $\ell = 8\lambda + 1$
    \item $\mu = 1$
    \item $\kappa = \frac{8\lambda+1}{1} = 8\lambda + 1$
\end{itemize}
Since the eigenvalues are bounded from below by some positive value, we have that the objective function is strongly convex, hence the condition number of the function is well defined and can easily be calculated. Clearly, the more blurring you wish to apply to the image, the larger the condition number becomes, hence the more ill-conditioned the problem becomes. This has negative consequences with regards to the convergence of algorithms when applied to this problem. 

Due to strong convexity this problem has a unique solution, so we can use the first order optimality condition (Theorem \ref{thm:FirstOrderOpt}) in order to get the following analytic solution. Firstly, we differentiate the objective function (\ref{eq:TikhObj}) with respect to $\mathbf{u}$:
\[(\mathbf{u}-\mathbf{f}) - \lambda \nabla^{T} \nabla \mathbf{u}\]
This can be set to zero and solved for $\mathbf{u}$ in order to obtain the analytic solution, $\mathbf{u}^{\star}$
\[\mathbf{u}^{\star} = (I - \lambda \nabla^{T} \nabla)^{-1}\mathbf{f}\]
This allows us to directly evaluate the performance of iterative algorithms using $f(\mathbf{u}^{(k)}) – f^{\star}$, where $f^{\star} = f(\mathbf{u}^{\star})$. Whilst a direct analytic solution can be found to this problem, it is useful to illustrate the application of gradient descent, and how it can be accelerated. These concepts can then be applied to more complex problems. Additionally, the process of inverting a matrix can be costly, therefore it may be favourable to use an iterative algorithm even when an analytical solution is available. Fortunately, due to the structure of this matrix, it can be efficiently inverted using a technique related to the trigonometric method used to calculate its eigenvalues - further details of this can also be seen in Appendix \ref{App:FourInv}.   

\subsection{Gradient Descent}
Vanilla gradient descent is a common technique for optimising smooth objective functions. The algorithm is listed below.
\bigskip    
\begin{algorithm2e}[h!]
\SetKwInOut{Input}{Input}\SetKwInOut{Output}{Output}
\caption{Vanilla Gradient Descent}\label{alg:GD}
\Input{$N > 0$, $t \leq \frac{1}{\ell}$, $\mathbf{u}^{(0)}\in \mathbb{R}^{mn}$}
\For{$k \leftarrow 0$ \KwTo $N$}{ 
$\mathbf{u}^{(k+1)} = \mathbf{u}^{(k)} - t \nabla f(\mathbf{u}^{(k)})$
}
\Output{$\mathbf{u}^{(N+1)}$}
\end{algorithm2e}

The algorithm is initialised at some point $\mathbf{u}^{(0)} \in \mathbb{R}^{mn}$, and the gradient of the function is calculated and evaluated at $\mathbf{u}^{(0)}$. Moving in the opposite direction to the gradient vector will give the largest decrease in the function value (for some small step size), hence $\mathbf{u}^{(0)}$ is stepped in the direction of the negative gradient in order to obtain the next iterate. This process is repeated until convergence, or the maximum number of iterations has been reached. I will only consider running algorithms for a set number of iterations, however, in many cases a termination condition can be used, which breaks the algorithm when a solution with suitable accuracy is obtained. 

Choosing the step size for this algorithm is key. The gradient only applies at the point at which it was calculated, so if we move by too large a step in any direction, we cannot guarantee the function will decrease - it will in many cases diverge. The step size for any given step is influenced by the magnitude of the gradient, therefore, at steep points of the function a large step will be taken. This can be problematic if the gradient then changes quickly so that the minimum is in a completely different direction. From this, we can infer that the rate of gradient change is strongly linked to how large a step size we can take. We know in the case of a smooth function, with an $\ell$-smooth gradient, the rate of change of the gradient is bounded above by $\ell$ (\ref{def:lSmooth}). It is known that the optimal step size for gradient descent is $\frac{1}{\ell}$. 

\begin{theorem}
Suppose the function $f: \mathbb{R}^{mn} \rightarrow \mathbb{R}$ is convex and differentiable, and that its gradient is Lipschitz with constant $\ell>0$. Then, if we run gradient descent for $k$ iterations with a fixed step size $t < \frac{1}{\ell}$, it will yield a solution $f(\mathbf{u}^{(k)})$ which satisfies
\[f(\mathbf{u}^{(k)}) - f^{\star} < \frac{\|\mathbf{u}^{(0)} - \mathbf{u}^{\star}\|^{2}_{2}}{2tk} \]
\label{thm:GD_conv}
\end{theorem}
\vspace{-20pt}
Theorem \ref{thm:GD_conv} illustrates that the algorithm converges with $\mathcal{O}(1/k)$ iterations, showing that gradient descent is a relatively slow method. It should also be highlighted that this convergence theorem depends on both the initialisation of the algorithm, as well as the step size. 

\subsection{Nesterov Acceleration}
Gradient descent is intuitively simple, but a comparatively slow baseline method for smooth optimisation problems. Nesterov proposed accelerated gradient descent, which improves the speed of the above algorithm significantly. Two variations of the scheme are denoted below
\bigskip
\begin{algorithm2e}[h!]
\SetKwInOut{Input}{Input}\SetKwInOut{Output}{Output}

\Input{$N > 0$, $t \leq \frac{1}{\ell}$, $\mathbf{v}^{(0)} \in \mathbb{R}^{mn}$, $\theta^{(0)} = 1$, $q \in [0,1]$ } 
\For {$i \leftarrow 0$ \KwTo $N$}{ 
$\mathbf{u}^{(k+1)} = \mathbf{v}^{(k)} - t \nabla f(\mathbf{v}^{(k)})$\\
$\theta^{(k+1)}$ solves $[\theta^{(k+1)}]^2 = (1-\theta^{(k+1)})[\theta^{(k)}]^{2} + q\theta^{(k+1)}$\\
$\beta^{(k+1)} = \theta^{(k)}(1-\theta^{(k)})/([\theta^{(k)}]^{2} + \theta^{(k+1)})$\\
$\mathbf{v}^{(k+1)} = \mathbf{u}^{(k+1)} + \beta^{(k+1)}(\mathbf{u}^{(k+1)} - \mathbf{u}^{(k)})$\\
}
\Output{$\mathbf{u}^{(N+1)}$}
\caption{Nesterov Acceleration Scheme 1}\label{alg:Nesterov Acceleration 1}
\end{algorithm2e}
\bigskip  

\begin{algorithm2e}[h!]
\SetKwInOut{Input}{Input}\SetKwInOut{Output}{Output}
\caption{Nesterov Acceleration Scheme 2}\label{alg:Nesterov Acceleration 2}
\Input{$N > 0$, $t\leq\frac{1}{\ell}$, $\mathbf{v}^{(0)} \in \mathbb{R}^{mn}$, $\beta^{\star} = (1-\sqrt{\mu/\ell})/(1+\sqrt{\mu/\ell})$}  
\For{$k \leftarrow 0$ \KwTo $N$}{ 
$\mathbf{u}^{(k+1)} = \mathbf{v}^{(k)} - t \nabla f(\mathbf{v}^{(k)})$ \\
$\mathbf{v}^{(k+1)} = \mathbf{u}^{(k+1)} + \beta^{\star}(\mathbf{u}^{(k+1)} - \mathbf{u}^{(k)})$
}
\Output{$\mathbf{u}^{(N+1)}$}
\end{algorithm2e}

Nesterov’s schemes involve a step forward on top of the regular gradient descent mechanism. This step forward is decided by the current velocity - the difference between the current iterate and the previous iterate, as well a momentum parameter, $\beta$. Such accelerated schemes obtain optimal results for first order optimisation algorithms applied to smooth convex problems.  This works on the assumption that the gradient at two consecutive positions will be similar. Both of the above acceleration schemes implement the same idea, but with a slightly different procedure and  slightly different hyperparameters. In Algorithm \ref{alg:Nesterov Acceleration 1}, we choose $q$ which dictates how $\beta^{(k)}$, the variable momentum parameter, changes as the algorithm progresses. However, in Algorithm \ref{alg:Nesterov Acceleration 2}, a constant optimal momentum parameter, $\beta^{\star}$, is selected. 

Some interesting analysis can be carried out on the values of $\beta$ in Algorithm \ref{alg:Nesterov Acceleration 1} and Algorithm \ref{alg:Nesterov Acceleration 2} to give an insight into how the parameters in each of them behave. The relationship can be seen in Figure \ref{fig:BetaComp}. Algorithm \ref{alg:Nesterov Acceleration 1} leads to $\beta^{(k)}$ converging to the optimal parameter $\beta^{\star}$ selected in Algorithm \ref{alg:Nesterov Acceleration 2}. The value of $q$ selected dictates the value which $\beta^{(k)}$ converges to. 

The value of $\beta$ dictates the behaviour of these acceleration algorithms. If $\beta>\beta^{\star}$, then Nesterov ripples are observed; this leads to sub-optimal convergence of the algorithm. If $\beta<\beta^{\star}$, then no oscillations are observed in the convergence. However, a sub-optimal amount of momentum is applied, leading to sub-optimal convergence. The value $\beta^{\star}$ is clearly the optimal momentum value - choosing optimal $q$, $q^{\star} = \frac{1}{\kappa}$, leads to $\beta^{(k)}$ converging to $\beta^{\star}$. When $q$ is set such that it is above the optimal value, then $\beta^{(k)}$ converges to a value smaller than $\beta^{\star}$, and when set such that it is below the optimal value, $\beta^{(k)}$ converges to a value above $\beta^{\star}$. At the extreme of $q=1$, gradient descent is obtained. 

\begin{theorem}
Suppose the function $f: \mathbb{R}^{mn} \rightarrow \mathbb{R}$ is strongly convex, with known strong convexity parameter, $\mu$, and is differentiable, with $\ell$-smooth gradient (therefore $\kappa = \frac{\ell}{\mu}$). Then, if we run Nesterov's acceleration scheme (Algorithm \ref{alg:Nesterov Acceleration 2}) with $\beta^{\star} = (1-\sqrt{1/\kappa})/(1+\sqrt{1/\kappa})$ for $k$ iterations with a fixed step size $t < \frac{1}{\ell}$, it will yield a solution $f(\mathbf{u}^{(k)})$ which satisfies
\[f(\mathbf{u}^{(k)}) - f^{\star} < \ell \left(1-\sqrt{\frac{1}{\kappa}}\right)^k \|\mathbf{u}^{0} - \mathbf{u}^{*}\|^{2}\]
\label{thm:GDAccConv}
\end{theorem}
\vspace{-10pt}
Such a convergence rate is known as linear convergence, due to linearity when plotted on a logarithmic scale. This method will clearly converge far quicker than the vanilla gradient descent method mentioned previously. Such accelerated gradient schemes depend on the condition number of the objective function, $\kappa$, the larger $\kappa$ the slower the convergence. 

\subsection{Restart}
In both acceleration schemes the strong convexity parameter is required in order to apply acceleration optimally, however, it is not always the case that this value is known. The strong convexity parameter may be very difficult to work out, in which case we would not know how to optimally choose the $\beta^{\star}$ or $q^{\star}$ parameters. Restart is a method of smoothing the Nesterov ripples in order to achieve quicker convergence. This essentially allows us to estimate the strong convexity parameter. 

Restart methods involve restarting the algorithm from the current iterate when a certain condition is met. To implement adaptive restart techniques, use Algorithm \ref{alg:Nesterov Acceleration 1} with $q = 0$. Note that we aren't required to know the strong convexity parameter in order to do so. One characteristic of this algorithm is that, in each loop, $\beta^{(k)}$ will proceed to increase above the optimum value $\beta^{\star}$. When this begins to have a detrimental effect on the convergence of the algorithm, we reset $\theta^{(k+1)} = 1$ (which is equivalent to restarting the algorithm at the current iterate). Two simple restart schemes can be used to identify when $\beta^{(k)}$ is having a detrimental effect on the convergence rate of the algorithm: 

\begin{itemize}
    \item Objective Scheme:
    \begin{center}
        $f(\mathbf{u}^{(k)})>f(\mathbf{u}^{(k+1)})$
    \end{center}
    \item Gradient Scheme:
    \begin{center}
        $\nabla f(\mathbf{v}^{(k+1)})^{T}(\mathbf{u}^{(k)} - \mathbf{u}^{(k-1)})>0$
    \end{center}
\end{itemize}

For the objective scheme, whenever the function is beginning to increase, the algorithm is overshooting and a Nesterov ripple is about to occur. Hence, the acceleration process is restarted.  However, it has the disadvantage that the function must be evaluated at every iterate which can potentially be costly. Therefore, the second method is often preferred, as all terms which appear in the rule require no additional computation. The intuition behind this rule is that the algorithm is restarted if the gradient and the velocity form an obtuse angle, i.e. the velocity has a direction very different to the current direction of greatest decrease. Both of these restart schemes offer comparable results, so for the reason mentioned above we will prefer the gradient restart scheme.

\vspace{10pt}
\begin{algorithm2e}[h!]
\caption{Nesterov Acceleration with Restart}\label{alg:Restart}
\SetKwInOut{Input}{Input}\SetKwInOut{Output}{Output}
\Input{$N > 0$, $t \leq \frac{1}{\ell}$, $\mathbf{v}^{(0)} \in \mathbb{R}^{mn}$, $\theta^{(0)} = 1$}
\For {$k \leftarrow 0$ \KwTo $N$} { 
    $\mathbf{u}^{(k+1)} = \mathbf{v}^{(k)} - t \nabla f(\mathbf{v}^{(k)})$ \\
    $\theta^{(k+1)} = 1 + \sqrt{(1+4[\theta^{(k)}]^{2})}/2$\\
    $\beta^{(k+1)} = (\theta^{(k)}-1)/\theta^{(k+1)}$\\
    $\mathbf{v}^{(k+1)} = \mathbf{u}^{(k+1)} + \beta^{(k+1)}(\mathbf{u}^{(k+1)} - \mathbf{u}^{(k)})$\\
    \If{$ \nabla f(\mathbf{v}^{(k+1)})^{T}(\mathbf{u}^{(k)} - \mathbf{u}^{(k-1)}) > 0$}{
        $\theta^{(k+1)} = 1$
    }
}
\Output{$\mathbf{u}^{(N+1)}$}
\end{algorithm2e}
\vspace{7pt}

\begin{figure}[h!]
     \centering
     \includegraphics[width = 10cm]{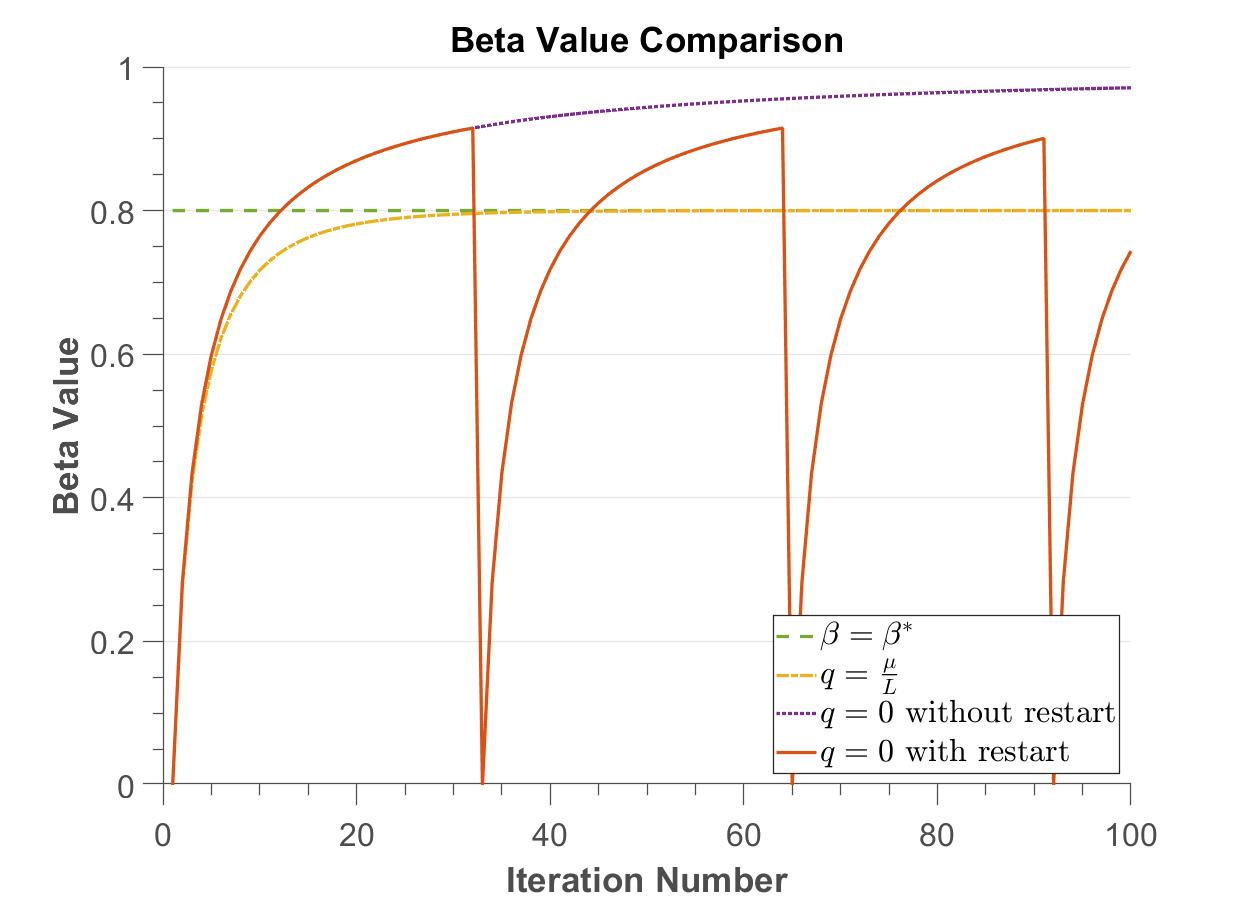}
     \vspace{-5pt}
     \caption{This graph illustrates how $\beta$ varies in the 3 proposed acceleration schemes.}
     \label{fig:BetaComp}
\end{figure}

An alternative way of viewing the restart methods is as a way of estimating the strong convexity parameter. The plots in Figure \ref{fig:BetaComp} allow us to see the effect restart has on $\beta$ and how this corresponds to the optimal $\beta^{\star}$ and $q^{\star}$ values. Restart allows the $\beta^{(k)}$ value to be increased past the optimal value, however, when this begins to have a detrimental effect on the convergence of the algorithm it is reset back to $\beta^{(k)} = 0$. When $q=0$ without restart, $\beta^{(k)}$ stays above the optimal value of $\beta^{\star}$, and this is why we observe Nesterov ripples in the convergence of this algorithm.  

\vspace{10pt}
\vspace{10pt}

Since the value of $\beta$ still exceeds the optimum value, some smoothed Nesterov ripples will be expected, however they are significantly reduced by the restart rule.  
\newpage
\subsection{Numerical Results}
 
\begin{figure}[h!]
    \centering
    \includegraphics[width = 0.75\textwidth]{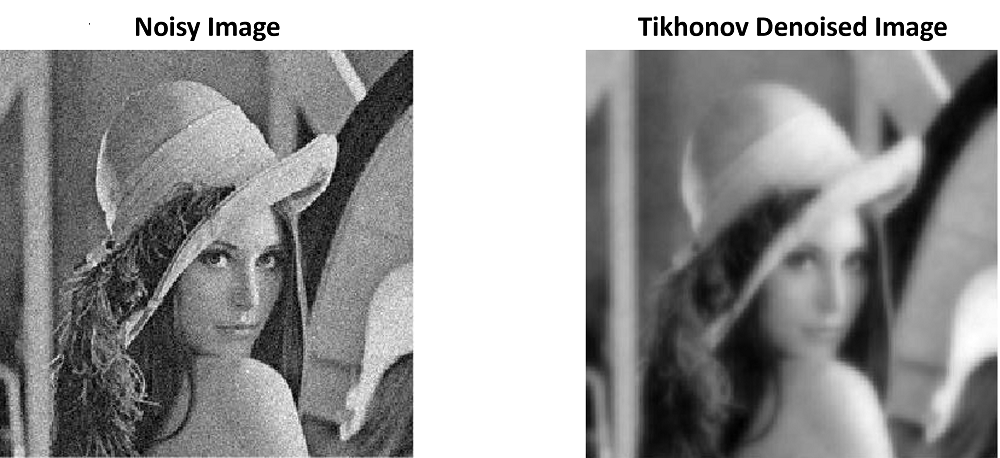}
    \vspace{-5pt}
    \caption{Lena image with Tikhonov denoising applied}
    \label{fig:TikhLena}
\end{figure}

The results in Figure \ref{fig:TikhLena} were obtained with $\lambda = 10$. The denoised image in this case is visibly smooth, and the Gaussian noise from the original image has been removed. It is clear that this image still corresponds to the original image, however, the edges from the original image appear blurred rather than the sharp edges we desire. This can be a problem in many computer vision tasks where edge detection may be necessary, such as in image segmentation. To highlight the poor edge retention of Tikhonov denoising, I have also explored the impact this has on a more geometric example.

\begin{figure}[h!]
    \centering
    \includegraphics[width = \textwidth]{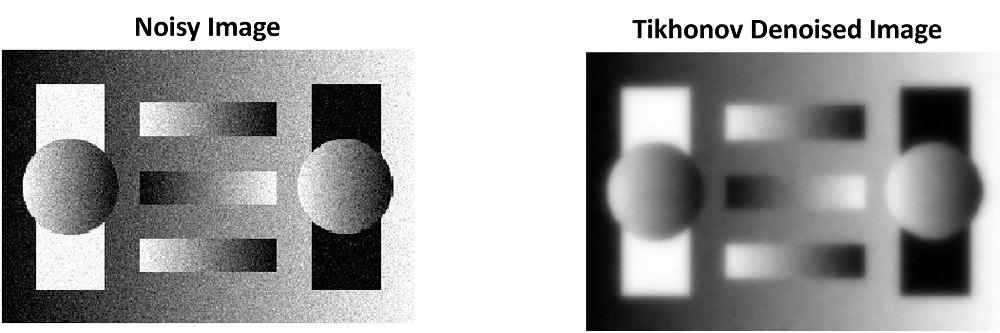}
    \vspace{-20pt}
    \caption{Geometric image with Tikhonov denoising applied}
    \label{fig:TikhGeom}
\end{figure}
\vspace{10pt}

From Figure \ref{fig:TikhGeom}, it is once again clear that the edges in the denoised image are blurred. However, it is also clear that the smooth gradients that exist in the image are retained with no artefacts, such as the intensity gradient from black to white in the background of the image. These properties can also be illustrated using a cross section of the denoised image. 

\begin{figure}[h!]
    \centering
    \includegraphics[width = \textwidth]{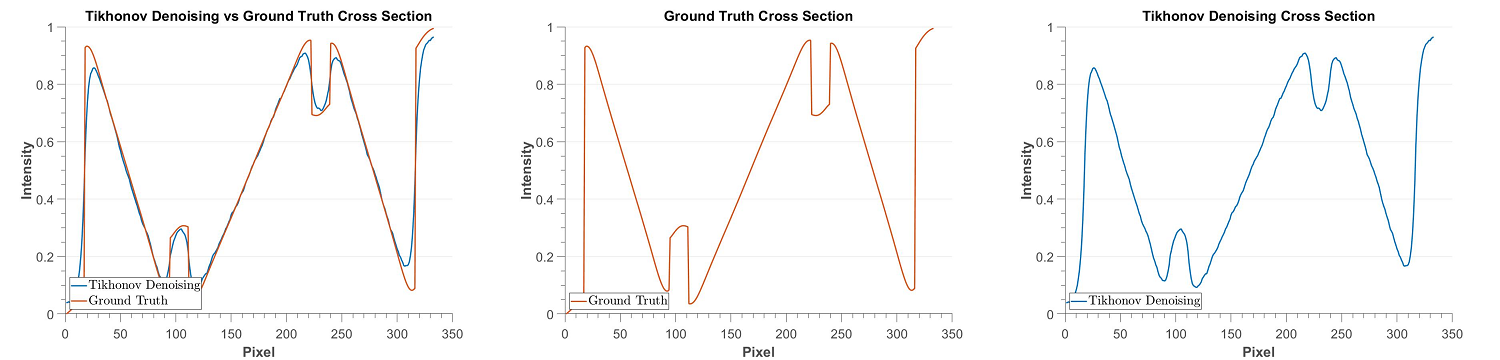}
    \vspace{-20pt}
    \caption{Tikhonov denoising cross section}
    \label{fig:TikhCross}
\end{figure}

Figure \ref{fig:TikhCross} highlights the properties of Tikhonov denoising by comparing it to the cross section of the ground truth image. The edges in the ground truth image are the sharp discontinuities - it can be observed that Tikhonov regularisation doesn't allow for these sharp discontinuities in the solution. Instead, the solution has a smooth transition from low to high intensity, therefore explaining the blurred appearance of the edges. However, when a smooth gradient is present in the image, for example, between pixels 125 and 225, the Tikhonov denoising solution performs well, capturing the smooth transition of the image.  

\begin{figure}[h]
    \centering
    \includegraphics[width = \textwidth]{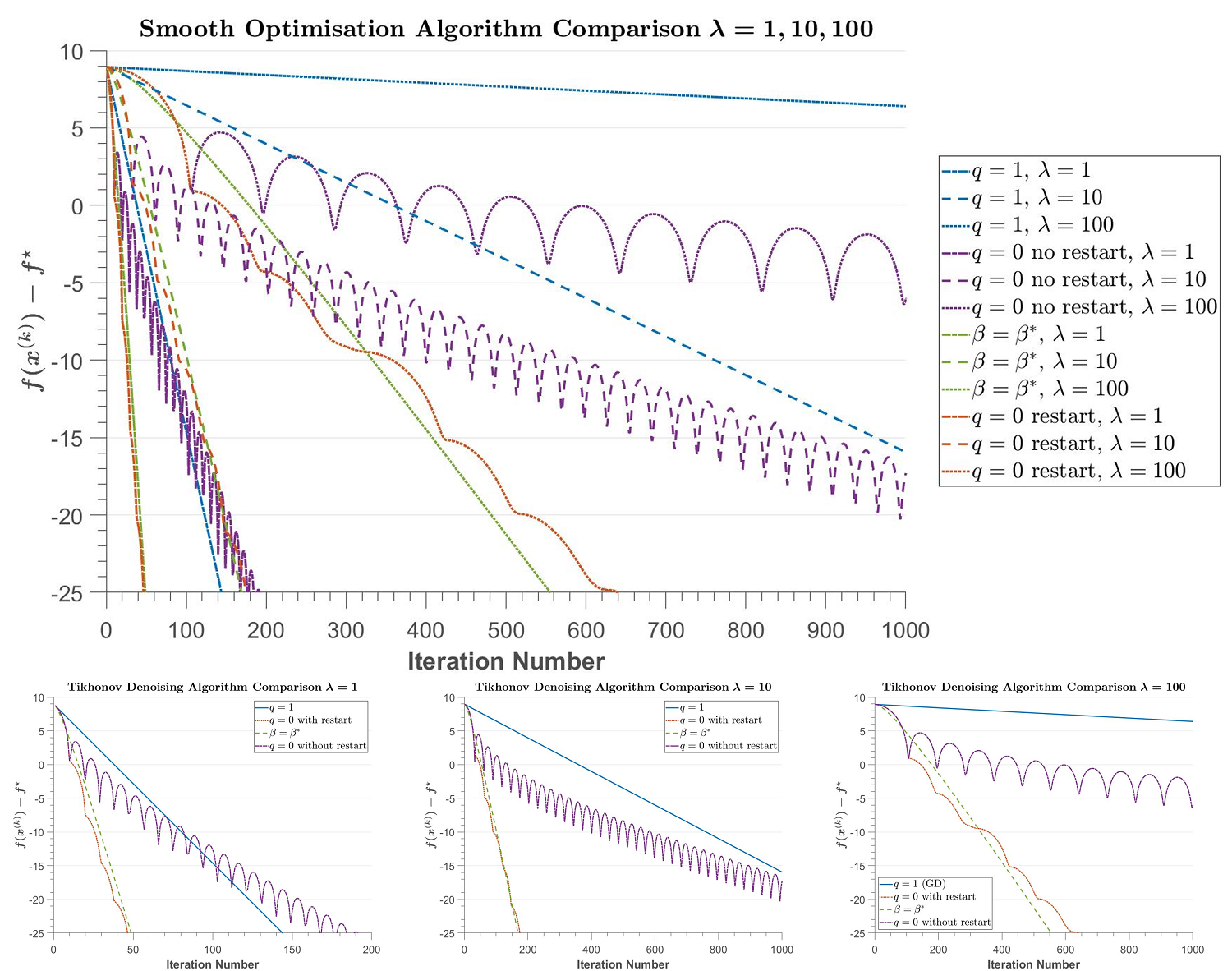}
    \vspace{-20pt}
    \caption{Tikhonov denoising algorithm comparison}
    \label{fig:TikhAlgComp}
\end{figure}

A graph illustrating how the different algorithms performed, with regards to convergence rate, on this problem can be seen in Figure \ref{fig:TikhAlgComp}. I have not used every algorithm above; similarly, performing schemes have been omitted. Both schemes which use the optimal parameter values, i.e. $q=q^{\star}$ and $\beta = \beta^{\star}$ perform comparably. The only difference is that less acceleration is applied to $q=q^{*}$ during the initial iterations - hence, this method is slightly slower. Both adaptive restart schemes perform similarly, and so the function scheme has been omitted.

\vspace{10pt}
\vspace{10pt}

From these results, it is clear that applying acceleration yields a significant improvement in the convergence rate of smooth gradient descent. The detrimental effect Nesterov ripples have on accelerated methods also becomes clear. The overshooting effect clearly leads to inefficiencies in the optimisation process, to the point where the initial advantage that the accelerated method has over the unaccelerated gradient descent is diluted. Another key point is to observe how similarly the accelerated method using restart performs to the accelerated method using optimal parameters. This shows that restart allows Nesterov acceleration to be applied to optimisation problems without calculating the strong convexity parameter, and provides an opportunity to accelerate many problems which previously appeared to be inaccessible. It is also clear that the method converges more slowly for larger values of $\lambda$ as the condition number increases, as expected.
 
After 1000 iterations, all four methods found results which are visibly very similar. However, it would be expected that the two most efficient algorithms would reach a visibly acceptable result far quicker than the other two methods - clearly within 200 iterations. With this in mind, Table \ref{tab:TikhConv} denotes the time in seconds it takes for each of the algorithms to achieve a predetermined, appropriate degree of accuracy, $1e^{-3}$. The image in Figure \ref{fig:TikhLena} shows the an image with this proximity to the true solution. 

\vspace{10pt}
\begin{table}[H]
     \centering
     \resizebox{\columnwidth}{!}{
     \begin{tabular}{|c||c|c|c|c|}
      \hline
      Algorithm & Gradient Descent & q = 0 without restart & q = 0 with restart & $\beta = \beta^{\star}$  \\ \hline
      Iterations & 452 & 90 & 59 & 66
      \\ \hline
      Time (Seconds) & 17.21 & 3.40 & 2.27 & 2.54 \\ \hline
     \end{tabular}}
     \vspace{-7pt}
     \caption{Convergence table for different algorithms}
     \label{tab:TikhConv}
\end{table}

These results are mostly to be expected, illustrating how cheap the restart decision in each iteration is, so much so that the restart method actually achieves the accuracy in fewer iterations than the optimum parameter. This further emphasises how well this method performs, and allows us to conclude this is an effective method to be used on more complex problems than Tikhonov denoising. 
\clearpage

\section{Non-Smooth Optimisation - Total Variation (TV) Denoising } \label{sec:TV}
\subsection{Introduction to TV Denoising}
Motivated by the shortfalls of Tikhonov denoising, we will now consider a partially non-smooth optimisation problem. As observed in the Figures \ref{fig:TikhLena} and \ref{fig:TikhGeom} in Section \ref{sec:Tikh}, Tikhonov regularisation doesn't allow the solution to contain any discontinuities and therefore sharp edges. We instead consider a more advanced regularisation term, Total Variation (TV) regularisation,  to overcome this shortfall. When applying this regularisation the problem of denoising then becomes
\[
\argminB_{\mathbf{u}} \frac{1}{2}\|\mathbf{u}-\mathbf{f}\|_{2}^{2}  + \lambda \rm{TV}(\mathbf{u})
\]
The TV regularisation term can come in two forms: isotropic and anisotropic, both of which are denoted below as follows
 
\begin{itemize}
    \item Isotropic:
        \[{\rm{TV}}(\mathbf{u}) =\sum_{j=1}^{mn}\sqrt{\bigg(\sum_{i=1}^{k}|\nabla u_{i,j}|^2\bigg)}\] 
    
    \item Anisotropic:
        \[{\rm{TV}}(\mathbf{u}) =\sum_{j=1}^{mn} \sum_{i=1}^{k} |\nabla u_{i,j}|\]
\end{itemize}

Where $u_{i,j}$ is the $j^{th}$ pixel of the $i^{th}$ channel. Since the values considered are scalars, $|\cdot|$ is simply the absolute value.
As indicated by the name, isotropic and anisotropic regularisation vary in their relationship with direction. The value of isotropic regularisation reacts the same to any direction, whereas anisotropic is direction dependent. The problem dictates which of these terms will lead to better results. Both isotropic and anisotropic TV terms can easily be optimised using the methods proposed later in this chapter. The only way in which the methods change is in the formulation of the dual function, which will also be discussed later in this chapter.

Whilst the data term in this case remains the same, the regularisation changes significantly from Tikhonov regularisation. TV regularisation allows for discontinuities in the solution, but the objective function is no longer smooth. In fact, TV denoising favours a sparse solution, therefore enforcing sparsity on the gradient of the image. This leads to many pixels for which there is no change between, and an image consisting of blocks of colour. In mathematical terms, applying TV regularisation promotes piecewise constant solutions. 
 
Since this objective function is not smooth, there is no way of obtaining a closed form solution. This highlights the neccessity for iterative algorithms to solve these problems, as well as enforcing the need for a different measure of convergence. However the data term in strongly convex with respect to $\mathbf{u}$, which is a property we can take advantage of when optimising this objective function. 
 
\subsection{Proximal Gradient Descent}
Proximal gradient descent is a method of optimising certain non-smooth objective functions. These partially non-smooth functions must be of the following form
\begin{equation}\label{eq:compFunc}
g(\mathbf{u}) + h(\mathbf{u})
\end{equation} 
 \begin{itemize}
     \item $g(\cdot)$ is convex and differentiable, with $\ell$-smooth gradient 
     \item $h(\cdot)$ is convex, not neccessarily differentiable 
 \end{itemize}
 
\vspace{10pt}
\begin{algorithm2e}[h!]
\SetKwInOut{Input}{Input}\SetKwInOut{Output}{Output}
\caption{Proximal Gradient Algorithm}\label{alg:prox_grad_alg}
\Input{$N > 0$, $t \leq \frac{1}{\ell}$, $\mathbf{u}^{(0)}\in \mathbb{R}^{mn}$}  
\For {$k \leftarrow 0$ \KwTo $N$}{ 
$\mathbf{u}^{(k+1)} = \mathbf{prox}_{h(\cdot),t}(\mathbf{u}^{(k)} - t\nabla g(\mathbf{u}^{(k)})))$
}
\Output{$\mathbf{u}^{(N+1)}$}
\end{algorithm2e}
\vspace{10pt}

 The proximal gradient algorithm is denoted in Algorithm \ref{alg:prox_grad_alg}. It can be observed that this is very similar to gradient descent with the additional step of applying the proximal operator to the gradient step. The proximal operator of a function, $h(\cdot)$, is denoted below
\[\mathbf{prox}_{h(\cdot),t}(\mathbf{x}) = \argminB_{\mathbf{z}}\frac{1}{2t}\|\mathbf{x}-\mathbf{z}\|^{2}_{2} + h(\mathbf{z})\]
At first glance, we may just appear to have converted the original optimisation problem to another. However, the proximal operator of a function will often have favourable properties, such as having a closed form solution or being able to be solved efficiently using an iterative algorithm. It should be noted that the step size $t$ remains the same in the gradient update and the proximal update. If the smooth function, $g$, is $\ell$-smooth, then the optimal step size will be $\frac{1}{\ell}$  . 

The proximal operator aims to find a solution which minimises the function it is applied to, whilst staying close to the input value. In the case of proximal gradient descent, this makes sense as the solution to this proximal step will be a compromise between the gradient step forward for the smooth function $g$ and the minimum of the non-smooth function $h$. Another important property of the proximal operator is that adding the squared $L_{2}$ norm to the non-smooth function, leads to the proximal operator being a strongly convex function; hence, a closed form solution can often be found. Since parallels can be drawn with gradient descent, it may be expected that the convergence rate of proximal gradient descent will be similar to that of gradient descent - this is in fact the case.
\begin{theorem}\label{thm:GDConv}
Suppose the function $f:\mathbb{R}^{mn} \mapsto \mathbb{R}$ is convex, differentiable, and it's gradient has a Lipschitz constant $\ell>0$. Then if we run gradient descent for $k$ iterations with a fixed step size $t \leq \frac{1}{\ell}$ it will yield a solution $f(\mathbf{u}^{(k)})$ which satisfies:
 \[f(\mathbf{u}^{(k)}) - f^{\star} \leq \frac{\|\mathbf{u}^{(0)} - \mathbf{u}^{\star}\|^{2}_{2}}{2tk}\]
\end{theorem}
The proximal gradient algorithm therefore has convergence rate $\mathcal{O}(\frac{1}{k})$. In this case, $k$ is the number of iterations as seen in Algorithm \ref{alg:prox_grad_alg}, this does not include any iterations taken to solve the proximal operator. Therefore, if the proximal operator is expensive then the proximal gradient algorithm will also be expensive, and so it is crucial that the proximal operator has either a closed form solution or a cheap iterative solution. This is a relatively slow convergence rate, and raises the question of whether the algorithm can be accelerated.

\subsection{The Dual Objective Function and Duality Gap}
In the case of TV denoising the proximal gradient algorithm cannot be applied directly to the primal objective function, since we would obtain a proximal operator which is identical to the original objective function. However, we can formulate the dual objective of this problem, then perform the dual proximal algorithm to find the maximum value of this function. 
\begin{equation}\label{eq:TVObj}
\argminB_{\mathbf{u}}\frac{1}{2}\|\textbf{u} - \textbf{f}\|^{2}_{2} + \lambda \rm{TV}(\mathbf{u})
\end{equation}
In order to derive the dual function for this primal objective function, we need to consider the dual of the TV regularisation term
\begin{equation} \label{eq:TVDual} 
{\rm{TV}}(\mathbf{u}) = \argminB_{\mathbf{u}}\argmaxB_{\mathbf{p}} \left\{\sum_{i=1}^{mn}\langle\nabla u_{i},p_{i}\rangle; |p_{i}|\leq 1\;{\rm{ for }}\;j = 1,...,mn\right\}
\end{equation}
For the purpose of maintaining a simple notation, we will generalise the dot product so that it can be applied to vectors in the image space. This is therefore a constrained maximisation problem. By making this constraint explicit, we can write this as an unconstrained optimisation problem, of the following form
\[{\rm{TV}}(\mathbf{u}) = \argminB_{\mathbf{u}}\argmaxB_{\mathbf{p}}{\langle \nabla \mathbf{u}, \mathbf{p} \rangle} + \delta(\mathbf{p})\]
Firstly, the TV regulariser can be substituted for its dual form introduced in (\ref{eq:TVDual}). This introduces a dual variable $\mathbf{p}$ with a constraint $\delta(\mathbf{p})$ placed upon it. This constraint can be included explicitly in the primal dual function. 
\begin{equation}\label{eq:TVPD}
\argminB_{\mathbf{u}} \argmaxB_{\mathbf{p}} \frac{1}{2}\|\mathbf{u} - \mathbf{f}\|^{2}_{2} + \lambda\langle\nabla \mathbf{u},\mathbf{p} \rangle + \delta(\mathbf{p})
\end{equation}
This new problem is smooth with respect to $\mathbf{u}$ - therefore gradient descent can be applied to iterate $\mathbf{u}$. Firstly, we find the gradient of the objective function, \ref{eq:TVObj}, with respect to $\mathbf{u}$
\[\mathbf{u} - \mathbf{f} - \lambda\nabla\cdot \mathbf{p}\]
        
From the gradient it can be determined that the Hessian matrix for this function with respect to $\mathbf{u}$ is the identity matrix. Since the eigenvalues of the identity matrix are all 1, the optimal step size for this gradient step will be 1. Fortunately, this means that we have a closed form solution for $\mathbf{u}$ in terms of $\mathbf{p}$ - using the optimal gradient descent step
\begin{align*}
    \mathbf{u}^{(k+1)} &= \mathbf{u}^{(k)} - (\mathbf{u}^{(k)} -\mathbf{f} - \lambda\nabla\cdot \mathbf{p}) \\
    &=\mathbf{f}+\lambda\nabla\cdot \mathbf{p}
\end{align*}
This value can then be plugged back into the primal dual equation (Equation \ref{eq:TVPD}) to obtain the dual function of TV denoising:
\begin{align*}
&\argminB_{\textbf{u}} \argmaxB_{\textbf{p}} \frac{1}{2}\|\textbf{u} - \textbf{f}\|^{2}_{2} + \lambda\langle\nabla \textbf{u},\textbf{p} \rangle + \delta(\textbf{p})\\
&=\argmaxB_{\textbf{p}} \frac{1}{2}\|\textbf{f}+\lambda\nabla\cdot \textbf{p} - \textbf{\textbf{f}}\|^{2}_{2} - \lambda\langle \textbf{f}+\lambda\nabla\cdot \textbf{p},\nabla \cdot \textbf{p} \rangle + \delta(\textbf{p})\\
&=\argmaxB_{\textbf{p}} \frac{\lambda \zwe{2}}{2} \: \|\nabla\cdot \textbf{p}\|^{2}_{2} - \lambda\langle \textbf{f},\nabla \cdot \textbf{p} \rangle - \lambda^{2}\langle\nabla\cdot \textbf{p},\nabla \cdot \textbf{p} \rangle + \delta(\textbf{p})\\
&=\argmaxB_{\textbf{p}} -\frac{\lambda \zwe{2}}{2} \: \|\nabla\cdot \textbf{p}\|^{2}_{2} - \lambda\langle \textbf{f},\nabla \cdot \textbf{p} \rangle + \delta(\textbf{p})\\
\end{align*}
Now that the dual problem has been derived, it can be observed that this function is also made up of a smooth component (the first 2 terms) and a non-smooth component (the final term). Therefore, we can directly apply the proximal gradient algorithm to this problem, assuming that the proximal operator of $\mathbf{\delta(p)}$ can be easily evaluated. Fortunately, this is the case - the proximal operator of this function is simply the projection of the input into the unit ball as seen in Equation (\ref{eq:ProxDelta}) (for a more detailed discussion of this fact see Appendix \ref{app:ProxDelt}). 
\vspace{-5pt}
\begin{equation} \label{eq:ProxDelta}
    \left[\mathbf{prox}_{\delta(\cdot),t}(\mathbf{p})\right]_j = \frac{p_{j}}{|p_{j}|}
\end{equation}
for $j = 1,2,...,mn$ for $\mathbf{p} \in \mathbb{R}^{mn}$. From now on we will denote the $\mathbf{prox}_{\delta(\cdot),t}(\cdot)$ as $\mathbf{proj}(\cdot)$.
It should be highlighted that the dual form of the TV regularisation term \ref{eq:TVDual}, the case considered here is that of the isotropic TV term. We can now formulate the update formula for Algorithm \ref{alg:prox_grad_alg} when applied to dual function of TV denoising: 
\begin{equation}
    \mathbf{p}^{(k+1)} = \mathbf{proj}(\mathbf{p}^{(k)} + t(\lambda^{2}\nabla \nabla\cdot \mathbf{p}^{(k)} + \lambda \nabla \mathbf{f}))
\end{equation}
Since the dual objective function must be maximised, we step in the direction of the positive gradient. Finally we must derive the $\ell$-smooth parameter for the smooth component of this function in order to calculate the optimal step size, t. The smooth component of the dual objective function is as follows
\[-\frac{\lambda^{2}}{2}\|\nabla\cdot \mathbf{p}\|^{2}_{2} - \lambda\langle \mathbf{f},\nabla \cdot \mathbf{p} \rangle\]
In order to make this easier to differentiate we can use the identity \ref{eq:AdjointIdentity}:
\[-\frac{\lambda^{2}}{2}\|\nabla\cdot \mathbf{p}\|^{2}_{2} + \lambda\langle \nabla \mathbf{f},\mathbf{p} \rangle\]
We can then find the gradient of this function with respect to $\mathbf{p}$:
\[\lambda^{2}\nabla \nabla\cdot \mathbf{p} + \lambda \nabla \mathbf{f}\]
Therefore we can easily see that the Hessian matrix of this function is
\[\lambda^{2}\nabla \nabla^{T}\]

By using the trigonometric analysis applied in Appendix \ref{App:FourInv}, we can see that the minimum eigenvalue of this matrix is $-8\lambda^{2}$ and the maximum is $0$. In this case all of the eigenvalues are negative due to the fact that we are now dealing with a concave maximisation problem. However, the same theory is applied to this problem as to convex minimisation problems, as we can minimise this function multiplied by -1. Hence the step size used in the proximal gradient algorithm will therefore be $\frac{1}{8\lambda^{2}}$. 

Now that the primal and dual objective functions have been calculated, we can consider the notion of the duality gap - the difference between the value of the primal function and the dual function. In this case, we have strong duality since the dual form of the TV regularisation is equal to the primal form of the regularisation itself. Therefore the optimal values of the primal and dual functions will both have the same value. We can use this as a measure of accuracy, since we know that the duality gap will converge to zero as both the primal and dual functions tend to their optimal values.

\subsection{Accelerating Proximal Gradient Descent}
Due to the iterative nature of this method, we can consider applying the same acceleration techniques which were applied to the smooth problem. The Accelerated Proximal Gradient Algorithm (APGA) is described in Algorithm \ref{alg:AccProxGrad}.

\begin{algorithm2e}[H]
\SetKwInOut{Input}{Input}\SetKwInOut{Output}{Output}
\caption{Accelerated Proximal Gradient Algorithm} \label{alg:AccProxGrad}
\Input{$N > 0$, $t \leq \frac{1}{\ell}$, $\mathbf{v}^{(0)}\in \mathbb{R}^{mn}$,$q \in [0,1]$} 
\For {$k \leftarrow 0$ \KwTo $N$}
{ 
    $\mathbf{u}^{(k+1)} = \mathbf{prox}_{h,t}(\mathbf{v}^{(k)} - t\nabla f(\mathbf{v}^{(k)})))$\\
    $\theta^{(k+1)}$ solves $[\theta^{(k+1)}]^{2} = (1 - \theta^{(k+1)})[\theta^{(k)}]^{2} + q\theta^{(k+1)}$\\
    $\beta^{(k+1)} = \theta^{(k)}(1-\theta^{(k)})/([\theta^{(k)}]^{2} + \theta^{(k+1)})$\\
    $\mathbf{v}^{(k+1)} = \mathbf{u}^{(k+1)} + \beta^{(k+1)}(\mathbf{u}^{(k+1)} - \mathbf{u}^{(k)})$
}
\Output{$\mathbf{u}^{N+1}$}
\end{algorithm2e}

As is the case for smooth optimisation, this improves the convergence rate of the function significantly, providing the result in Theorem (\ref{thm:AccProxGradConv}).
\begin{theorem} \label{thm:AccProxGradConv}
Suppose the function $f:\mathbb{R}^{mn} \mapsto \mathbb{R}$ is of the same composite form as described in \ref{eq:compFunc}, where it's smooth component has Lipschitz constant $\ell>0$. Then running the accelerated proximal gradient algorithm for k iterations with a fixed step size $t\leq\frac{1}{\ell}$ will yield a solution $f(\mathbf{u}^{(k)})$ which satisfies:  
\[f(\mathbf{u}^{(k)}) - f^{\star} \leq \frac{2\|\mathbf{u}^{(0)} - \mathbf{u}^{\star}\|^{2}_{2}}{t(k+1)^{2}}\]
\end{theorem}

This illustrates how the convergence rate of the accelerated proximal gradient method has been improved to $\mathcal{O}(\frac{1}{k^{2}})$. However, due to the nature of acceleration, we will once again expect to observe inefficiencies when the momentum applied becomes too great. Having witnessed the undesirable effects of excess momentum in the domain of smooth optimisation, we propose that applying restart to the accelerated proximal gradient algorithm would improve the algorithm. To add acceleration to this method, we must use the gradient update scheme specified in Section \ref{sec:Tikh}. Due to the proximal step, the following generalised restart decision is used:
\[(\mathbf{v}^{(k)}-\mathbf{u}^{(k+1)}) \cdot (\mathbf{u}^{(k)} - \mathbf{u}^{(k-1)}) > 0 \]
Since this method is often applied to the dual objective function of problems, we will refer to it as the Accelerated Dual Proximal Algorithm (ADPA), however in the cases when it is applied to primal objective functions, it will be referred to as the Accelerated Proximal Gradient Algorithm (APGA). The final ADPA algorithm which is used to optimise the TSV denoising problem can be observed in Algorithm \ref{alg:TVADPA}.

\begin{algorithm2e}[h!]
\SetKwInOut{Input}{Input}\SetKwInOut{Output}{Output}
\caption{Accelerated Proximal Gradient Algorithm applied to TV Denoising} \label{alg:TVADPA}
\Input{$N > 0$, $t \leq \frac{1}{\ell}$, $\mathbf{\hat{p}}^{(0)}\in \mathbb{R}^{mn}$, $q \in [0,1]$} 
\For {$k \leftarrow 0$ \KwTo $N$}
{ 
    $\mathbf{p}^{(k+1)} =  \mathbf{\hat{p}}^{(k+1)} = \mathbf{proj}(\mathbf{\hat{p}}^{(k)} + t(\lambda^{2}\nabla \nabla\cdot \mathbf{\hat{p}}^{(k)} + \lambda \nabla \mathbf{f}))$\\
    $\theta^{(k+1)}$ solves $[\theta^{(k+1)}]^{2} = (1 - \theta^{(k+1)})[\theta^{(k)}]^{2} + q\theta^{(k+1)}$\\
    $\beta^{(k+1)} = \theta^{(k)}(1-\theta^{(k)})/([\theta^{(k)}]^{2} + \theta^{(k+1)})$\\
    $\mathbf{\hat{p}}^{(k+1)} = \mathbf{p}^{(k+1)} + \beta^{(k+1)}(\mathbf{p}^{(k+1)} - \mathbf{p}^{(k)})$ \\
    \If{$(\mathbf{\hat{p}}^{(k)}-\mathbf{p}^{(k+1)}) \cdot (\mathbf{p}^{(k)} - \mathbf{p}^{(k-1)}) > 0$}{
        $\theta^{(k+1)} = 1$
    }
}
$\mathbf{u} = \mathbf{f}+\lambda\nabla\cdot \mathbf{p}^{(N+1)}$ \\
\Output{$\mathbf{u}$}
\end{algorithm2e}
\newpage
\subsection{Numerical Results}
\vspace{10pt}
\begin{figure}[h!]
    \centering
    \includegraphics[width = 0.75\textwidth]{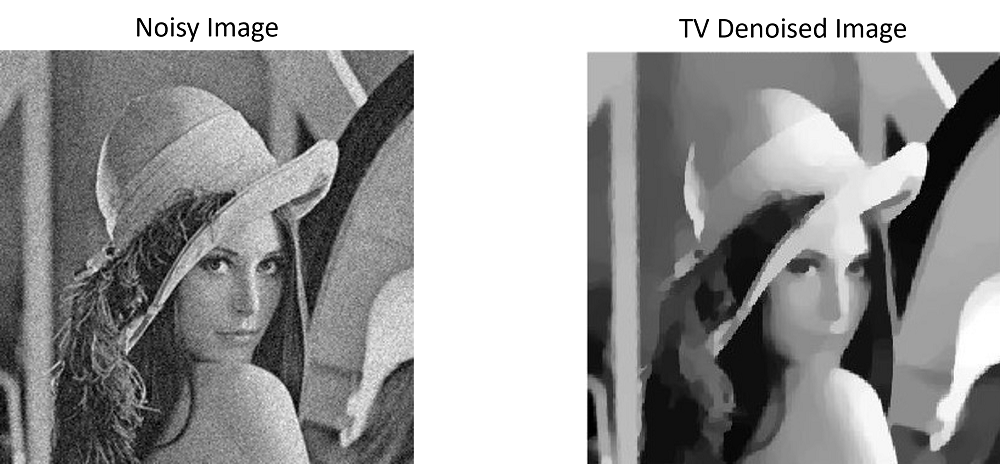}
    \vspace{-5pt}
    \caption{Lena image with TV denoising applied}
    \label{fig:TVLena}
\end{figure}
\begin{figure}[h!]
    \centering
    \includegraphics[width = \textwidth]{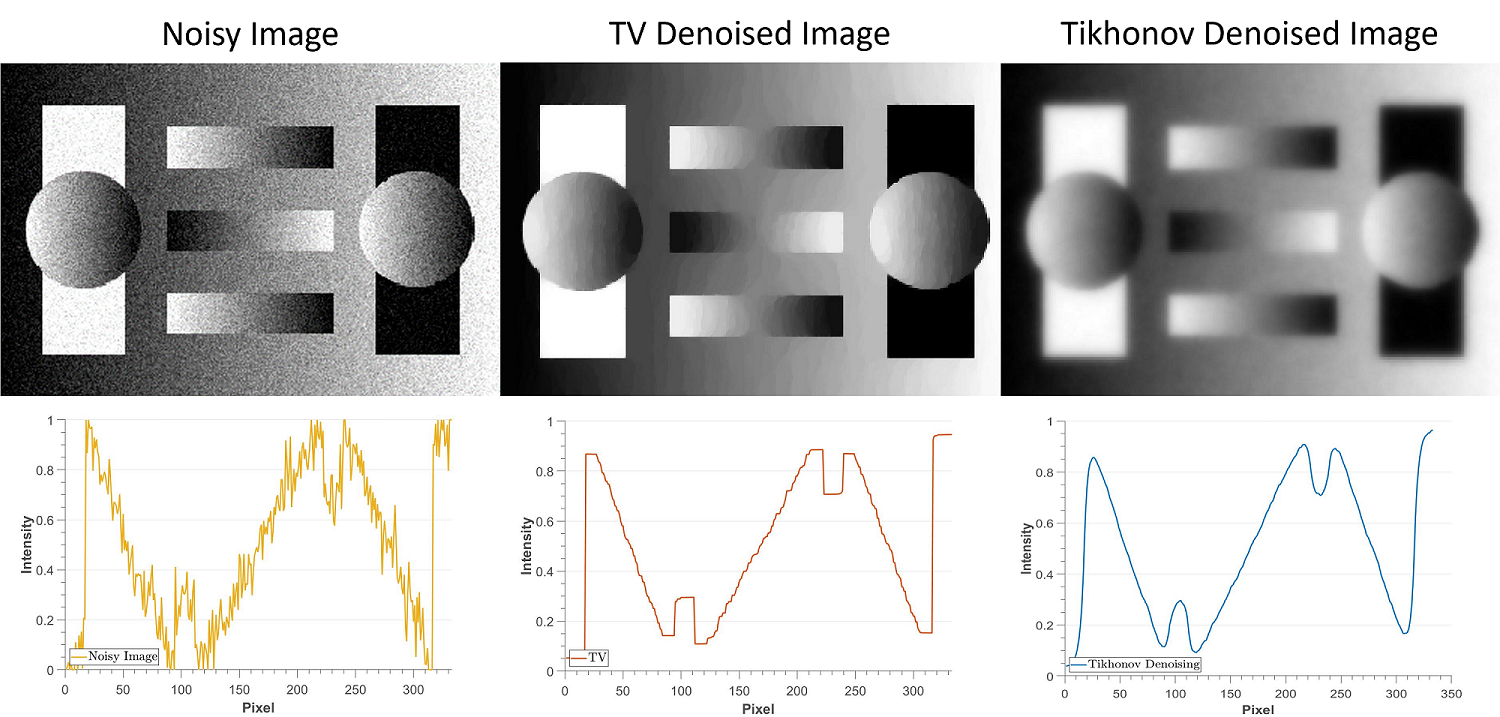}
    \vspace{-20pt}
    \caption{Geometric image with TV denoising applied.}
    \label{fig:TVGeom}
\end{figure}

\vspace{10pt}
The results obtained from TV denoising clearly show much stronger edge retention. The cartoon effect discussed earlier can also be seen, as the image is dominated by large blocks of colour. One point of note which can be observed is that the original image has a smooth gradient between shades of white and black, however, in the denoised image staircase artefacts can be observed. As mentioned before, TV regularisation encourages a piecewise constant function. This can fit blocks of colour well, however when there is a smooth intensity gradient in the image, the best approximation a piecewise constant function can produce leads to these staircase artefacts. Such artefacts can be significantly detrimental in some imaging applications. The aforementioned properties of this denoised image can be easily identified in a plot of the cross section of the geometric denoised image (Figure \ref{fig:TVGeom}), especially when compared to the same cross section of the Tikhonov denoised image.

\begin{figure}[h!]
    \centering
    \includegraphics[width = \textwidth]{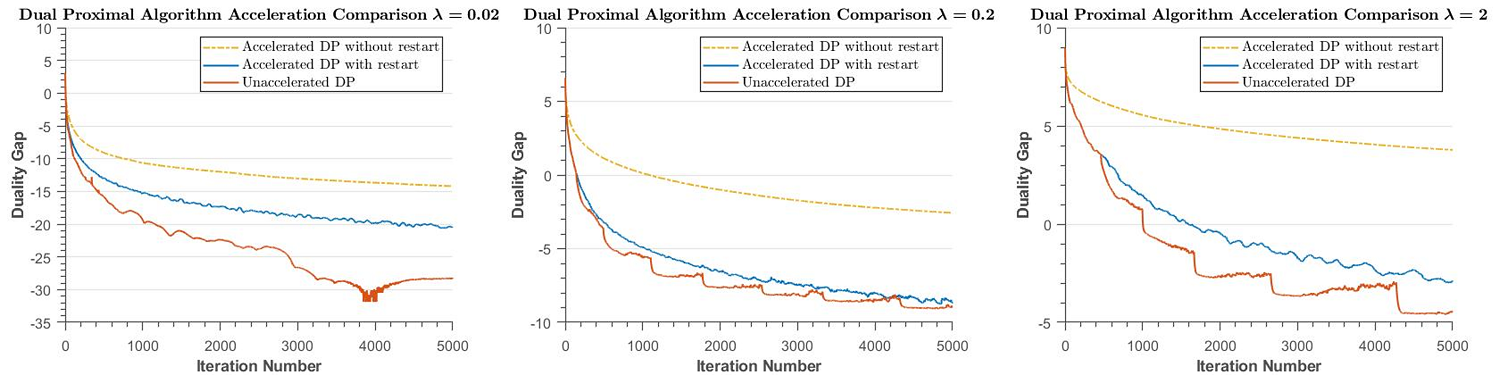}
    \vspace{-20pt}
    \caption{Comparison of the Dual Proximal Algorithm on the TV denoising problem.}
    \label{fig:DPGraphs}
\end{figure}

After a sufficient number of iterations, all 3 of the algorithms in Figure \ref{fig:DPGraphs} achieve similar results. However, the number of iterations and the time taken vary significantly between each method. The graphs containing the convergence rate for each of the three methods for $\lambda = 0.02,0.2,2$ can be observed in Figure \ref{fig:DPGraphs}.

ADPA, both with and without restart, significantly improves on the unaccelerated proximal gradient algorithm as stated in Theorem \ref{thm:AccProxGradConv}. We also observe that by applying restart to the accelerated proximal gradient method, the performance is again significantly improved. Additionally, the convergence rate decreases as $\lambda$ increases, which is a consequence of the step size decreasing (as the $\ell$-smooth parameter increases). Moreover, the effects of restart become less as $\lambda$ increases, and more generally as $\ell$ increases. 
 
\subsection{Comparison with state-of-the-art}
In this section I will compare the accelerated dual proximal gradient algorithm with other state-of-the-art methods for non-smooth optimisation. The two state-of-the-art methods which I have selected to compare to the dual proximal gradient algorithm are ADMM and Chambolle and Pock's primal dual method. The two methods differ slightly in the sense that ADMM is applied directly to the primal objective function whereas, as the name suggests, Chambolle and Pock's method must be applied to the primal dual function. Both of the following methods will be used as comparisons to other algorithms later in the thesis, so I will now offer a brief background into both of these methods.
 
 \subsubsection{Chambolle and Pock's Primal Dual Algorithm}
 Chambolle and Pock's primal dual algorithm acts on the primal dual function of the original primal objective function. This function will, by definition, be a mini-max problem as we will attempt to minimise the function with respect to the primal variable and maximise the function with respect to the dual variable. The algorithm is applied to functions of the following form
 \[\argminB_{\mathbf{u}} \argmaxB_{\mathbf{p}} \Big\{ \langle K\mathbf{u},\mathbf{p}\rangle + f(\mathbf{u}) - g(\mathbf{p})\Big\}\]
 where $g:X \rightarrow [0,+\infty]$ and $f: Y \rightarrow [0,+\infty]$ are convex functions. 
 
\bigskip
\begin{algorithm2e}[h!]
\caption{Primal Dual}\label{alg:Chambolle and Pock Primal Dual}
\SetKwInOut{Input}{Input}\SetKwInOut{Output}{Output}
\Input{$\tau,\sigma \leq \frac{1}{\ell}$, $\theta \in [0,1]$ and $(\mathbf{u}^{(0)},\mathbf{p}^{(0)}) \in U \times P$, $\bar{\mathbf{u}}^{(0)} = \mathbf{u}^{(0)}$}
\For {$k \leftarrow 0$ \KwTo $N$}{ 
$\mathbf{p}^{(k+1)} = \mathbf{prox}_{g,\sigma}(\mathbf{p}^{(k)} + \sigma K \bar{\mathbf{u}}^{(k)})$\\
$\mathbf{u}^{(k+1)} = \mathbf{prox}_{f,\tau}(\mathbf{u}^{(k)} - \tau K \mathbf{p}^{(k+1)})$\\
$\bar{\mathbf{u}}^{(k+1)} = \mathbf{u}^{(k+1)} + \theta(\mathbf{u}^{(k+1)} - \mathbf{u}^{(k)})$\\
}
\Output{$\mathbf{u}^{(N+1)}$}
\end{algorithm2e}
\bigskip
Some care must be taken when selecting the step size for both the primal and the dual steps in this algorithm. The step sizes must be selected such that $\tau \sigma L^{2} < 1$ where $L = \|K\| = {\rm{max}}\{|Kx|: x \in X$ with $|x| \leq 1\}$. In the case of TV denoising we have the following primal dual formula (which was derived above when deriving the dual function):
\[\argminB_{\mathbf{u}} \argmaxB_{\mathbf{p}} \frac{1}{2}\|\textbf{u} - \textbf{f}\|^{2}_{2} + \lambda\langle\nabla \mathbf{u},\mathbf{p} \rangle + \delta(\mathbf{p})\]
It is easy to see that $f(\mathbf{u}) = \frac{1}{2}\|\textbf{u} - \textbf{f}\|^{2}_{2}$, $K=\lambda \nabla$, and $g(\mathbf{p}) = \delta(\mathbf{p})$. In this case we have that $L = \lambda \sqrt{8}$, therefore we can set $\tau,\sigma = \frac{1}{\lambda \sqrt{8}}$. 

\subsubsection{Alternating Direction Method of Multipliers}
ADMM is a state-of-the-art method in optimisation. It is widely used because of the wide range of problems it can be applied to, as well as its strong performance qualities. ADMM is applied to problems of the following form:
\vspace{-5pt}
\[\argminB_{\mathbf{u,w}} f(\mathbf{u}) + g(\mathbf{w})\] subject to $A\mathbf{u} +B\mathbf{w} = \mathbf{b}$

where $\mathbf{u}\in \mathbb{R}^{n}$, $\mathbf{w} \in \mathbb{R}^{m}$, $\mathbf{b} \in \mathbb{R}^{p}$ $A \in \mathbb{R}^{p\times n}$ and $B \in \mathbb{R}^{p \times m}$ and $f$ and $g$ are convex functions. To formulate ADMM we must introduce the augmented Lagrangian of the objective function above.
\vspace{-20pt}
\begin{center}
    \[L_{\rho} = f(\mathbf{u}) + g(\mathbf{w}) +\mathbf{y}^{T}(A\mathbf{u} + B\mathbf{w} - \mathbf{b}) + \frac{\rho}{2}\|A\mathbf{u} + B\mathbf{w} - \mathbf{b}\|^{2}_{2}\]
\end{center}
This function can be broken down into the original objective function, the Lagrange multiplier and the augmented Lagrangian regularisation (the $L_{2}$ norm). The parameter $\rho$ is a hyperparameter which must be manually selected - it is normally selected between 3 and 5. This has moved us from a constrained optimisation problem to an unconstrained optimisation problem. The addition of the augmented Lagrangian term also has the added benefit of introducing an element of strong convexity to the problem, without effecting the optimal solution. As suggested by the `alternating direction' in ADMM, we can now solve this problem with respect to each of the 3 variables, $\mathbf{u}$,$\mathbf{w}$ and $\mathbf{y}$. The variables $\mathbf{u}$ and $\mathbf{w}$ in this case are our primal variables and $\mathbf{y}$ is our dual variable. The method proceeds by solving the following 3 subproblems:
\begin{align}
    \mathbf{u}^{(k+1)} &:= \argminB_{\mathbf{u}}L_{\rho}(\mathbf{u},\mathbf{w}^{(k)},\mathbf{y}^{(k)}) \\
    \mathbf{w}^{(k+1)} &:= \argminB_{\mathbf{v}}L_{\rho}(\mathbf{u}^{(k+1)},\mathbf{w},\mathbf{y}^{(k)}) \\
    \mathbf{y}^{(k+1)} &:= \mathbf{y}^{(k)} + \rho(A\mathbf{u}^{(k+1)}+B\mathbf{w}^{(k+1)}-\mathbf{b})
\end{align}
Algorithm \ref{alg:ADMM} uses the scaled form of each of the above formulas, however they are equivalent, despite using a slightly different dual variable ($\mathbf{q}$). 

\vspace{10pt}
\begin{algorithm2e}[h!]
\SetKwInOut{Input}{Input}\SetKwInOut{Output}{Output}
\caption{General ADMM}\label{alg:ADMM}
\Input{$N > 0$, $\rho > 0$, $\mathbf{u}^{(0)} \in \mathbb{R}^{n}$, $\mathbf{w} \in \mathbb{R}^{m}$, $\mathbf{q} \in \mathbb{R}^{p}$}  

\For{$k \leftarrow 0$ \KwTo $N$}{
$\mathbf{u}^{(k+1)} = \argminB_{\mathbf{u}}f(\mathbf{u}) + \frac{\rho}{2}\|A\mathbf{u}+B\mathbf{w}^{(k)}-\mathbf{b} + \mathbf{q}^{(k)}\|_{2}^{2}$ \\
$\mathbf{w}^{(k+1)} = \argminB_{\mathbf{w}}g(\mathbf{w}) + \frac{\rho}{2}\|A\mathbf{u}^{(k+1)}+B\mathbf{w}-\mathbf{b}+\mathbf{q}^{(k)}\|_{2}^{2}$ \\
$\mathbf{q}^{(k+1)} = \mathbf{q}^{(k)} + A\mathbf{u}^{(k+1)} + B\mathbf{w}^{(k+1)} - \mathbf{b}$ 
}
\Output{$\mathbf{u}^{(N+1)}$}
\end{algorithm2e}

For the update formulas of both primal variables, it is clear that they are strongly convex - the sum of a convex and strongly convex function results in a strongly convex function. This clearly offers an advantage as in many cases it allows closed form solutions to be derived for these two sub-problems. 

One point of note when applying ADDM is how potentially costly the matrix inversion for the first primal update can be. In order to find a cheap solution, the highly structured form of the matrix allows for quick DCT inversion (see Appendix \ref{App:FourInv}) to be applied. Whilst this method is far quicker than naive matrix inversion, it is key to note that this method involves discrete cosine transforms, which can become expensive when applied to high dimensional images. This is one motivation for using a gradient based method such as the dual proximal algorithm over ADMM, as it doesn't require any discrete trigonometric transforms. On the other hand, since there are no gradient steps in this algorithm, there is no dependence on the $\ell$-smooth parameter or the strong convexity parameter. This means that ADMM is less sensitive to the properties of the function (some reliance on the condition number remains due to the matrix inversions). 

\subsubsection{Numerical Results}
In order to measure the rate of convergence, a ground truth image was used. To obtain this ground truth I ran the accelerated dual proximal gradient algorithm for 500,000 iterations. Whilst this wouldn't necessarily give the exact ground truth because to get to machine accuracy, it may take many more iterations, it offers a point at which we can compare all of the other algorithms. The duality gap was used to ensure that the ground truth obtained was suitably accurate. Note, as expected, all values for $f(\mathbf{x}^{(k)}) - f^{\star}$ are positive which re-enforces our assumption that $f^{\star}$ is a very accurate solution. 

\vspace{10pt}
\begin{figure}[h!]
    \centering
    \includegraphics[width = \textwidth]{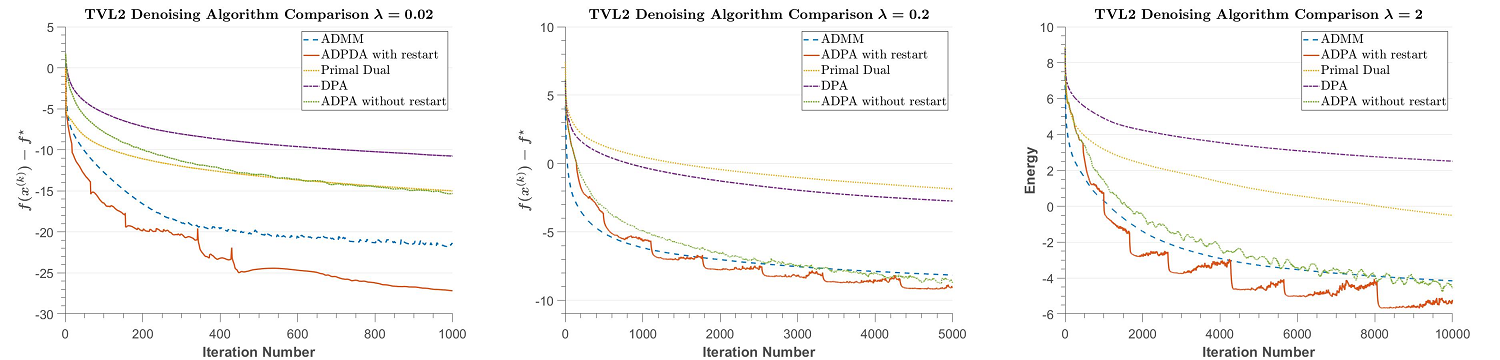}
    \vspace{-20pt}
    \caption{The error against iteration plots of different state-of-the-art algorithms when applied to TV denoising.}
    \label{fig:TVAlgCompIt}
\end{figure}
\vspace{10pt}
From the results in Figure \ref{fig:TVAlgCompIt} it is clear that as the number of iterations increase, ADPA outperforms ADMM on this problem. However, for a small number of iterations ADMM performs well. By applying restart to ADPA, not only does this improve the convergence of the algorithm, but it also increases it past that of ADMM. This pattern is something which can be observed often as ADMM offers very fast initial convergence, although it can be very slow to converge to machine accuracy. ADPA also provides a significant improvement on the primal dual algorithm applied to this problem. 

I collected data on the convergence of the algorithm for 3 different smoothing parameters. The main purpose of this was to observe how the algorithms performed on functions when the smooth component has different $\ell$-smooth values, since as $\lambda$ increases, the $\ell$-smooth parameter increases also. Firstly, it is clear that the convergence rate of all algorithms becomes lower as the $\ell$-smooth parameter increases. It can also be observed that ADPA with restart takes more iterations to cross ADMM as $\lambda$ increases. This is to be expected since the as the $\ell$-smooth parameter increases, the step size of ADPA decreases, whereas this is not the case for ADMM. Therefore we draw the conclusion that the smaller the $\ell$-smooth parameter the better ADPA performs in comparison to ADMM.  

\vspace{10pt}
\begin{figure}[h!]
    \centering
    \includegraphics[width = \textwidth]{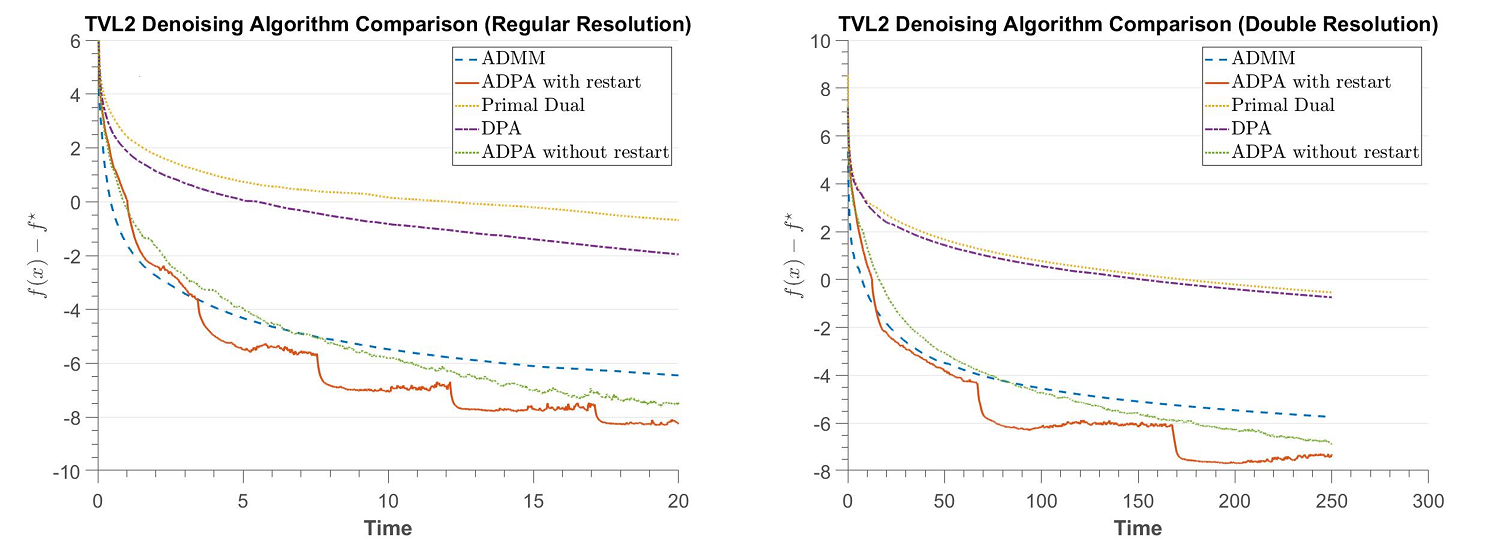}
    \vspace{-20pt}
    \caption{The convergence against time plots of different state-of-the-art algorithms when applied to TV denoising.}
    \label{fig:TVAlgCompTime}
\end{figure}
\vspace{10pt}
I also include a study of convergence against time in Figure \ref{fig:TVAlgCompTime}. I found that ADPA performed very well compared to other algorithms in terms of time taken to converge. Both experiments for the convergence rate were carried out on the noisy Lena image, with $\lambda = 0.2$. I tested the algorithms on this image in regular resolution, and then doubled the resolution. When the resolution increases, ADPA with restart crosses ADMM at a higher error value (-1 compared to -3 in the regular resolution image). As expected, this illustrates the that ADPA will be beneficial on higher dimensional images due to the fact it contains no DCT inversions. Since these values are higher than would be desired in terms of this solution, it can be concluded that ADPA with restart is the strongest performing algorithm, when $\lambda = 0.2$.

It is important to note that there is a trade-off in this case. For higher dimensional problems, ADPA outperforms ADMM, whereas for problems with a larger $\ell$-smooth parameter, ADMM outperforms ADPA. Therefore, in order to decide which of these algorithms will be best for any given problem, these properties need to be considered. 
\clearpage

\section{Total Smooth Variation (TSV) Denoising}\label{sec:TSV}
\subsection{Introduction to TSV Denoising}
As stated in Chapter \ref{sec:TV}, TV regularisation has some shortfalls, specifically due to the presence of staircase artefacts found in the solution. Such artefacts are a result of TV enforcing sparsity on the gradient vector. A signal with such a gradient will be a piecewise constant function, and will struggle to model smooth intensity gradients within an image. So whilst TV regularisation is effective for modelling images with piecewise constant intensity,  a different type of regularisation will be required to incorporate smooth gradients in an image.  One method of overcoming these artefacts is using Total Generalised Variation (TGV) regularisation:
\begin{equation} \label{eq:TGVObj}
\argminB_{\mathbf{u}}\frac{1}{2}\|\mathbf{u} - \mathbf{f}\|_{2}^{2} + \lambda\sum_{j=1}^{mn}\sqrt{((\nabla \mathbf{u})_{1,j}-\mathbf{w}_{1,j})^2 + ((\nabla \mathbf{u})_{2,j}-\mathbf{w}_{2,j})^2} + \beta\|\mathcal{E}(\mathbf{w})\|_{1}
\end{equation}
where $\mathcal{E}(\mathbf{w}) = \frac{1}{2}(\nabla \mathbf{w} + \nabla \mathbf{w}^{T})$ denotes the symmetric derivative of $\mathbf{w}$. This will in fact be a 2 by 2 matrix and by taking the $L_{1}$ norm we sum the absolute values of all 4 entries at any given pixel. 

The parameters $\lambda$ and $\beta$ are used to balance the first and second derivatives. In \textit{Second Order Total Generalized Variation (TGV) for MRI} by \cite{knoll2011second}, it is found via experimental results that the ratio of these parameters can be kept constant (with $\beta = 2\lambda$), therefore effectively no additional hyperparameter optimisation is required. In the original paper proposing the TGV denoising problem, the primal dual algorithm is used to optimise this function. We also know that ADMM  can be effectively used to optimise this function as shown in \cite{duan2016denoising} and \cite{lu2016implementation}, however there are no effective ways of accelerating these methods. 

My goal is to construct a regularisation term which overcomes the staircase artefacts which occur when applying TV regularisation, and that can be efficiently optimised using gradient based methods (therefore utilising Nesterov acceleration and adaptive restart methods discussed up to this point). To do so, we can consider constructing a regularisation term similar to TGV for which we can formulate the dual objective function. By addressing the non-smoothness and non-strong convexity of the function with respect to $\mathbf{p}$, we arrive at Total Smooth Variation (TSV), a regularisation term we can optimise using the same dual proximal algorithm applied to TV denoising.
\vspace{-20pt}
\begin{center}
\begin{align} \label{eq:TSVObj}
\begin{split}
\argminB_{\mathbf{u},\mathbf{w}} \frac{1}{2}\|\mathbf{u}-\mathbf{f}\|^{2}_{2} +&\lambda \sum_{j=1}^{mn} \sqrt{((\nabla \mathbf{u})_{(1),j}-\mathbf{w}_{(1),j})^2 + ((\nabla \mathbf{u})_{(2),j}-\mathbf{w}_{(2),j})^2} \\
&+ \frac{\beta}{2}\|\partial_{x} \mathbf{w}_{1}\|^{2}_{2} +\frac{\beta}{2}\|\partial_{y} \mathbf{w}_{2}\|^{2}_{2} + \frac{\gamma}{2}\|\mathbf{w}\|^{2}_{2}
\end{split}
\end{align}
\end{center}
\vspace{-10pt}
We address the non-smoothness of the function by swapping the $L_{1}$ norm applied to the symmetric derivative of $\mathbf{w}$ for the $L_{2}$ squared norm. The symmetric derivative is also replaced by the $x$ and $y$ derivatives of $\mathbf{w}_{1}$ and $\mathbf{w}_{2}$, respectively. This means all terms in the regularisation term can be differentiated, other than the TV term (which we know can easily be transferred to its dual form). The term $\frac{\gamma}{2}\|\mathbf{w}\|^{2}_{2}$ is then an additional term which is used to ensure that the function is strongly convex with respect to $\mathbf{w}$. This also adds another hyperparameter, $\gamma$; however, during the experiments carried out, we found that the function is not sensitive to this hyperparameter, so throughout we select $\gamma = 1$. The objective function is now strongly convex with respect to $\mathbf{w}$. 

One disadvantage is that this regularisation term appears to be more sensitive to the choice of $\beta$ than the original TGV regularisation, therefore $\beta$ must be selected independently for every application.

Below, I outline the formulation of the dual objective function as well as how this function can then be optimised using the ADPA. Firstly, we can apply the dual property of the TV regulariser (\ref{eq:TVDual}) to obtain the following primal dual problem.
\[\argminB_{\mathbf{u},\mathbf{w}} \argmaxB_{\mathbf{q}} \frac{1}{2}\|\mathbf{u}-\mathbf{f}\|^{2}_{2} + \lambda\langle \nabla \mathbf{u} - \mathbf{w} , \mathbf{q}\rangle + \frac{\beta}{2}\|\partial_{x} \mathbf{w}_{1}\|^{2}_{2} +\frac{\beta}{2}\|\partial_{y} \mathbf{w}_{2}\|^{2}_{2} + \frac{\gamma}{2}\|\mathbf{w}\|^{2}_{2} + \delta(\mathbf{q}) \]
This can then be differentiated with respect to each of the variables to obtain
\begin{itemize}
    \item With respect to \textbf{u}:
    \begin{equation} \label{eq:TSVuGrad}
        (\mathbf{u}-\mathbf{f}) - \lambda \nabla^{T}\mathbf{q}
    \end{equation}
    \item With respect to \textbf{w}:
    \begin{align}\label{eq:TSVpGrad}
    &-\lambda \mathbf{q}_{1} - \beta \partial_{x}^{T}\partial_{x}\mathbf{w}_{1} + \gamma \mathbf{w}_{1} \\
    &-\lambda \mathbf{q}_{2} - \beta \partial_{y}^{T}\partial_{y}\mathbf{w}_{2} + \gamma \mathbf{w}_{2}
    \end{align}
    \item With respect to \textbf{q} (only considering the smooth component):
    \begin{equation} \label{eq:TSVqGrad}
    \lambda(\nabla \mathbf{u} - \mathbf{w})
    \end{equation}
\end{itemize}

The Hessian matrix of the function with respect to $\mathbf{u}$ is clearly the identity matrix, of which all of the eigenvalues have value 1. Therefore, we can apply gradient descent with the optimal step size to find a closed form solution for $\mathbf{u}$ in terms of $\mathbf{q}$:
\begin{align}\label{eq:TSVuSoln}
\begin{split}
\mathbf{u}^{(k+1)} &= \mathbf{u}^{(k)} - \mathbf{u}^{(k)} + \mathbf{f} + \lambda \nabla^{T}\mathbf{q}^{(k)} \\
\mathbf{u}^{(k+1)} &= \mathbf{f} + \lambda \nabla^{T} \mathbf{q}^{(k)}
\end{split}
\end{align}

The other primal variable $\mathbf{w}$ must be handled in a slightly different manner since the Hessian matrix for $\mathbf{w}_{1}$ and $\mathbf{w}_{2}$ are $(\gamma I - \beta \partial_{x}^{T} \partial_{x})$ and $(\gamma I - \beta \partial_{y}^{T} \partial_{y})$, respectively. The eigenvalues of these matrices range from $\gamma$ to $4\beta + \gamma$. It is clear that including $\frac{\gamma}{2}\|\mathbf{w}\|^{2}_{2}$ is what facilitates the strong convexity of this function with respect to $\mathbf{w}$, and allows a closed form solution of $\mathbf{w}$ in terms of $\mathbf{q}$ to be derived. 
 Since both of the Hessian matrices are positive definite, i.e. have eigenvalues which are bounded from below by some value greater than 0, the problem is strongly convex with respect to both $\mathbf{w_{1}}$ and $\mathbf{w_{2}}$. Therefore, using the first order optimality condition (Theorem \ref{thm:FirstOrderOpt}) we can derive the following closed form solutions. Below I only derive the closed form solution for $\mathbf{w_{1}}$, however the solution is very similar for $\mathbf{w_{2}}$.
 
 \vspace{-20pt}
\begin{align}\label{eq:TSVpSoln}
\begin{split}
-&\lambda \mathbf{q}_{1} - \beta \partial_{x}^{T}\partial_{x}\mathbf{w}_{1} + \gamma \mathbf{w}_{1} = 0 \\
&(\lambda I - \beta \partial_{x}^{T}\partial_{x})\mathbf{w}_{1} = \lambda \mathbf{q}_{1} \\
&\mathbf{w}_{1} = (\gamma I - \beta \partial_{x}^{T}\partial_{x})^{-1}\lambda \mathbf{q}_{1} \\
\end{split}
\end{align}
Fortunately, the structure of the matrix can be utilised and the inversion can be performed quickly using a DCT inversion (see Appendix \ref{App:FourInv} with some modification). Using these two solutions we can then formulate the proximal gradient update for Algorithm \ref{alg:AccProxGrad} with respect to $\mathbf{q}$ (as \ref{eq:TSVpSoln} and \ref{eq:TSVuSoln} and be subbed into the below equation):
\[\mathbf{q}^{(k+1)} = \mathbf{prox}_{\delta(\cdot),t_{q}}(\mathbf{q}^{(k)} + t_{q} \lambda (\nabla \mathbf{u}^{(k+1)} - \mathbf{w}^{(k+1)}))\]
This algorithm can then be further accelerated using the ideas of restart which were applied to TV denoising also. 
One disadvantage of TSV when compared to TV, is that by solving $\mathbf{w}$ in terms of $\mathbf{q}$ for each loop we introduce a matrix inversion in every iteration. Whilst we can still solve this inversion quickly by using the DCT inversion, this still introduces a discrete cosine transform for each iteration. However, there will still be fewer applications of DCT when solving the ADPA than when applying ADMM to the same problem.  

Nevertheless, we are still required to calculate the step size, $t_q$, which entails calculating the $\ell$-smooth parameter of the objective function with respect to $\mathbf{q}$. We know the gradient of the function with respect to $\mathbf{q}$ is as follows from (\ref{eq:TSVqGrad})
\[\lambda(\nabla \mathbf{u}^{(k+1)} - \mathbf{w}^{(k+1)})\]
Since we know the closed form solutions of $\mathbf{u}^{(k+1)}$ from (\ref{eq:TSVuSoln}) and $\mathbf{w}^{(k+1)}$ from (\ref{eq:TSVpSoln}) with respect to $\mathbf{q}$ we know that the gradient can also be written as follows
\begin{center}
    \hfill$\nabla(\mathbf{f}+\lambda \nabla^{T}\mathbf{q}_{1}) - (\gamma I - \beta \partial_{x}^{T}\partial_{x})^{-1}\lambda \mathbf{q}_{1}$ \hfill     $\nabla(\mathbf{f}+\lambda \nabla^{T}\mathbf{q}_{2}) - (\gamma I - \beta \partial_{y}^{T}\partial_{y})^{-1}\lambda \mathbf{q}_{2}$\hfill   
\end{center}
for $\mathbf{q}_{1}$ and $\mathbf{q}_{2}$, respectively. Therefore the Hessian matrices for both $\mathbf{q}_{1}$ and $\mathbf{q}_{2}$ can easily be computed
\begin{center}
\hfill $\lambda(\lambda\nabla \nabla^{T} - (\gamma I - \beta \partial_{x}^{T}\partial_{x})^{-1}\lambda \mathbf{q}_{1})$ \hfill $\lambda(\lambda\nabla \nabla^{T} - (\gamma I - \beta \partial_{y}^{T}\partial_{y})^{-1}\lambda \mathbf{q}_{2})$  \hfill  
\end{center}

From here, we can easily see that the $\ell$-smooth parameter will be:
\[\lambda^{2}(8+\frac{1}{\gamma + 4 \beta})\]
Therefore the optimum step size will be the reciprocal of this value. We can once again see that the larger $\lambda$ is, the smaller the step size, leading to a slower convergence rate for this algorithm. 
\begin{algorithm2e}[h!]
\SetKwInOut{Input}{Input}\SetKwInOut{Output}{Output}
\caption{Accelerated Proximal Gradient Algorithm applied to TSV Denoising} \label{alg:TSVADPA}
\Input{$N > 0$, $t \leq \frac{1}{\ell}$, $\mathbf{\hat{q}}^{(0)}\in \mathbb{R}^{mn}$, $q \in [0,1]$} 
\For {$k \leftarrow 0$ \KwTo $N$}
{   
    $\mathbf{u}^{(k+1)} = \mathbf{f} + \lambda \nabla^{T} \mathbf{\hat{q}}^{(k)}$\\
    $\mathbf{w}_{1}^{(k+1)} = (\gamma I - \beta \partial_{x}^{T}\partial_{x})^{-1}\lambda \mathbf{\hat{q}}_{1}$\\
    $\mathbf{w}_{2}^{(k+1)} = (\gamma I - \beta \partial_{y}^{T}\partial_{y})^{-1}\lambda \mathbf{\hat{q}}_{2}$\\
    $\mathbf{q}^{(k+1)} =  \mathbf{\hat{q}}^{(k+1)} = \mathbf{proj}(\mathbf{\hat{q}}^{(k)} + t(\lambda(\nabla \mathbf{u}^{(k+1)} - \mathbf{w}^{(k+1)})))$\\
    $\theta^{(k+1)}$ solves $[\theta^{(k+1)}]^{2} = (1 - \theta^{(k+1)})[\theta^{(k)}]^{2} + q\theta^{(k+1)}$\\
    $\beta^{(k+1)} = \theta^{(k)}(1-\theta^{(k)})/([\theta^{(k)}]^{2} + \theta^{(k+1)})$\\
    $\mathbf{\hat{q}}^{(k+1)} = \mathbf{q}^{(k+1)} + \beta^{(k+1)}(\mathbf{q}^{(k+1)} - \mathbf{q}^{(k)})$ \\
    \If{$(\mathbf{\hat{q}}^{(k)}-\mathbf{q}^{(k+1)}) \cdot (\mathbf{q}^{(k)} - \mathbf{q}^{(k-1)}) > 0$}{
        $\theta^{(k+1)} = 1$
    }
}
$\mathbf{u} = \mathbf{f}+\lambda\nabla\cdot \mathbf{q}^{(N+1)}$ \\
\Output{$\mathbf{u}$}
\end{algorithm2e}

\subsection{Numerical Results}
\vspace{10pt}
\begin{figure}[H]
    \centering
    \includegraphics[width=\textwidth]{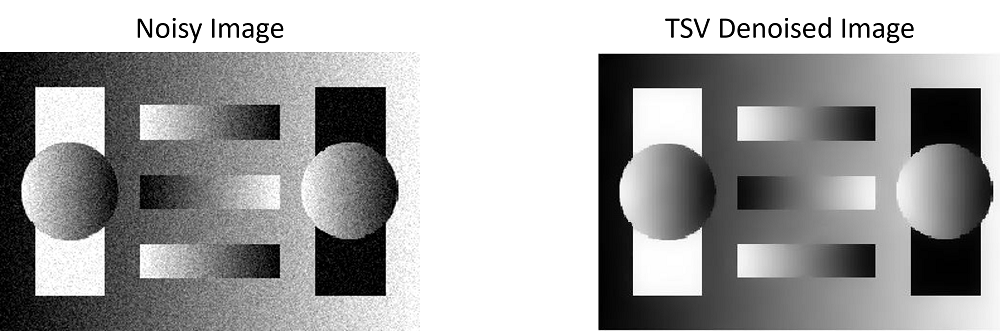}
    \vspace{-20pt}
    \caption{TSV denoising applied to a noisy geometric image.}
    \label{fig:TSVGeom}
\end{figure}

\begin{figure}[h!]
    \centering
    \includegraphics[width = \textwidth]{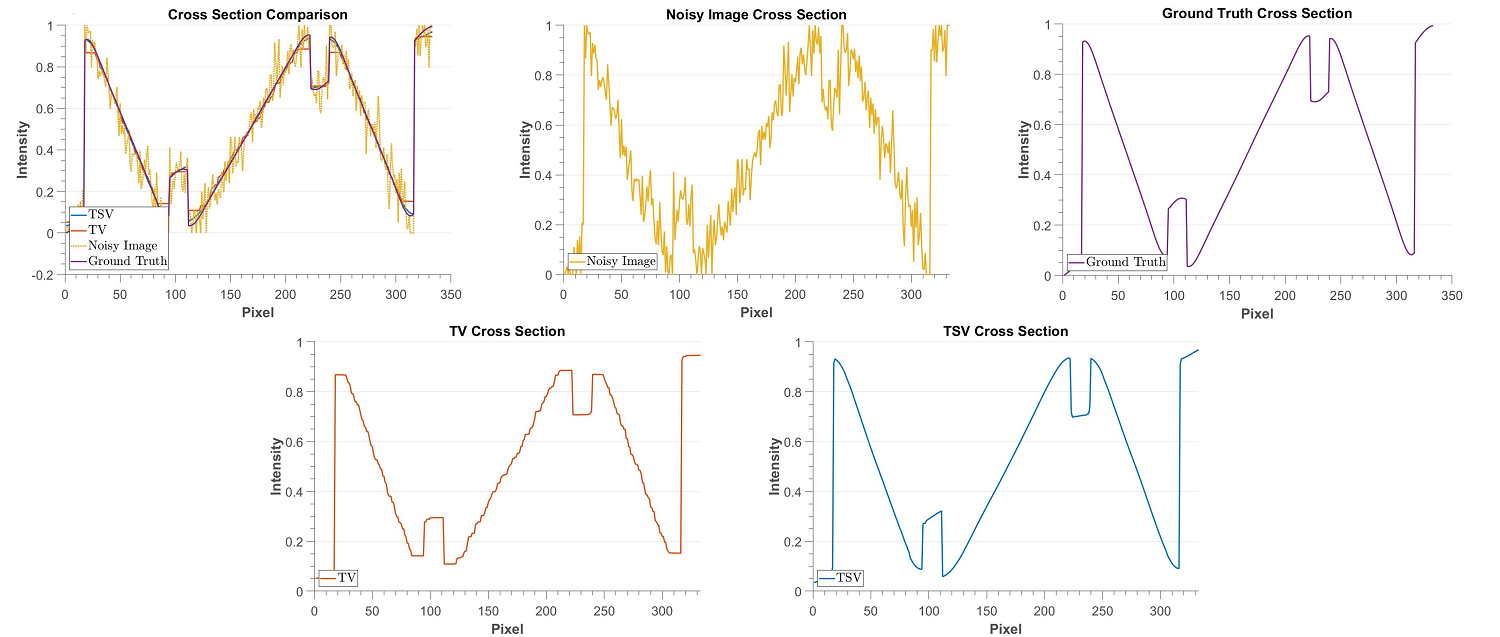}
    \vspace{-20pt}
    \caption{Cross sections of the geometric image denoised using a variety of regularisation terms.}
    \label{fig:TSVCross}
\end{figure}

Figure \ref{fig:TSVGeom} was obtained by applying TSV denoising with $\lambda = 0.1$ and $\beta = 100 = 1000\lambda$, it illustrates a lot of good properties of TV denoising, similar to the properties of TV denoising. For example, all of the noise has been removed, as well as having crisp edges. However, it is also clear that the staircase artefacts which were present in TV denoising are no longer an issue. Whilst this result is apparent from Figure \ref{fig:TSVGeom}, an interesting way of seeing this result is considering a cross section of the image, as seen in Figure \ref{fig:TSVCross}.

\vspace{10pt}
\begin{figure}[h!]
    \centering
    \includegraphics[width = \textwidth]{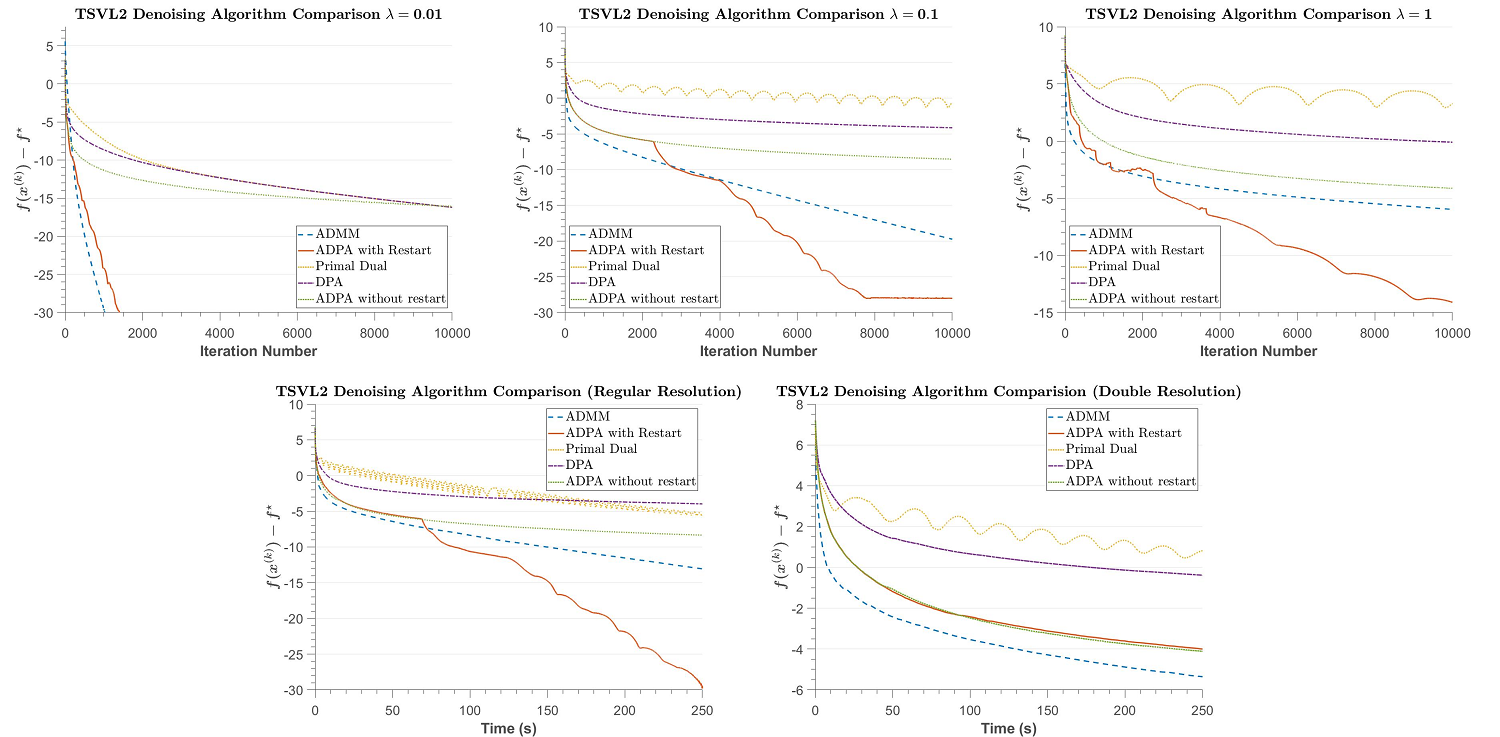}
    \vspace{-20pt}
    \caption{Comparison the convergence against iterations and time for the TSV denoising algorithm.}
    \label{fig:TSVAlgComp}
\end{figure}

\vspace{10pt}
In the graphs in Figure \ref{fig:TSVCross}, the piecewise nature of the TV denoising solution becomes obvious - whilst they both remove the noise from the original signal, the smooth gradient produced by TSV is clearly a much better fit for the smooth gradient in the original image. With regards to the convergence of the dual proximal algorithm acting on the TSV denoising problem, the following results can be seen in Figure \ref{fig:TSVAlgComp}.

Once again, I carried out tests for $\lambda = 0.01,0.1,1$ (with $\beta = 1000\lambda$). As $\lambda$ increases, the rate of convergence decreases for all algorithms. Similarly to TV, we have that ADMM performs the strongest out of the algorithms at first, and is then overtaken by ADPA after a certain number of iterations. The number of iterations and accuracy obtained by the algorithms varies with $\lambda$. In this case, as $\lambda$ increases, the point at which ADPA crosses ADMM occurs when the error is larger. This is the opposite of what is observed with TV denoising, and can be put down to the influence of $\beta$ on the $\ell$-smooth parameter. For all values of $\lambda$, it is clear that ADPA outperforms all other algorithms in terms of convergence to machine level accuracy. ADPA would be the preferred algorithm if machine accuracy is required for an application. ADPA would also be preferred on problems where large $\lambda$ is required. 

A number of conclusions can be drawn from Figure \ref{fig:TSVAlgComp}. Firstly, it reinforces the results from Chapter \ref{sec:TV}, which illustrate that applying restart to ADPA increases the performance in comparison to ADPA without restart. We can also see that while ADMM achieves better results for approximately the first 2000 iterations, ADPA with restart actually approaches machine accuracy significantly quicker than any of the other algorithms. 

I also showed the time taken for the algorithms to converge on both the regular resolution image and the double resolution image. As expected, doubling the resolution slows the convergence of all of the algorithms. For the regular resolution image, ADPA with restart and ADMM perform comparatively until around 75 seconds, at which point ADPA with restart begins to outperform ADMM significantly. However, in the case of the double resolution image, restart doesn't appear to impact ADPA, therefore ADMM is the strongest performing algorithm up to this point. Further investigation is required to analyse the performance on higher resolution images over larger timescales. 

\begin{figure}[h!]
    \centering
    \includegraphics[width = \textwidth]{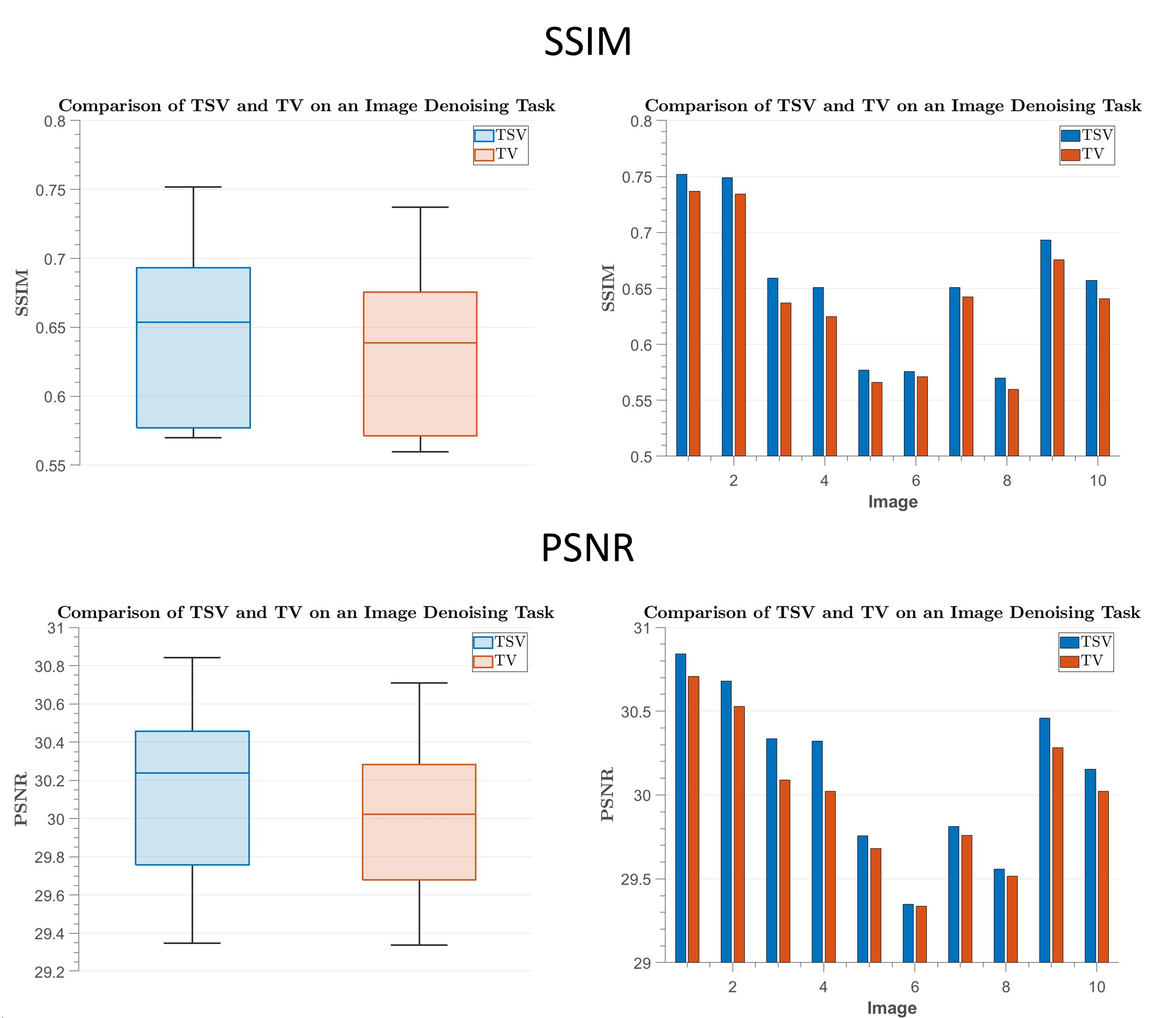}
    \vspace{-20pt}
    \caption{Quantative Results collected from TV and TSV denoising}
    \label{fig:TSVDenoiseData}
\end{figure}
I also carried out a study on a set of MRI images from the FastMRI dataset (\cite{zbontar2018fastmri}), which can be seen in Figure \ref{fig:TSVDenoiseData}. In order to simulate the intensity gradient present in images collected from a single coil MRI scanner, I applied a mask to each of the images which added a smooth gradient from high intensity at the bottom of the image, to low intensity at the top of the image. I then added Gaussian noise with a mean of $0$ and a variance of $0.005$ to each of the images. I applied both TV and TSV, with the smoothing parameter $\lambda$ set to $0.075$ for both methods. For TSV I set $\beta = 150$ and $\gamma = 1$. I then denoised each of the $10$ images using the ADPA algorithm for TV and TSV for $5000$ iterations. The results are compared using SSIM and PSNR between the denoised image and the original image before having Gaussian noise applied.In terms of both SSIM and PSNR TSV improves on TV for this denoising problem.  
\newpage
\begin{figure}[h!]
    \centering
    \includegraphics[width = 0.99\textwidth, height = 0.6\textwidth]{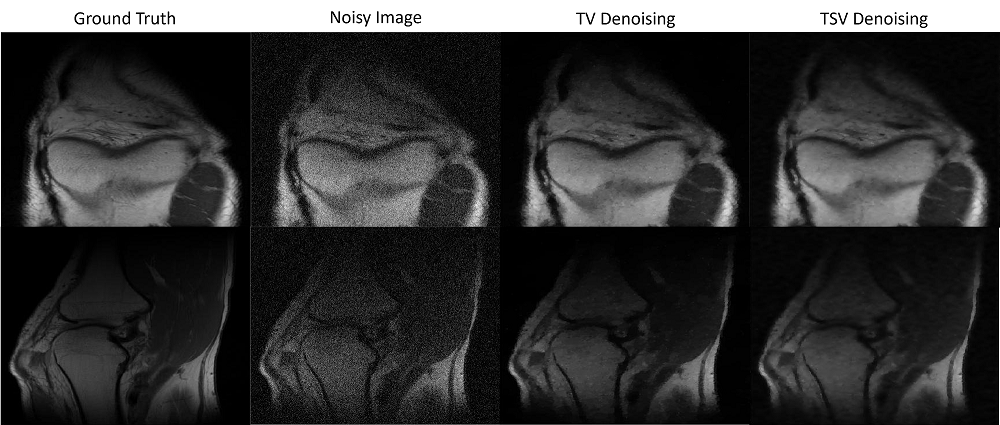}
    \vspace{-5pt}
    \caption{The results of applying TSV and TV denoising MRI images.}
    \label{fig:TSVDenoiseData}
\end{figure}

The images in Figure \ref{fig:TSVDenoiseData} show smoothed staircase artefacts in the TSV denoised image compared to the TV denoised image, and better noise reduction. Therefore, Figure \ref{fig:TSVDenoiseData} support the quantitative results that TSV performs better on this image denoising task. 

\clearpage
\section{Advanced Imaging Applications}\label{sec:Advanced}
\subsection{Introduction to Advanced Imaging Applications}
In Chapter \ref{sec:TSV} we dealt with denoising problems only, but variational methods can be extended to more advanced imaging problems. However, the data term used in these problems becomes more complex, and is no longer strongly convex, making them more difficult than the denoising problems. Below, we will consider the general form of advanced imaging applications and apply it to both MRI reconstruction and optical flow. Again, I aim use the ideas of acceleration and restart to produce an efficient algorithm to optimise these problems. 

The general form of advanced imaging problems is as follows
\begin{equation} \label{eq:ADVObj}
\argminB_{\mathbf{u}} \frac{1}{2}\|A\mathbf{u}-\mathbf{f}\|_{2}^{2} + \lambda {\rm TSV}(\mathbf{u})
\end{equation}
 where $A \in \mathbb{R}_{mn \times mn}$, $\mathbf{u,f} \in \mathbb{R}^{mn}$. 
 
 In the case of denoising, $A = I$, however, more generally, $A$ is a matrix which is not full rank. The properties of matrix $A$ have a big impact on how we can solve the problem. Generally, A will be some transformation on $\mathbf{u}$, therefore $\mathbf{f}$ will be some data we want the transformed image to be similar to, not necessarily in the image space. However, as is always the case with variational methods, the regularisation is a function of the gradient of the image. So whilst we want the image to be close to the data in terms of some other space, we want the image to be smooth in the image space.

By differentiating \ref{eq:ADVObj} with respect to $\mathbf{u}$ we obtain:
\[A^{T}(A\mathbf{u}-\mathbf{f}) \]
From this is can easily be seen that the Hessian matrix with respect to $\mathbf{u}$ is $A^{T}A$. We can consider two cases: when $A$ is full rank and when $A$ isn't full rank. When $A$ is full rank this implies that $A^{T}A$ will also be full rank. This has the implication that the objective function will be strongly convex with respect to $\mathbf{u}$ and we can therefore find a closed form solution using the first order optimality condition (Theorem \ref{thm:FirstOrderOpt}). However it is common that $A$ is not full rank, meaning the problem is not strongly convex. For this reason, we can consider applying the proximal gradient algorithm to this problem. In which case we get
\begin{align}
    \Tilde{\mathbf{u}}^{(k+1)} &= \mathbf{u}^{k} - t(A^{T}(A\mathbf{u}^{(k)}-\mathbf{f})) \\
    \mathbf{u}^{(k+1)} &= \mathbf{prox}_{{\rm{TSV}(\cdot)}}(\Tilde{\mathbf{u}}^{(k+1)})
\end{align}
We are required to solve the following proximal operator
\[\argminB_{\mathbf{u}}\frac{1}{2t}\|\mathbf{u}-\Tilde{\mathbf{u}}^{(k+1)}\|_{2}^{2} + \lambda {\rm{TSV}}(\mathbf{u})\]
Upon closer inspection, we can see that this is the TSV denoising problem (\ref{eq:TSVObj}) which we were able to find an efficient solution to in Chapter \ref{sec:TSV}. Therefore, in order to solve this problem we can apply accelerated proximal gradient algorithm to the original problem, and then solve the proximal step using the ADPA. The algorithm we propose to use to solve this problem can easily be generalised to TV regularisation following the same reasoning as above.
The step size for the outer loop will be the reciprocal of the largest eigenvalue of the matrix $A^{T}A$ and will change on an application-by-application basis. 
\bigskip
\begin{algorithm2e}[h!]
\caption{APGA for Advanced Applications}\label{alg:Accelerated Dual Proximal Algorithm}
\SetKwInOut{Input}{Input}
\SetKwInOut{Output}{Output}
\Input{$\mathbf{u}^{(0)} \in \mathbb{R}^{mn}$, $\mathbf{v}^{(0)} = \mathbf{0} \in \mathbb{R}^{mn}$ and $\theta^{(0)}=1$ }
\For {$k \leftarrow 0$ \KwTo N}{ 
$\theta^{(k+1)} = 1 + \sqrt{(1+4[\theta^{(k)}]^{2})/2}$ \\
$\Tilde{\mathbf{u}}^{(k+1)} = \mathbf{u}^{(k)} + ((\theta^{(k)}-1)/\theta^{(k+1)})\mathbf{v}^{(k)}$ \\
$\mathbf{u}^{(k+1)} = {\rm{TSVDenoising}}(\Tilde{\mathbf{u}}^{(k)} - t_{u}A^{T}(A\Tilde{\mathbf{u}}^{(k)}-\mathbf{f}))$ \\
$\mathbf{v}^{(k+1)} = \mathbf{u}^{(k+1)} - \mathbf{u}^{(k)}$ \\
}
\Output{$\mathbf{u}^{(N+1)}$}
\end{algorithm2e}

One disadvantage of Algorithm \ref{alg:Accelerated Dual Proximal Algorithm} is the need to solve the proximal operator using an iterative algorithm. This does mean that the algorithm is a 2-loop algorithm, so it may take longer to solve, however, the results below show efficient convergence to high accuracy solutions in a smaller number of outer iterations. We also know that the inner loop can be solved efficiently using either ADMM or ADPA, and the number of iterations should not be too great. 

\subsection{MRI Reconstruction}
MRI scans are commonly used in many areas of medicine, and play a significant role in medical diagnosis. The images obtained from MRI scans may be analysed by a medical professional directly or, due to the recent advance of AI technology, may be passed into machine learning algorithms. It is of the upmost importance that the images used are free of noise and artefacts. However, MRI has an inherently expensive data acquisition process, which can lead to: artefacts from patient movement, patient discomfort and additional cost. For this reason it is desirable to reduce the time taken to obtain images as much as possible.

The data that MRI scanners collect is in the $k$-space rather than the image space. Due to the global nature of the $k$-space, the data can be heavily undersampled and yet we maintain the possibility of reconstructing the original image from the undersampled data. By undersampling the $k$-space, the process of obtaining an MRI scan is sped up significantly; if an undersampling rate of 4 is measured (i.e. only 25\% of the $k$-space is collected) then the time taken to carry out the scan decreases by a factor of 4. However, by collecting less data, the output image becomes noisy, and contains undersampling artefacts.

In order to obtain an artefact free image, we can use prior knowledge to formulate a regularisation term to allow us to produce an image with desirable properties. Artefacts often manifest themselves as significant changes of intensity in the image, and so we can use variational methods in order to reconstruct MRI images. The objective function we look to minimise is as follows
\begin{equation}\label{eq:MRObj}
\argminB_{\mathbf{u}} \|{\mathcal{DF}}\mathbf{u} - \mathbf{m}\|^{2}_{2} + \lambda {\rm{TSV}}(\mathbf{u})
\end{equation}
Where $\mathbf{u,m}\in \mathbb{R}^{mn}$ and $\lambda \in \mathbb{R}$, $\mathbf{m}$ is the undersampled k-space data. By comparing this objective function (\ref{eq:MRObj}) to the general objective function (\ref{eq:ADVObj}), it is clear that in this case $A = {\mathcal{DF}}$, where ${\mathcal{D}}$ is the undersampling pattern used. This will be a diagonal matrix whos only entries are ones and zeros, and ${\mathcal{F}}$ is the discrete Fourier transform matrix. We can note that in the case where ${\mathcal{D}}$ is the identity, we get a fully sampled $k$-space - such a problem would be trivial since $A = {\mathcal{F}}$ is full rank and therefore invertible and there would be no need for regularisation. In the case when $A = {\mathcal{DF}}$, the eigenvalues of the matrix $A^{T}A$ will all either be 1 or 0. This tells us that the problem is only convex rather than strongly convex, and also that the optimal step size will be 1. 

In this problem we can see clearly that the data term ensures that the solution is close to the data in the $k$-space. So in this case, $\mathbf{m}$ would be the $k$-space data collected from the MRI scanner. It is important to reiterate that in this case we want the image to be smooth rather than the $k$-space to be smooth. 

On top of noise removal, it is important to consider what other properties a regularisation term is required to have to be successfully be applied to MRI reconstruction. One characteristic of MRI scans is that they contain intensity gradients; the image becomes dimmer as it moves away from the MRI coil. This immediately indicates that TSV would have an advantage over TV regularisation, as it would eradicate undesirable staircase artefacts on this intensity gradient.

The results I will compare for this task are a combination of my findings from earlier in the paper. Firstly, I will compare the convergence rates of accelerated proximal gradient descent with restart to other optimisation methods, as well as illustrate the efficacy of TSV as a regularisation method for MRI reconstruction using both perceptual and quantative results. 

\subsubsection{MRI Reconstruction Numerical Results}
\begin{figure}[h!]
    \centering
    \includegraphics[width = \textwidth]{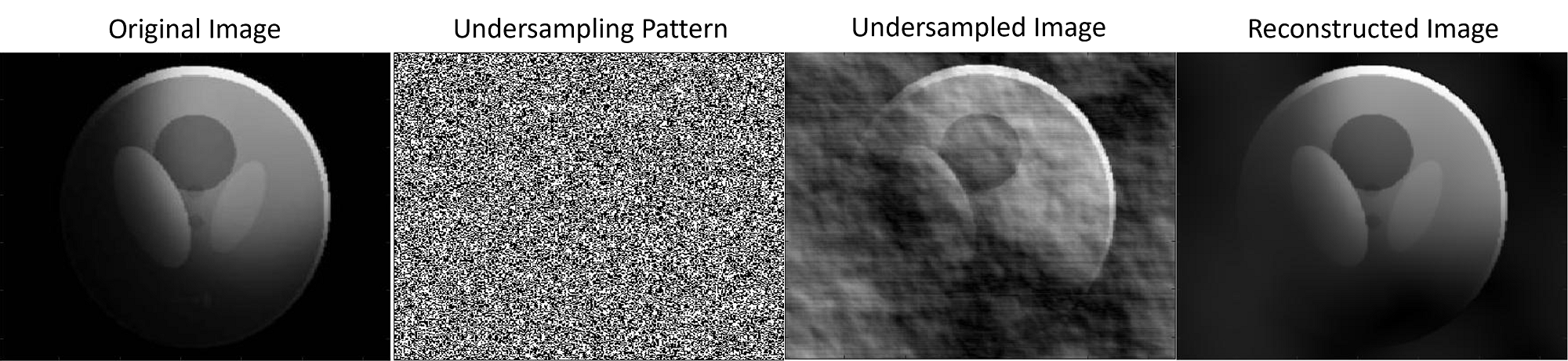}
    \vspace{-20pt}
    \caption{TSV MRI Reconstruction applied to an example image aimed to simulate and MRI image.}
    \label{fig:MRReconToy}
\end{figure}

\begin{figure}[h!]
    \centering
    \includegraphics[width = \textwidth]{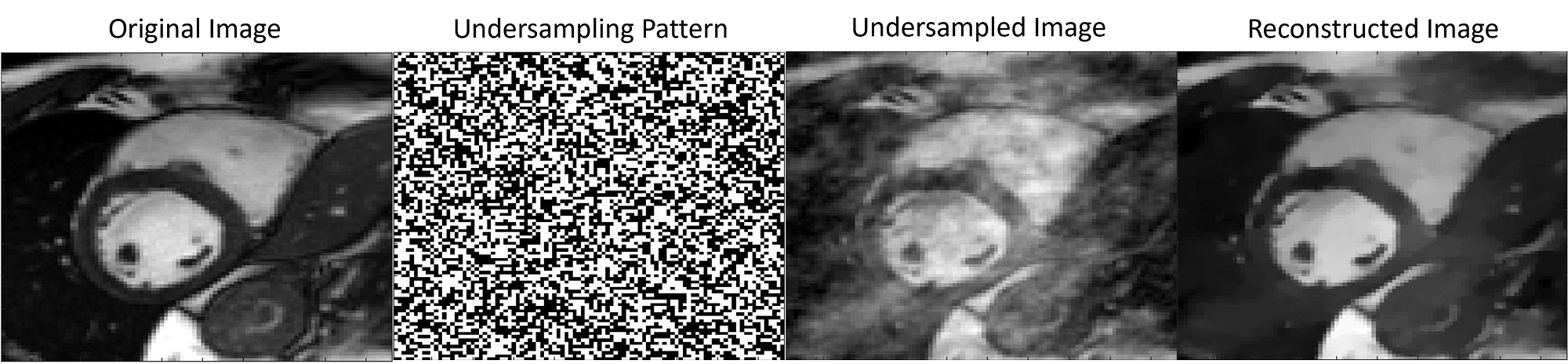}
    \vspace{-20pt}
    \caption{TSV MRI Reconstruction applied to a cardiac image.}
    \label{fig:MRReconCard}
\end{figure}

Figure \ref{fig:MRReconToy} and Figure \ref{fig:MRReconCard} both illustrate perceptually how well TSV works on some simple examples. In both images the artefacts, caused by the undersampling of the $k$-space, are removed. The images retain sharp edges from the original image without containing any staircase artefacts where smooth gradients are present in the image. 
\vspace{10pt}
\vspace{10pt}
\vspace{10pt}

\begin{figure}[h!]
\centering
\includegraphics[width=.75\textwidth]{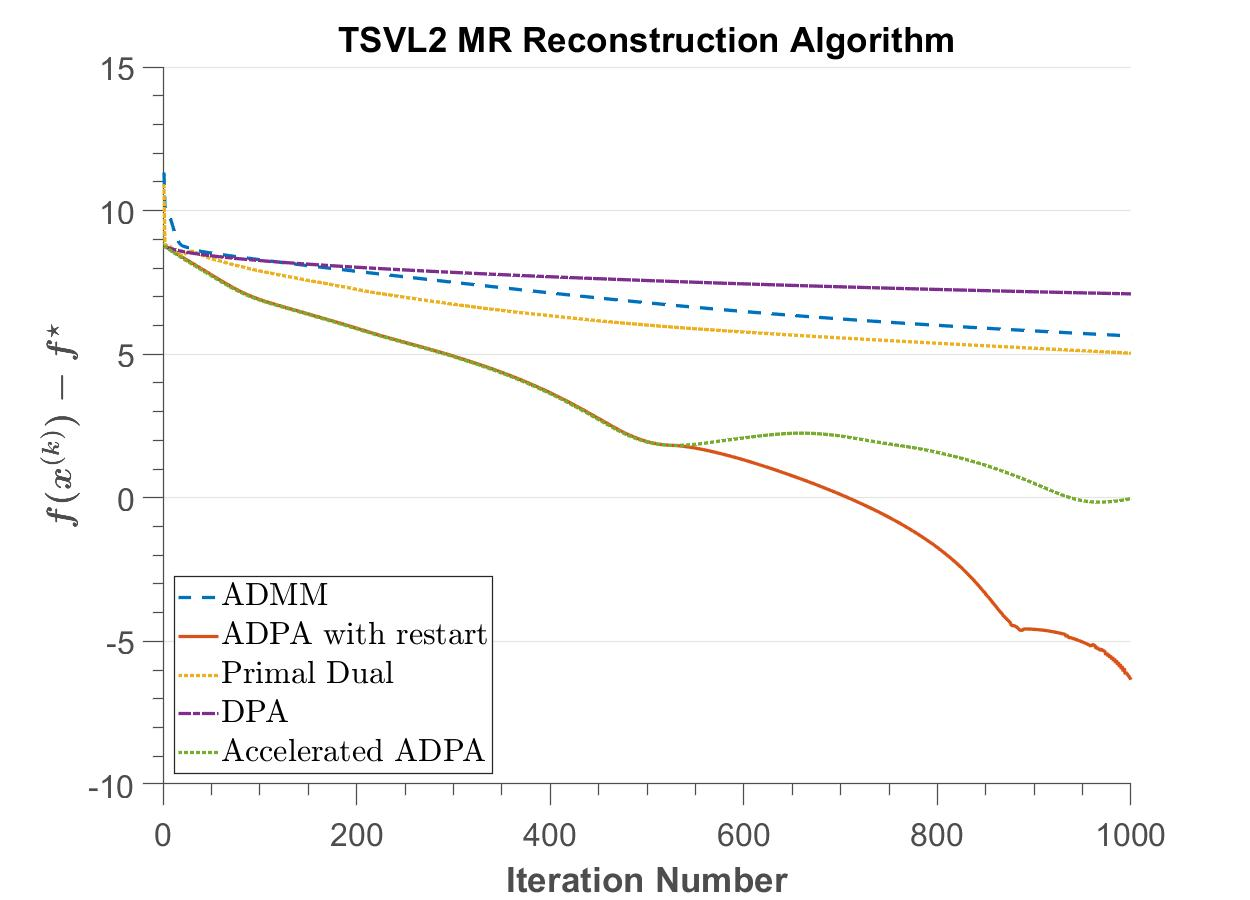}\hfill

\caption{Comparison of state-of-the-art algorithms applied to the MRI Reconstruction task.}
\label{fig:MRReconAlgComp}

\end{figure}

It can be seen in Figure \ref{fig:MRReconAlgComp} that in terms of the outer loop, the convergence of ADPA towards machine accuracy is significantly quicker than the other algorithms, infact in this case it converges significantly quicker than the other algorithms full stop. This mainly shows the impact of applying the proximal gradient algorithm to the general advanced imaging problem. It increases the convergence rate signicantly. Acceleration clearly has a big impact on applying the proximal gradient algorithm to this problem as without acceleration, applying the proximal gradient algorithm performs worse than any of the other algorithms. Note that there is still the fact that the inner loops may take longer to solve than the other methods, however, this outer loop convergence is very promising. 

\begin{figure}[h!]
    \centering
    \includegraphics[width = 0.99\textwidth, height = 0.8\textwidth]{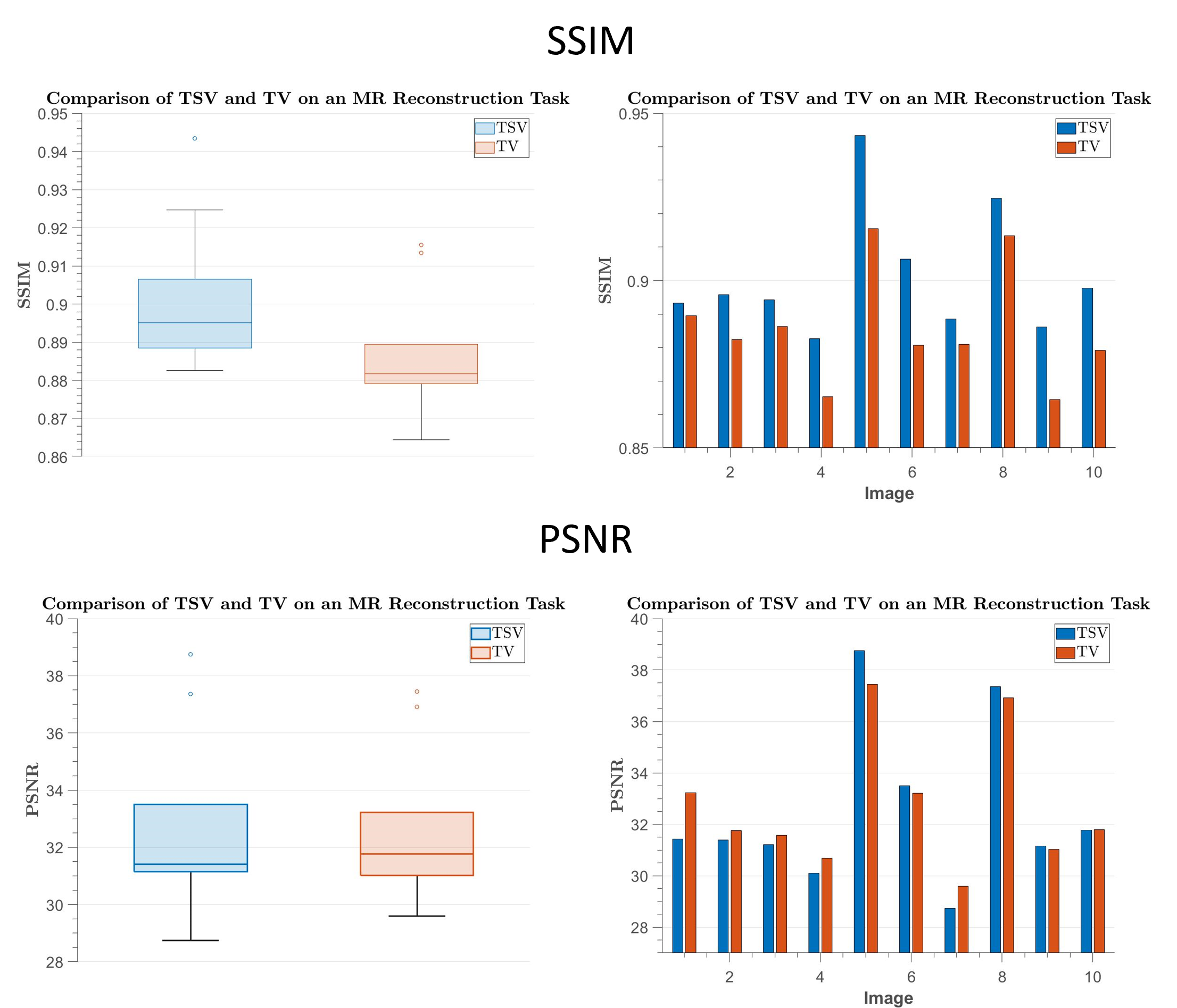}
    \vspace{-5pt}
    \caption{Quantitative results collected from TV and TSV denoising}
    \label{fig:TSVMRData}
\end{figure}

Some quantitative results for the performance of TSV on the MRI reconstruction task were also collected on the same subset from the FastMRI data set used for image denoising. These results can be observed in Figure \ref{fig:TSVMRData}. The same process of masking the images to simulate their collection from a single coil MRI scanner was carried out. However, in this case compressed sensing was applied. For both TSV and TV, the smoothing parameter $\lambda$ was set to 0.075, however, in this case I found the TSV performed more effectively when $\beta$ was set to 15, illustrating how $\beta$ is required to be tuned depending on the application. Once again, $\gamma$ was set to 1. I used ADPA, with 100 iterations for both the inner and outer loops, to optimise the objective functions of both TSV and TV. 

\begin{figure}[h!]
    \centering
    \includegraphics[width = 0.99\textwidth, height = 0.6\textwidth]{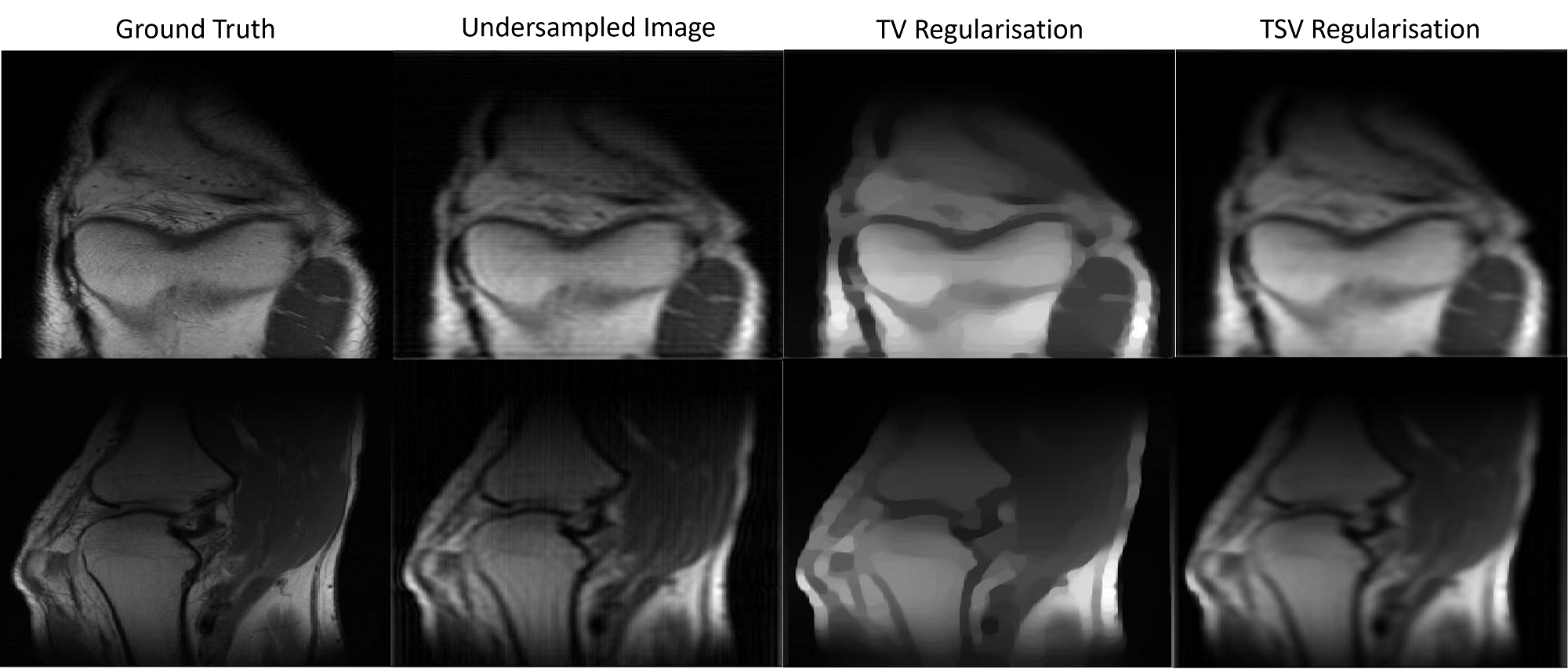}
    \vspace{-5pt}
    \caption{Results for 2 images from the FastMRI dataset.}
    \label{fig:TSVMRIm}
\end{figure}

These results show that TSV outperforms TV in terms of SSIM and the two methods perform similarly in terms of PSNR. Two examples from this data set are illustrated in Figure \ref{fig:TSVMRIm}. The images which TSV regularisation is applied to appear far closer to the original image, compared to the images with TV applied to them which have a cartoon-like effect. These images reinforce the quantitative results given in \ref{fig:TSVMRData}.

\newpage
\subsection{Optical Flow}
Optical flow is the task of tracking the apparent movement of brightness patterns between a pair of images. Optical flow has many different applications, ranging from video compression to medical image analysis; it is therefore another taks for which there is a great need to accelerate. The problem is formulated by considering the Horn-Schunk formulation of the problem. We can see that the assumption that the optical flow field will be smooth is a a valid constraint as we would expect nearby pixels to have the same motion. However, there may be discontinuities in this motion - for example, at the edge of an object. For this reason TSV is deemed to be an appropriate regularisation for this task. 

The objective function for Optical flow is as follows:
\begin{equation}\label{eq:OFObj}
\argminB_{\mathbf{u},\mathbf{v}}\|\langle I_{x},\mathbf{u}\rangle + \langle I_{y},\mathbf{v} \rangle + I_{t}\|_{2}^{2} + \lambda {\rm{TSV}}(\mathbf{u},\mathbf{v})
\end{equation}
where $\mathbf{u,v} \in \mathbb{R}^{mn}$ and $I_{x},I_{y},I_{t} \in \mathbb{R}^{mn}$. 

This problem is slightly more challenging than those faced before since it is now a two-channelled problem. Here, $\mathbf{u}$ represents the motion of each pixel in the $x$ direction and $\mathbf{v}$ represents the motion of each pixel in the $y$ direction. However, the methods applied to one-channelled problems can easily be generalised to this two-channelled problem. 
In this case, the gradient updates for the outer loop can be performed on $\mathbf{u}$ and $\mathbf{v}$ respectively, as follows 
\begin{align}
    \Tilde{\mathbf{u}}^{(k+1)} &= \mathbf{u}^{(k)} - t\langle I_{x}, \langle I_{x},\mathbf{u}^{(k)} \rangle + \langle I_{y}, \mathbf{v}^{(k)} \rangle + I_{t} \rangle \\
    \Tilde{\mathbf{v}}^{(k+1)} &= \mathbf{v}^{(k)} - t\langle I_{y}, \langle I_{x},\mathbf{u}^{(k)} \rangle + \langle I_{y}, \mathbf{v}^{(k)} \rangle + I_{t} \rangle
\end{align}

Now the proximal step associated with TSV denoising must be applied to the problem. If anisotropic TSV is applied then $\Tilde{\mathbf{u}}^{(k+1)}$ and $\Tilde{\mathbf{v}}^{(k+1)}$ can have TSV denoising applied independently. In the case of isotropic TSV, the projection step must be carried out jointly. However, the modification is simple, illustrating that the algorithm can easily be generalised to two-channelled images.

Since the objective function in this case is made up of 2 variables, $\mathbf{u}$ and $\mathbf{v}$, we must find the maximum eigenvalues of the following block matrix to find the step size $t$ for this algorithm.
\begin{equation}
\begin{pmatrix}
{\rm{diag}}(I_{x} \odot I_{x}) & {\rm{diag}}(I_{x} \odot I_{y}) \\
{\rm{diag}}(I_{y} \odot I_{x}) & {\rm{diag}}(I_{y} \odot I_{y})
\end{pmatrix}
\end{equation}
Where $\odot$ is the Hadamard product. Using the knowledge that $I_{x}$ and $I_{y}$ are diagonal matrices with values between 0 and 1 and therefore all of the blocks which make up this matrix are also diagonal matrices, bounded between the same values.I identified, experimentally, that the eigenvalues of this matrix are bounded above by 2, therefore the step size $t = \frac{1}{2}$. 

\subsubsection{Optical Flow Numerical Results}
\begin{figure}[h]
    \centering
    \includegraphics[width = 0.9\textwidth]{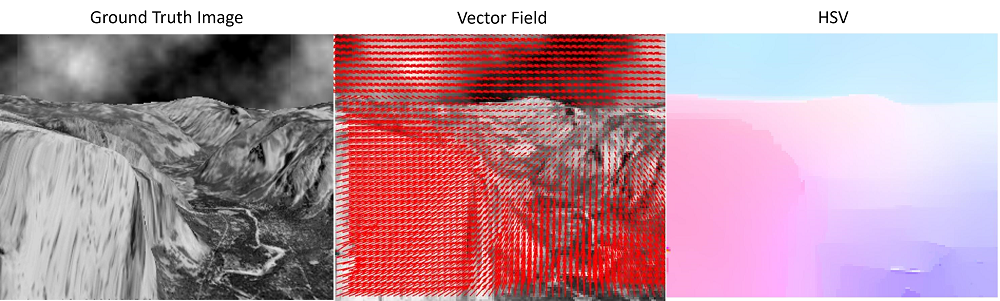}
    \vspace{-5pt}
    \caption{Optical flow on 2 images of a landscape}
    \label{fig:YOSOFlandscape}
\end{figure}

\begin{figure}[h]
    \centering
    \includegraphics[width = 0.9\textwidth]{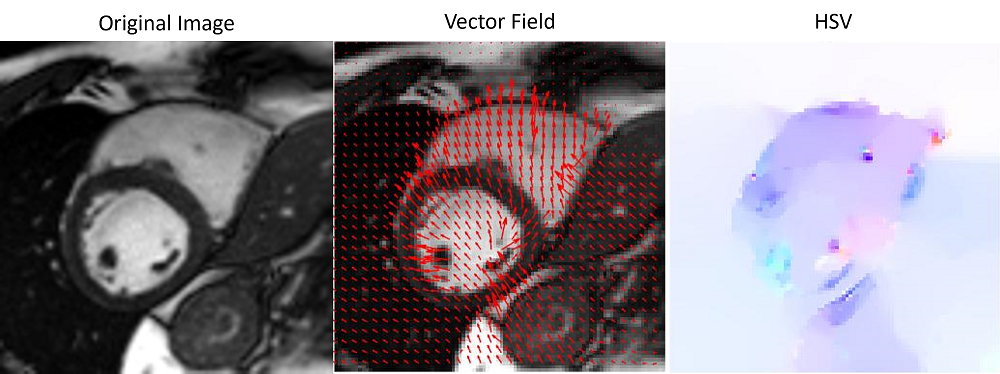}
    \vspace{-5pt}
    \caption{Optical flow on 2 cardiac images}
    \label{fig:YOSOFcardiac}
\end{figure}

The Optical flow results in Figures \ref{fig:YOSOFlandscape} and \ref{fig:YOSOFcardiac} were collected using TSV with $\lambda=0.003$ and $\beta = 3000\lambda$, they illustrate how TSV can capture discontinuities in results without incurring the staircase artefacts done so by TV. Both of the results are collected between a source image and a target image, between which there is a slight motion. The vector field displays the motion of each pixel using a vector. The HSV image uses colour to display the direction and the magnitude of such vectors. The hue of any pixel represents the angle of motion and the value is indicative of the magnitude of the vector at this point.

\vspace{10pt}
We see in both examples that using TSV regularisation captures the motion between the two images accurately. In Figure \ref{fig:YOSOFlandscape}, a sharp discontinuity between the valley and the sky indicates TSV's ability to capture discontinuities. In this same image, the continuous change between the left and right of the image shows that TSV effectively captures a smooth change of motion.

\begin{figure}[h]
    \centering
    \includegraphics[width = .75\textwidth]{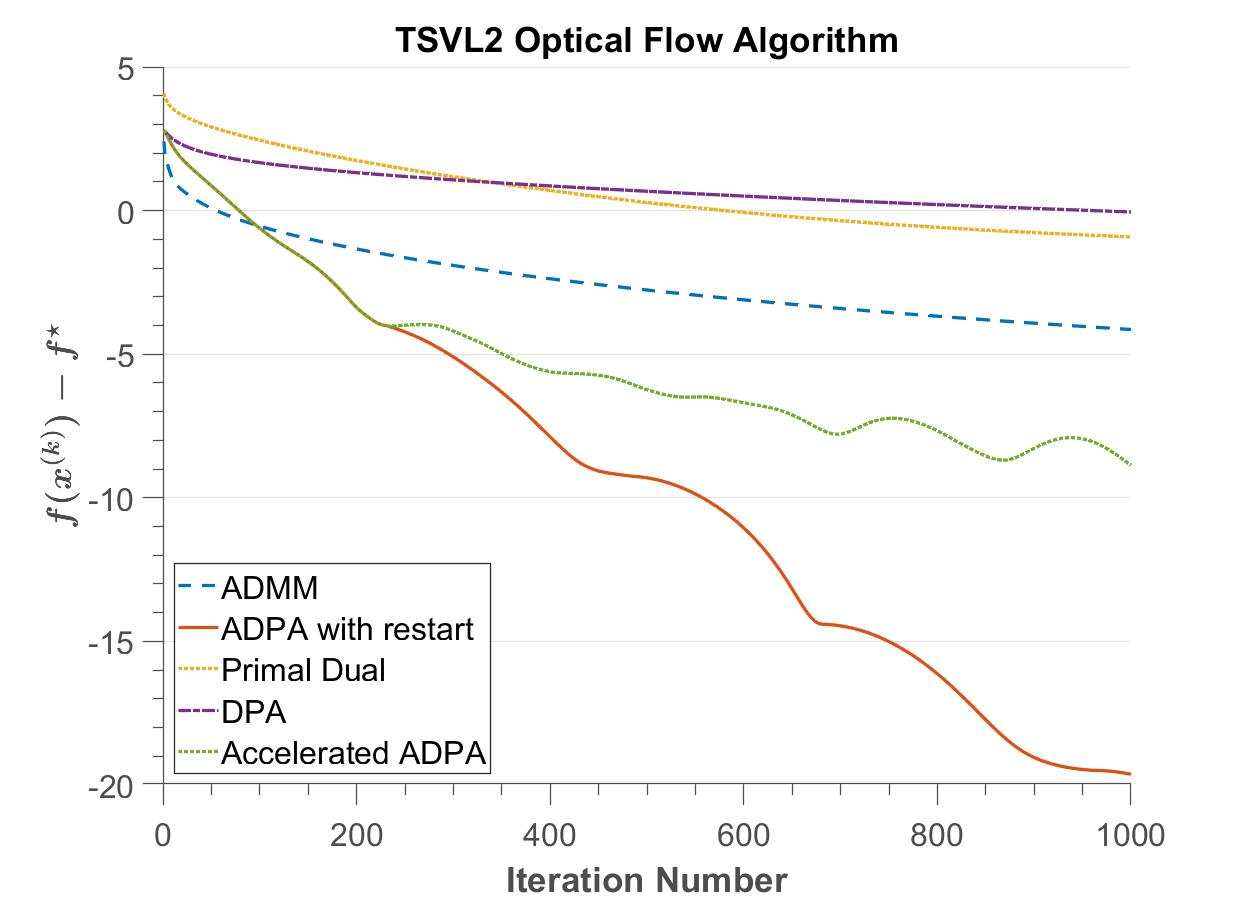}
    \vspace{-5pt}
    \caption{A plot of convergence against iteration data for state-of-the-art algorithms applied to the optical flow problem with TSV regularisation.}
    \label{fig:OFPlot}
\end{figure}

The convergence results in Figure \ref{fig:OFPlot} show how applying restart to ADPA increases the rate of convergence of the outer loop significantly compared to ADPA without restart. We can also observe that whilst ADMM converges quicker than ADPA with restart, for the first few iterations ADPA soon becomes the most effective algorithm, converging towards machine accuracy far quicker than any of the other algorithms. It is also worth noting that ADPA exceeds ADMM in terms of accuracy when the iterations are around 100 and the error is at around -1. Since this is insufficiently accurate, we can conclude that ADPA outperforms ADMM in terms of solving this optical flow problem.  

\clearpage
\section{Conclusion and Summary}\label{sec:conc}
Throughout this thesis, I have utilised state-of-the-art methods applied to variational imaging methods. I have combined adaptive restart methods used in optimisation, and conducted a study applying them to a range of regularisation techniques, each of which builds on the previous to overcome certain shortfalls. Two main ideas were covered - firstly, how adaptive restart methods improve acceleration when applied to almost any problem. This finding then motivated the need for an adaptation of TGV to utilise the efficacy of adaptive restart techniques. 

The results on adaptive restart when compared to state-of-the-art methods show that it competes in a range of situations. Such adaptive restart methods showed a clear advantage over acceleration with no adaptive restart. It improved the convergence of the proximal gradient algorithm whilst adding minimal extra expense per iteration. This study is the first in the literature to show how effective adaptive restart can be when applied to the proximal gradient algorithm. This conclusion was illustrated in all tests carried out, therefore we can draw the conclusion that restart can always be applied to improve the performance of the accelerated proximal gradient algorithm. 

TSV showed impressive performance on a range of tasks. Firstly, it performed very well on the denoising tasks. It produced perceptually very good results on some toy problems, retaining sharp edges, but removing previously present staircase artefacts. The resulting image also achieved improved results quantitatively compared to TV in terms of SSIM values, specifically on both denoising and MRI reconstruction problems. Optimising problems with TSV regularisation applied often lead to quicker results, especially in terms of outer iterations. 

I also highlighted the method used to convert primal problems to dual problems, upon which proximal gradient algorithm can be applied. This is a framework which can be applied to many problems, especially those containing the total variation term. Although this method isn't novel, we are able to use these ideas to motivate TSV - a novel solution to removing the staircase artefacts found when using the total variation regularisation technique. This idea of converting the problem into its dual form using the strongly convex property of the primal variables is then exploited when formulating TSV to allow the application of the efficient algorithm ADPA. Manipulating the regularisation term to include strongly convex terms is an idea which could be extended to to other functions to provide efficient optimisation algorithms. 

Throughout this project I have covered a large range of topics, in the areas of regularisation terms and optimisation algorithms. Bringing these two fields together offers an insight into how they are linked together and can motivate discoveries in the other respective field. However, covering such a broad spectrum comes with the sacrifice of depth. In some areas such as MRI reconstruction, more quantitative results could be collected, in order to obtain more concrete conclusions and further emphasise the impact TSV can have in this field. 

Linked to this is the fact that all code written for this project was executed on a CPU. Iterative algorithms such as these can be accelerated greatly when executed on a GPU. This would have enabled me to perform more experiments and conduct more in-depth studies on each application, allowing me to add more depth to the numerical results collected. On top of this, due to the time taken to run some of the algorithms, such as the algorithms for MRI reconstruction, I could only tune hyperparameters with limited accuracy. A GPU implementation, for example, would have overcome this problem.  
 
 Another area of potential research is combining algorithms when applying the proximal gradient method to advanced imaging applications. In order to apply accelerated proximal gradient algorithm to advanced imaging applications, the inner loop must be solved efficiently. Two findings are required in order to optimally apply such an algorithm. Firstly, to what degree of accuracy is it necessary to solve the inner loop in order to get desired results, and secondly, once this degree of accuracy is known, which algorithm can most efficiently achieve this accuracy. If only a low level of accuracy is required then ADMM may be more suitable than the dual proximal algorithm, however, if machine accuracy is required then ADPA will be more effective. It may also be possible that the level of accuracy required changes as the algorithm progresses. To get a moderate level of accuracy required in earlier iterations, only a small number of inner iterations may be required, however, to get machine accuracy more inner loop iterations may be required. This would open the door to potentially exploring adaptive methods to decide the number of inner iterations at any point.
 
 Going forward, the applications of the new regularisation term have only just scratched the surface. To truly unlock its potential, it would have to be extended to more challenging domains such as image registration, which are non-convex problems. There is also the application of TSV regularisation to non-smooth data terms to be considered. One area of further research which can be carried out is to investigate the relationship between the optimisation algorithms and the data term. The ideas in this thesis could naturally be extended to problems with non-smooth data terms using the APPA algorithm highlighted in \cite{lin2020near}.\\\\
\begin{center}
\Large{``We can only see a short distance ahead,\\but we can see plenty there that needs to be done."}\\\normalsize{\textsc{Alan Turing}}
\end{center}

\clearpage
\bibliography{ref}

\begin{thebibliography}{}

\bibitem[Balakrishnan et~al., 2019]{balakrishnan2019voxelmorph}
Balakrishnan, G., Zhao, A., Sabuncu, M.~R., Guttag, J., and Dalca, A.~V.
  (2019).
\newblock Voxelmorph: a learning framework for deformable medical image
  registration.
\newblock {\em IEEE transactions on medical imaging}, 38(8):1788--1800.

\bibitem[Boyd et~al., 2011]{boyd2011distributed}
Boyd, S., Parikh, N., and Chu, E. (2011).
\newblock {\em Distributed optimization and statistical learning via the
  alternating direction method of multipliers}.
\newblock Now Publishers Inc.

\bibitem[Bredies et~al., 2010]{bredies2010total}
Bredies, K., Kunisch, K., and Pock, T. (2010).
\newblock Total generalized variation.
\newblock {\em SIAM Journal on Imaging Sciences}, 3(3):492--526.

\bibitem[Chambolle and Pock, 2011]{chambolle2011first}
Chambolle, A. and Pock, T. (2011).
\newblock A first-order primal-dual algorithm for convex problems with
  applications to imaging.
\newblock {\em Journal of mathematical imaging and vision}, 40(1):120--145.

\bibitem[Chambolle and Pock, 2016]{chambolle2016introduction}
Chambolle, A. and Pock, T. (2016).
\newblock An introduction to continuous optimization for imaging.
\newblock {\em Acta Numerica}, 25:161--319.

\bibitem[Duan et~al., 2016]{duan2016denoising}
Duan, J., Lu, W., Tench, C., Gottlob, I., Proudlock, F., Samani, N.~N., and
  Bai, L. (2016).
\newblock Denoising optical coherence tomography using second order total
  generalized variation decomposition.
\newblock {\em Biomedical Signal Processing and Control}, 24:120--127.

\bibitem[Duan et~al., 2019]{duan2019vs}
Duan, J., Schlemper, J., Qin, C., Ouyang, C., Bai, W., Biffi, C., Bello, G.,
  Statton, B., O’regan, D.~P., and Rueckert, D. (2019).
\newblock Vs-net: Variable splitting network for accelerated parallel mri
  reconstruction.
\newblock In {\em International Conference on Medical Image Computing and
  Computer-Assisted Intervention}, pages 713--722. Springer.

\bibitem[Fortun et~al., 2015]{fortun2015optical}
Fortun, D., Bouthemy, P., and Kervrann, C. (2015).
\newblock Optical flow modeling and computation: A survey.
\newblock {\em Computer Vision and Image Understanding}, 134:1--21.

\bibitem[Gabay and Mercier, 1976]{gabay1976dual}
Gabay, D. and Mercier, B. (1976).
\newblock A dual algorithm for the solution of nonlinear variational problems
  via finite element approximation.
\newblock {\em Computers \& mathematics with applications}, 2(1):17--40.

\bibitem[Goldstein et~al., 2014]{goldstein2014fast}
Goldstein, T., O'Donoghue, B., Setzer, S., and Baraniuk, R. (2014).
\newblock Fast alternating direction optimization methods.
\newblock {\em SIAM Journal on Imaging Sciences}, 7(3):1588--1623.

\bibitem[Hammernik et~al., 2018]{hammernik2018learning}
Hammernik, K., Klatzer, T., Kobler, E., Recht, M.~P., Sodickson, D.~K., Pock,
  T., and Knoll, F. (2018).
\newblock Learning a variational network for reconstruction of accelerated mri
  data.
\newblock {\em Magnetic resonance in medicine}, 79(6):3055--3071.

\bibitem[Horn and Schunck, 1981]{horn1981determining}
Horn, B.~K. and Schunck, B.~G. (1981).
\newblock Determining optical flow.
\newblock {\em Artificial intelligence}, 17(1-3):185--203.

\bibitem[Horn and Johnson, 1990]{horn1990norms}
Horn, R.~A. and Johnson, C.~R. (1990).
\newblock Norms for vectors and matrices.
\newblock {\em Matrix analysis}, pages 313--386.

\bibitem[Huang et~al., 2008]{huang2008fast}
Huang, Y., Ng, M.~K., and Wen, Y.-W. (2008).
\newblock A fast total variation minimization method for image restoration.
\newblock {\em Multiscale Modeling \& Simulation}, 7(2):774--795.

\bibitem[Knoll et~al., 2011a]{https://doi.org/10.1002/mrm.22595}
Knoll, F., Bredies, K., Pock, T., and Stollberger, R. (2011a).
\newblock Second order total generalized variation (tgv) for mri.
\newblock {\em Magnetic Resonance in Medicine}, 65(2):480--491.

\bibitem[Knoll et~al., 2011b]{knoll2011second}
Knoll, F., Bredies, K., Pock, T., and Stollberger, R. (2011b).
\newblock Second order total generalized variation (tgv) for mri.
\newblock {\em Magnetic resonance in medicine}, 65(2):480--491.

\bibitem[Kobler et~al., 2020]{kobler2020total}
Kobler, E., Effland, A., Kunisch, K., and Pock, T. (2020).
\newblock Total deep variation for linear inverse problems.
\newblock In {\em Proceedings of the IEEE/CVF Conference on Computer Vision and
  Pattern Recognition}, pages 7549--7558.

\bibitem[Lin et~al., 2020]{lin2020near}
Lin, T., Jin, C., and Jordan, M.~I. (2020).
\newblock Near-optimal algorithms for minimax optimization.
\newblock In {\em Conference on Learning Theory}, pages 2738--2779. PMLR.

\bibitem[Lu et~al., 2016]{lu2016implementation}
Lu, W., Duan, J., Qiu, Z., Pan, Z., Liu, R.~W., and Bai, L. (2016).
\newblock Implementation of high-order variational models made easy for image
  processing.
\newblock {\em Mathematical Methods in the Applied Sciences},
  39(14):4208--4233.

\bibitem[Lustig et~al., 2007]{lustig2007sparse}
Lustig, M., Donoho, D., and Pauly, J.~M. (2007).
\newblock Sparse mri: The application of compressed sensing for rapid mr
  imaging.
\newblock {\em Magnetic Resonance in Medicine: An Official Journal of the
  International Society for Magnetic Resonance in Medicine}, 58(6):1182--1195.

\bibitem[Nesterov, 2013]{nesterov2013gradient}
Nesterov, Y. (2013).
\newblock Gradient methods for minimizing composite functions.
\newblock {\em Mathematical Programming}, 140(1):125--161.

\bibitem[{Nesterov}, 1983]{zbMATH03850830}
{Nesterov}, Y.~E. (1983).
\newblock {A method of solving a convex programming problem with convergence
  rate \(0(1/k^ 2)\)}.
\newblock {\em {Sov. Math., Dokl.}}, 27:372--376.

\bibitem[O’Donoghue and Candes, 2015]{o2015adaptive}
O’Donoghue, B. and Candes, E. (2015).
\newblock Adaptive restart for accelerated gradient schemes.
\newblock {\em Foundations of computational mathematics}, 15(3):715--732.

\bibitem[Rudin, 1987]{rudin1987images}
Rudin, L.~I. (1987).
\newblock {\em Images, numerical analysis of singularities and shock filters}.
\newblock PhD thesis, California Institute of Technology.

\bibitem[Rudin et~al., 1992]{rudin1992nonlinear}
Rudin, L.~I., Osher, S., and Fatemi, E. (1992).
\newblock Nonlinear total variation based noise removal algorithms.
\newblock {\em Physica D: nonlinear phenomena}, 60(1-4):259--268.

\bibitem[Schlemper et~al., 2017]{schlemper2017deep}
Schlemper, J., Caballero, J., Hajnal, J.~V., Price, A.~N., and Rueckert, D.
  (2017).
\newblock A deep cascade of convolutional neural networks for dynamic mr image
  reconstruction.
\newblock {\em IEEE transactions on Medical Imaging}, 37(2):491--503.

\bibitem[Sun et~al., 2016]{sun2016deep}
Sun, J., Li, H., Xu, Z., et~al. (2016).
\newblock Deep admm-net for compressive sensing mri.
\newblock {\em Advances in neural information processing systems}, 29.

\bibitem[Tikhonov, 1963]{tikhonov1963solution}
Tikhonov, A.~N. (1963).
\newblock On the solution of ill-posed problems and the method of
  regularization.
\newblock In {\em Doklady Akademii Nauk}, volume 151, pages 501--504. Russian
  Academy of Sciences.

\bibitem[Ye, 2019]{ye2019compressed}
Ye, J.~C. (2019).
\newblock Compressed sensing mri: a review from signal processing perspective.
\newblock {\em BMC Biomedical Engineering}, 1(1):1--17.

\bibitem[Zach et~al., 2007]{zach2007duality}
Zach, C., Pock, T., and Bischof, H. (2007).
\newblock A duality based approach for realtime tv-l 1 optical flow.
\newblock In {\em Joint pattern recognition symposium}, pages 214--223.
  Springer.

\bibitem[Zbontar et~al., 2018]{zbontar2018fastmri}
Zbontar, J., Knoll, F., Sriram, A., Murrell, T., Huang, Z., Muckley, M.~J.,
  Defazio, A., Stern, R., Johnson, P., Bruno, M., et~al. (2018).
\newblock fastmri: An open dataset and benchmarks for accelerated mri.
\newblock {\em arXiv preprint arXiv:1811.08839}.

\end{thebibliography}
\newpage
\appendix

\section{Discrete Cosine Transform Inverse} \label{App:FourInv}
The structure of the matrix we are trying to invert is as follows:
\[I+\lambda \nabla \nabla^{T}\]
we can first make the observation that I is diagonal and if we can diagonalise $\lambda \nabla \nabla^{T}$ then the eigenvalues, and therefore the inverse of this matrix can easily be found. We will first consider the 1D formulation of this problem, which can easily be generalised to 2D using the Kronecker product. The matrix we will try to diagonalise (and therefore invert is as follows):

\begin{equation} \label{eq:T}
{\cal {T}}_n = \left( {\begin{array}{*{20}{r}}
{-1}&1& & &\\
1&{-2}&1& &\\
 & \cdot & \cdot & \cdot &\\
&&1&{-2}&1\\
&&&1&{-1}
\end{array}} \right).
\end{equation}
a matrix of this structure can be analysed using the following identity: consider $c_{k} = \cos((k+\frac{1}{2})\theta)$ and $s_{k} = \sin((k+\frac{1}{2})\theta)$ 
\begin{center}
    $c_{k-1} =\cos(\theta)c_{k} + \sin(\theta)s_{k}$ and $c_{k+1} = \cos(\theta)c_{k}-\sin(\theta)s_{k}$
\end{center}
We can then construct the following identity:
\[c_{k-1} -2\cos(\theta)c_{k}+ c_{k+1} = 0\]
Now it is clear that this equation shares some properties with the above matrix, as the coefficients of the equation are the same as all but the top and bottom rows. So we can now consider the the following lemma:

\begin{lemma}
Suppose we have $\theta \in \mathbb{R}$ if $c_{k} = \cos((k+\frac{1}{2})\theta)$ for $k = 0,...,n-1$ we have:
\begin{equation} \label{eq:eigen1}
{\cal {T}}_n \left[ \begin{array}{c}
{c_0}\\
{c_1}\\
\, \vdots \\
{c_{n - 2}}\\
{c_{n - 1}}
\end{array} \right] = 2\left( {\cos (\theta)  - 1} \right)\left[ \begin{array}{c}
{c_0}\\
{c_1}\\
\, \vdots \\
{c_{n - 2}}\\
{c_{n - 1}}
\end{array} \right] + \left[ \begin{array}{c}
{c_0} - {c_{ - 1}}\\
0\\
\, \vdots \\
0\\
{c_{n - 1}} - {c_n}
\end{array} \right].
\end{equation}  
\end{lemma}

We can see that we are close to observing that the eigenvalues of ${{\cal{T}}_{n}}$ are $\lambda_{j} = 2(\cos(\theta_{j})-1)$ where $\theta_{j} = \frac{j\pi}{n}$, however the top and bottom terms of the remainder are preventing this from being the case. However, upon closer inspection we have:

    \[c_{0} - c_{-1} = \cos(\frac{1}{2}\theta) - \cos(-\frac{1}{2}\theta) = 0\]

As well as:
\[{c_{n - 1}} - {c_n} = \cos \left( {\left( {n - \frac{1}{2}} \right)\theta } \right) - \cos \left( {\left( {n + \frac{1}{2}}   \right)\theta} \right) 
=2\sin \left( {n\theta } \right)\sin \left( {\frac{\theta }{2}} \right)=0\]
We have now show that this for the 1D case the eigenvalues of ${\cal{T}}_{n}$ are $\lambda_{j}$ as stated before, and can formulate the following eigensystem:
\[{\cal{V}}^{-1}({\cal{I}}_{n} - {\cal{D}}_{n}){\cal{V}}\]
Where ${\cal{V}}$ is formed from the eigenvectors of the system above.

This is useful because we can now derive the following formula using the eigensystem above:
\[({\cal{I}}_{n} - {\cal{T}}_{n})^{-1} = {\cal{V}}({\cal{I}}_{n} - {\cal{D}}_{n})^{-1}{\cal{V}}^{-1}\]
Where ${\cal{D}}_n$ = diag($\lambda_{0},...,\lambda_{n-1}$). We can use this to solve linear equations of the form $({\cal{I}}_{n} - {\cal{T}}_{n})u = f$ in the following way:
\begin{equation} \nonumber
\begin{array}{l}
u \leftarrow {{\cal {V}}^{ - 1}}f,\\
u \leftarrow {({\cal {I}}_n-{\cal {D}}_n)^{-1}}u,\\
u \leftarrow {\cal {V}} u.
\end{array}
\end{equation}
By definition, one knows that the multiplication of a vector by $\cal{V}$ is  equivalent  to  performing  an  inverse discrete cosine transform (DCT) with to a discrete 1D signal. 

Throughout this thesis however we are dealing with the 2D case. The above ideas about eigenvalues can be transferred to the 2D case using the properties of the kronecker product. Since in the 2D case, we have:
\[{\cal{M}} = \nabla \nabla^{T} = ({\cal{T}}_{m} \otimes {\cal{I}}_{n} + {\cal{I}}_{m} \otimes {\cal{T}}_{n})\]
The eigenvalues can then be calculated using the following theorem (which follows directly from properties of the kronecker product):
\begin{theorem}
Suppose ${\cal {M}} = {\cal {I}}_m \otimes {\cal {I}}_n - ({\cal {T}}_m \otimes {\cal {I}}_n + {\cal {I}}_m \otimes {\cal {T}}_n) $ is non-singular with ${\cal {T}}_m \in \mathbb{R}^{m \times m}$ and ${\cal {T}}_n \in \mathbb{R}^{n \times n}$. If ${\cal {V}}_m^{-1} {\cal {T}}_m {\cal {V}}_m = {\cal {D}}_m$, ${\cal {V}}_n^{-1} {\cal {T}}_n {\cal {V}}_n = {\cal {D}}_n$, and $\bm{f} \in \mathbb{R}^{mn}$, then the solution to ${\cal {M}} \bm{u}=\bm{f}$ is given by 
\begin{equation}\nonumber
\bm{u} = \left( {{\cal {V}}_m \otimes {\cal {V}}_n} \right){\cal{A}}\left( {{\cal {V}}_m^{-1} \otimes {\cal {V}}_n^{-1}} \right) \bm{f},
\end{equation}
where ${\cal{A}} = \left[ {\cal {I}}_m \otimes {\cal {I}}_n - ({\cal {D}}_m \otimes {\cal {I}}_n + {\cal {I}}_m  \otimes  {\cal {D}}_n)\right]^{-1}$.
\end{theorem}

So we have the following procedure when calculating u:
\begin{equation} \label{eq:2Dsolution}
\begin{array}{l}
\bm{u} \leftarrow ({{\cal {V}}_m^{-1} \otimes {\cal {V}}_n^{-1}})\bm{f},\\
\bm{u} \leftarrow [{\cal {I}}_m \otimes {\cal {I}}_n - ({\cal {D}}_m \otimes {\cal {I}}_n + {\cal {I}}_m  \otimes  {\cal {D}}_n)]^{-1}\bm{u},\\
\bm{u} \leftarrow ({{\cal {V}}_m \otimes {\cal {V}}_n}) \bm{u}.
\end{array}
\end{equation}
Since we have that the multiplication of a vector by ${{\cal {V}}_m^{-1} \otimes {\cal {V}}_n^{-1}}$ is equivalent to applying the 2D DCT to the 2D matrix, the above proccess can easily be performed on an image. Since  ${\cal {I}}_m \otimes {\cal {I}}_n - ({\cal {D}}_m \otimes {\cal {I}}_n + {\cal {I}}_m  \otimes  {\cal {D}}_n)]^{-1}$ is a diagonal matrix, this calculation can simply be expressed as an elementwise division on an image the discrete cosine transform of an image $f$:
\begin{equation} \nonumber
\begin{split}
  [{\cal {I}}_m \otimes {\cal {I}}_n - ({\cal {D}}_m \otimes {\cal {I}}_n + {\cal {I}}_m  \otimes  {\cal {D}}_n)]^{-1}\bm{u} \\
  = \frac{{\cal {U}}}{1- \left( {2\cos \left( {\frac{{q\pi }}{m}} \right) + {2\cos \left( {\frac{{r\pi }}{n}} \right)}  - 4} \right)}
\end{split}
\end{equation}
where $r\in[1,n]$ and $q\in[1,m]$ are the integer indices and ${\cal {U}} \in \mathbb{R}^{m \times n}$ is equivalent to $\bm{u}$ after reshaping back to the $m \times n$ sized image and having the 2 dimensional Fourier transform applied. ${1- \left( {2\cos \left( {\frac{{q\pi }}{m}} \right) + {2\cos \left( {\frac{{r\pi }}{n}} \right)}  - 4} \right)}$ and the eigenvalues of the matrix ${\mathcal{M}}$. In some cases in the thesis this idea is generalised, however the same idea can be applied. 
The handling of this method gives us one insight into how the boundary conditions can dictate a method. If the boundary conditions we were using were periodic instead of symmetric, then the matrix ${{\cal{T}}_{n}}$ would be a matrix of a different form. Both the top and bottom rows would change to follow the pattern of the rest of the matrix. In this case, the trigonometric identities discussed discussed no longer fit the matrix. When this is the case, it actually turns out this method can be modified and the discrete fourier transform can be used to find the inverses rather than the discrete cosine transform.

The above appendix outlines how both the inverse and the eigenvalues of these matrices can be calculated. This appendix is referred to throughout the report, it should be noted that often modifications of this method are used to derive the eigenvalues and inverses of matrices with similar structural properties to ${\mathcal{T}_{n}}$. 

\section{Proximal operator of $\delta(\cdot)$} \label{app:ProxDelt}
In order to evaluate the proximal operator for $\delta(\cdot)$ evaluated at some vector $\mathbf{p} \in \mathbb{R}^{mn}$ then we must solve the following minimisation problem:
\[\mathbf{prox}_{\delta(\cdot),t}(\mathbf{p}) = \argminB_{\mathbf{z}}\frac{1}{2t}\|\mathbf{p}-\mathbf{z}\|^{2}_{2} + \delta(\mathbf{z})\]
where
\begin{equation}
\delta(z)_{i} =
\begin{cases}
\infty & \sqrt{\sum_{j}z_{i,j}^{2}}>1 \\
0 & \sqrt{\sum_{j}z_{i,j}^{2}}\leq1
\end{cases}
\end{equation}
Where we are summing over the number of channels, in most cases throughout this thesis the number of channels is 2. 
In order to minimise this function we can now deal with each pixel separately, since they don't interact. This means the process can be considered in a element-wise fashion. So at that given point, if $p_{i} \in [-1,1]$ then no further action is needed. However if $p_{i} \notin [-1,1]$ the value of $z_{i}$ which minimises $\|p_{i}-z_{i}\|^{2}_{2}$ and is in the set $[-1,1]$ must be found. This value is in fact the projection of $p_{i}$ into the unit ball. So we conclude that the proximal operator of $\delta(p)$ is simply the pixelwise projection of $\mathbf{p}$ into the unit ball:
\begin{equation} \label{eq:ProxDelta}
    \left[\mathbf{prox}_{\delta(\cdot),t}(\mathbf{p})\right]_i = \frac{p_{i}}{|p_{i}|}
\end{equation}

Note in the case of the isotropic TV regularisation term the constraint becomes:
\begin{equation}
\delta(z)_{i} =
\begin{cases}
\infty & |z_{i,j}|>1 \text{ for any j} \\
0 & |z_{i,j}|\leq 1 \text{ for any j}
\end{cases}
\end{equation}
Where $i$ is the pixel value and $j$ represents the channel, therefore it is clear that the projection in this case becomes:
\begin{equation} \label{eq:ProxDelta}
    \left[\mathbf{prox}_{\delta(\cdot),t}(\mathbf{p})\right]_{i,j} = \frac{p_{i,j}}{|p_{i,j}|}
\end{equation}
\section{Code}
The code can be found at \href{https://github.com/Jbartlett6/Accelerated-First-Order-Method-for-Variational-Imaging}{https://github.com/Jbartlett6/Accelerated-First-Order-Method-for-Variational-Imaging}.
All code can simply be ran by opening the function and running it in Matlab. The image the algorithm is acting on as well as the hyperparameters can be changed internally within each function.

\end{document}